\documentclass{article}
\usepackage[ruled,vlined,linesnumbered]{algorithm2e}
\usepackage[T1]{fontenc}
\usepackage{microtype}
\usepackage{graphicx}
\usepackage{booktabs}
\usepackage{array}
\usepackage{placeins}
\usepackage{pifont}
\usepackage{tikz}
\usepackage{multirow}
\usepackage{capt-of}
\usepackage{amsmath,amssymb}
\usepackage{hyperref}

\usepackage[accepted]{mlsys2025}
\makeatletter\renewcommand{\Notice@String}{}\makeatother
\hypersetup{pdfsubject={}}
\usepackage{xcolor}
\usepackage{xspace}

\definecolor{RefRed}{HTML}{B00000}
\definecolor{LinkBlue}{HTML}{0000FF}

\hypersetup{
    colorlinks=true,
    linkcolor=LinkBlue,
    citecolor=violet,
    urlcolor=LinkBlue
}

\usepackage{enumitem}
\newenvironment{denseitemize}{
\begin{itemize}[topsep=2pt, partopsep=0pt, leftmargin=1.5em]
  \setlength{\itemsep}{2pt}
  \setlength{\parskip}{0pt}
  \setlength{\parsep}{0pt}
}{\end{itemize}}

\newcommand{\xmark}{\ding{55}}
\newcommand{\hbnone}{\tikz[baseline=-0.6ex]\draw (0,0) circle (0.6ex);}
\newcommand{\hbhalf}{\tikz[baseline=-0.6ex]{\draw (0,0) circle (0.6ex);\fill (0,-0.6ex) arc (-90:90:0.6ex) -- cycle;}}
\newcommand{\hbfull}{\tikz[baseline=-0.6ex]\fill (0,0) circle (0.6ex);}

\usepackage[normalem]{ulem}
\newcommand{\zul}{\bgroup\markoverwith{\textcolor{teal}{\rule[-0.55ex]{2pt}{0.5pt}}}\ULon}

\usepackage{xcolor}

\definecolor{SoftOrange}{RGB}{230, 145, 56} 
\definecolor{MutedOrange}{RGB}{200, 130, 70}
\definecolor{darkorange}{rgb}{1.0, 0.55, 0.0}
\definecolor{deepcarrotorange}{rgb}{0.99, 0.41, 0.17}
\definecolor{ppurple}{rgb}{0.6, 0.0, 0.5}
\definecolor{plum}{rgb}{0.56, 0.27, 0.52}
\definecolor{cardinal}{rgb}{0.77, 0.12, 0.23}

\SetCommentSty{mycommfont}
\def\name{Reflex\xspace}

\mlsystitlerunning{Reflex: Fast Robot Policy Serving by Reusing and Correcting Past Actions}

\begin{document}
\twocolumn[
\mlsystitle{Reflex: Fast Robot Policy Serving by Reusing and Correcting Past Actions}
\begin{mlsysauthorlist}
\mlsysauthor{Ziyang Lin}{uiuc}
\mlsysauthor{Zixuan Sun}{uiuc}
\mlsysauthor{Rami Nasser}{uiuc}
\mlsysauthor{Jianli Jin}{uiuc}
\mlsysauthor{Xuanle Li}{uiuc}
\mlsysauthor{Jayanth Srinivasa}{cisco}
\mlsysauthor{Fan Lai}{uiuc}
\end{mlsysauthorlist}
\mlsysaffiliation{uiuc}{University of Illinois at Urbana-Champaign}
\mlsysaffiliation{cisco}{Cisco}
\mlsyscorrespondingauthor{Ziyang Lin}{ziyang10@illinois.edu}
\mlsyscorrespondingauthor{Fan Lai}{fanlai@illinois.edu}
% Other author emails: zixuans8@illinois.edu, rnass2@illinois.edu, jianlij2@illinois.edu, xuanlel2@illinois.edu, jasriniv@cisco.com
\mlsyskeywords{robot policies, inference caching, efficient serving}
\vskip 0.3in
\begin{abstract}
Modern robot policies often perform expensive inference to generate action chunks, making serving
a latency bottleneck for real-time control and a
throughput bottleneck for multi-robot serving.
Building on the success of caching in image, video,
and audio generation, recent robot-policy advances
have reported promising speedups by reusing intermediate
denoising states to avoid redundant computation.
Surprisingly, we find that such reuse often offers little advantage over simply running fewer denoising steps from fresh noise, without any cache. 
Moreover, accelerating denoising leaves
the expensive policy backbone intact, which can
account for 70\% of GPU time.

We present \name, a serving layer that turns
past policy executions into a backbone-free
path for future decisions. Rather than reusing
intermediate computations, \name caches complete
action chunks to bypass both the backbone and
action head. This introduces two challenges:
similar observations can demand different actions,
and small action deviations (errors) in reuse can compound into execution
stalls. 
\name addresses these through two-level cache
correction. 
At the action level, it learns action-aware
similarity from past policy decisions, assembles
compatible chunks, and applies lightweight
residual correction to limit action mismatch.
At the execution level, it captures task progress
and adaptively refreshes the original policy to limit
accumulated errors and balance inference cost.
Without modifying the underlying policy,
\name improves inference latency by
1.4$\times$--11$\times$ and throughput by 2.0$\times$--2.2$\times$ on representative workloads and real robots, while preserving comparable task success.
\end{abstract}
]
\printAffiliationsAndNotice{}

\section{Introduction}
\label{sec:intro}

Modern robot policies are increasingly capable, but their computational cost poses a challenge for real-time deployment. Recent advances in diffusion-based policies \cite{chi2024diffusionpolicyvisuomotorpolicy}, vision--language--action (VLA) models \cite{kim2024openvlaopensourcevisionlanguageactionmodel,black2026pi0visionlanguageactionflowmodel,nvidia2025gr00tn1openfoundation}, and world-action models (WAMs) \cite{ye2026worldactionmodelszeroshot,yuan2026fastwamworldactionmodels,kim2026cosmospolicyfinetuningvideo} enable robots to generate chunks of future controls from camera observations, task instructions, and robot states, often through iterative denoising.
Unlike one-shot generative inference, robot control
invokes the policy repeatedly throughout task execution,
placing inference on the critical path of closed-loop
control. For example, a robot operating at 20\,Hz
and executing five actions per chunk must replan
every 250\,ms.
At fleet scale, robots performing warehouse fulfillment,
industrial assembly, or laboratory automation may share
an on-premises GPU server or a cloud-hosted model rather than each hosting a
large policy model \cite{geminiroboticsteam2025geminiroboticsbringingai,jiang2026rosaroboticsfoundationmodel,adamopoulos2026efficientvisionlanguageactionmanagementserving}.
Each policy call delays its requesting robot while
consuming GPU capacity needed by others, making
inference both a latency and a throughput bottleneck~\cite{kairos-arxiv26}.

A natural principle is to reuse intermediate states
of the iterative denoising process. This approach has
shown successful premise in image, video, and audio generation
\cite{agarwal2023approximatecachingefficientlyserving,ma2023deepcacheacceleratingdiffusionmodels,liu2025timestepembeddingtellsits,zhao2025realtimevideogenerationpyramid,barik2026soundweavercompositionalwarmstartingtexttoaudio}, and recent robot-policy advances
have adopted it to reduce action-generation cost
\cite{oi2026actioncachetrainingfreeaccelerationvisionlanguageaction,chen2025falconfastvisuomotorpolicies,zhao2026c3acheacceleratingworldaction,liu2026efficientworldactionmodel}.
However, our measurements reveal two (surprising) limitations.
First, denoising reuse offers little advantage over simply running fewer denoising steps from fresh
noise. At matched step budgets, denoising reuse
can even reduce task success by 13.4 percentage
points.
Second, accelerating denoising leaves the backbone
cost intact, which can account for 70\% of GPU time, limiting
the speedup from eliminating the entire action head
to just 1.17$\times$ (\S\ref{sec:motivation}).

In this paper, we explore a different
approach: \emph{reusing complete policy decisions instead of
intermediate computations}. Our key insight is that robot 
tasks often repeat, even when trajectories do not.
In a familiar workcell, different initial object
configurations often produce different trajectories, yet
many individual decisions still call for similar
actions. Caching action chunks from past successful
executions can serve new decisions without
running the policy backbone.

Making complete-action caching reliable poses two
fundamental challenges. First, \emph{action mismatch}: similar
observations may require different actions.
A slight change in object pose can alter the
required grasp, while substantial background
changes may leave the appropriate motion unchanged.
The cache must identify action-compatible
states rather than merely visually similar ones. Second, \emph{execution drift}: even individually
plausible cached actions can compound errors
across decisions, eventually
stalling execution. Reliable reuse must address per-decision action deviations and their effects
over an entire task trajectory (episode).

\name caches complete action chunks from past successful
policy executions and reuses them using features
from the policy's own vision encoder, so it generalizes to the popular VLA, WAM, and diffusion policies. 
At the action level, it learns a task-specific
similarity metric from past decisions whose actions
agree, assembles compatible cached chunks, and
applies lightweight residual correction to reduce
action deviation. Importantly, past
decisions already provide supervision for both selection
and correction, without additional policy inference
or modification of policy weights (\S\ref{sec:reuse}).

At the execution level, \name monitors task progress
by aligning recent observations with successful
episodes. When estimated progress falls below
the past, it selectively invokes
the original policy to address potential stalls.
Task-adaptive periodic refresh further limits
error accumulation before stalls become apparent,
balancing cache reuse against inference cost.
These mechanisms allow \name to bypass
expensive policy computation on cache hits while
retaining policy feedback when needed
(\S\ref{sec:correct}).

We evaluate \name across five policies spanning VLAs,
WAMs, and diffusion policies, on LIBERO-Spatial,
LIBERO-Long, and MetaWorld MT50. At comparable task success targets, \name reduces amortized inference time by
1.4--11$\times$ (3.3$\times$ on average).
Under settings with fewer denoising steps and longer
execution horizons, it achieves 9.3$\times$ average latency
savings over default pure inference.
On a shared RTX 4090, \name improves multi-robot serving throughput by
2.0--2.2$\times$ over pure inference. On a physical SO-101
arm, it serves 79.3\% of decisions from the cache
and reduces decision latency by 3.1$\times$ (\S\ref{sec:eval}).

Overall, this paper makes three key contributions:
\begin{denseitemize}
    \item
    We show the (surprising) limits of denoising reuse relative
    to plain step reduction and motivate caching
    complete actions to bypass backbone computation (\S\ref{sec:motivation}).

    \item 
    We introduce a novel backbone-free action reuse mechanism, proposing action-aware
    chunk assembly, lightweight residual correction,
    and progress-guided policy refresh without modifying the policy model (\S\ref{sec:method}--\S\ref{sec:impl}).

    \item 
    We demonstrate \name's substantial latency and throughput improvements in real-world workloads (\S\ref{sec:eval}).
    
\end{denseitemize}
% We will release the code, the cache libraries, and our trained policy checkpoints upon publication.

\begin{figure}[t]
\centering
\includegraphics[width=\columnwidth]{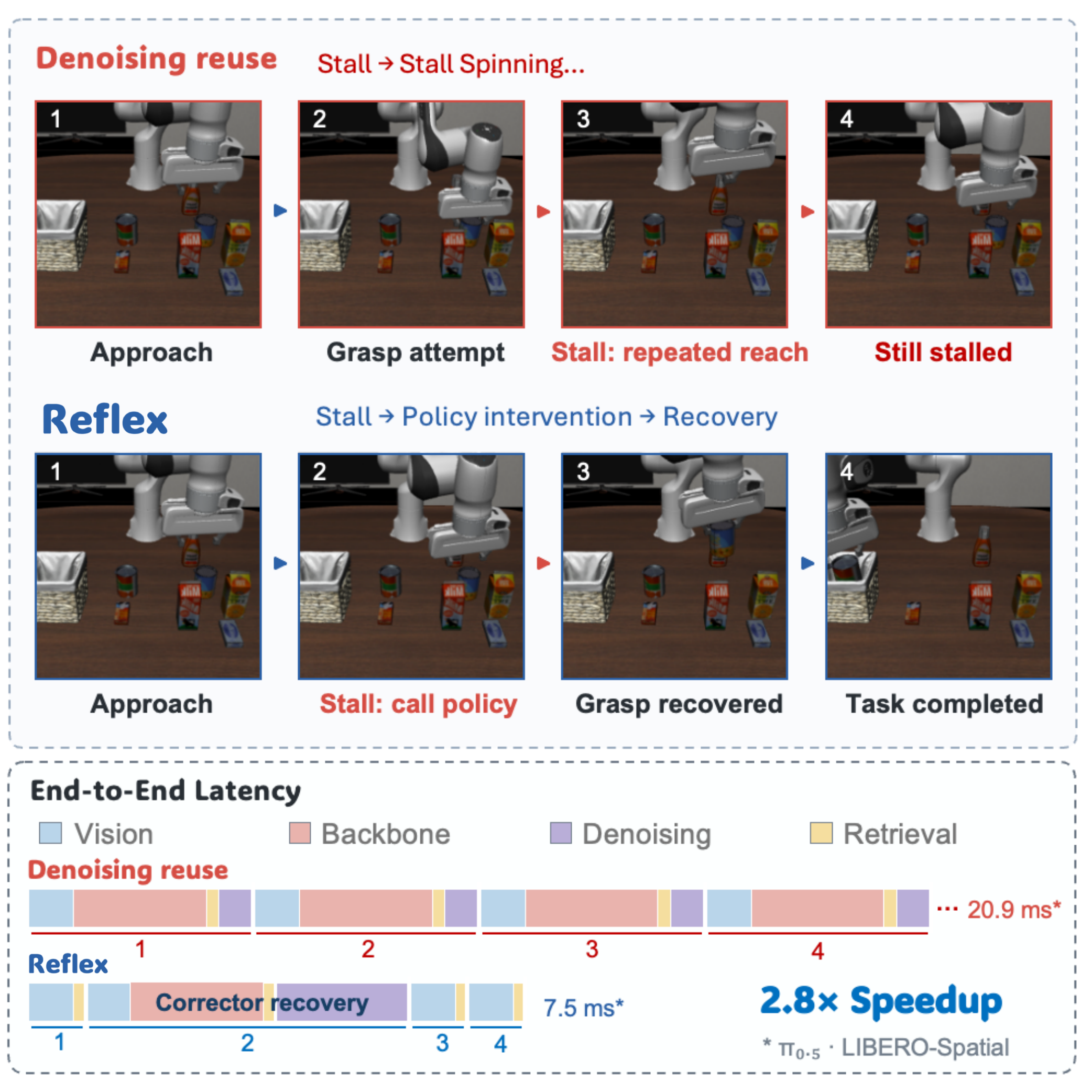}
\vspace{-.6cm}
\caption{\name reuses complete action chunks bypassing both backbone and action head, with selective policy calls to recover from stalls. \name reduces latency from 20.9 ms to 7.5 ms ($2.8\times$) versus inference with denoising  reuse (\S\ref{sec:e2e}).}
\vspace{-.4cm}
\label{fig:teaser}
\end{figure}

\section{Background and Motivation}\label{sec:motivation}

\subsection{Robot Policy Inference}
\label{subsec:robot}

Modern robot policies often generate action chunks,
sequences of future controls conditioned on the current
observation and task instruction
\cite{chi2024diffusionpolicyvisuomotorpolicy,black2026pi0visionlanguageactionflowmodel}.
An action-chunking policy predicts $H$ future controls,
but the robot may execute only the first $h\le H$
before observing and replanning. Predicting a longer
chunk captures a coherent sequence of actions, while
executing a shorter horizon allows the robot to respond
to new observations and correct deviations. Thus, executing $T$ control steps requires approximately $T/h$ policy calls.

As shown in Figure~\ref{fig:teaser}, the VLA and WAM policies divide each inference call
into three stages: (i) a vision encoder processes
camera observations; (ii) a backbone combines visual
features with the task instruction and robot state; 
and (iii) an action head generates the chunk through
iterative denoising. 
By iterative denoising, the action head repeatedly refines
the actions using conditioning computed once by the
backbone. For example, $\pi_{0.5}$ uses a 2B-parameter
backbone and a 0.3B-parameter action expert, with ten
denoising steps by default \cite{intelligence2025pi05visionlanguageactionmodelopenworld}.
Considering the costs of these stages as $t_v$, $t_b$,
and $t_h$, a policy call takes $t_v+t_b+t_h$.
Reducing denoising steps lowers only $t_h$,
leaving the vision encoder and backbone unchanged.

\subsection{Limitations of Denoising Caches}

Caching denoising states is a natural way to cut computation and has accelerated image,
video, and audio generation \cite{agarwal2023approximatecachingefficientlyserving,ma2023deepcacheacceleratingdiffusionmodels,liu2025timestepembeddingtellsits,barik2026soundweavercompositionalwarmstartingtexttoaudio}. Recent robot-policy
advances adopt a similar approach: reuse intermediate
states from previous decisions to skip early denoising
steps, while refining the reused states for current
observation \cite{oi2026actioncachetrainingfreeaccelerationvisionlanguageaction,chen2025falconfastvisuomotorpolicies,Duan_2025}.
The intuition is appealing, as similar observations may call for similar actions, but we identify two limitations: \emph{denoising reuse offers no consistent advantage
over plain step reduction}, and
\emph{accelerating denoising leaves substantial
backbone computation untouched}. 

% The most natural way to cut this cost is to cache the denoising process.
% Approximate caching in image generation reuses the intermediate noisy states of earlier requests to skip denoising steps \todo{cite Agarwal 2024}, and several training-free accelerations of robot policies start the action head from a cached or earlier intermediate state \todo{cite Oi 2026; Chen 2025; Duan 2025}.
% We reproduce this warm start from the nearest entry of our library and compare it with plain step reduction from noise at the same number of steps.

\begin{figure}[t]
\centering
\includegraphics[width=\columnwidth]{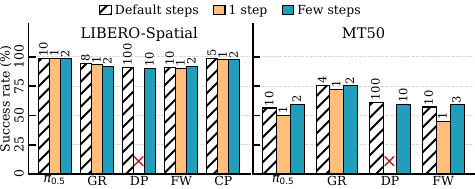}
\vspace{-.6cm}
\caption{Plain step reduction barely changes the success rate, and a few more steps recover what one step loses. Policies are $\pi_{0.5}$, GR00T N1.5, Diffusion Policy, Fast-WAM, and Cosmos Policy. Bar labels show the denoising steps, where DP fails with one step.
}
\label{fig:steps}
\end{figure}

\paragraph{Plain step reduction already keeps the success rate.}
Existing denoising-reuse advances report speedups over full denoising at nearly the same success rate \cite{oi2026actioncachetrainingfreeaccelerationvisionlanguageaction,chen2025falconfastvisuomotorpolicies,Duan_2025}. Yet such comparisons conflate the savings from caching with those from executing fewer denoising steps. This plain step reduction samples from fresh noise using fewer steps, without a cache, which yet has been largely overlooked \cite{Duan_2025,zhao2026c3acheacceleratingworldaction,zhou2026rmuscleroboticmusclememory,yang2025efficientvlatrainingfreeaccelerationcompression,he2026vlarollingfewerdenoisingsteps}. We evaluate this appealing baseline with five recent policies on LIBERO \cite{liu2023liberobenchmarkingknowledgetransfer} and MetaWorld MT50 \cite{yu2021metaworldbenchmarkevaluationmultitask}. Surprisingly, as shown in Figure~\ref{fig:steps}, one step changes the success rate by at most 1.4 points on LIBERO-Spatial. On MT50, one step can lose as much as 12.5 points, but two or three steps recover it. Diffusion Policy likewise retains most of its success with ten DDIM steps, while Cosmos Policy and X-WAM {\cite{guo2026unified4dworldaction}} lose at most 4.4 points with one step on RoboCasa {\cite{nasiriany2024robocasalargescalesimulationeveryday}} and RoboTwin 2.0 {\cite{chen2025robotwin20scalabledata}}.

\begin{figure}[t]
\centering
\includegraphics[width=\columnwidth]{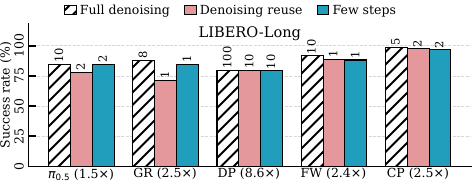}
\vspace{-.6cm}
\caption{At matched denoising-step budgets, denoising reuse
shows no consistent success-rate advantage over plain
step reduction, losing up to 13.4 percentage points.
Parentheses report the speedup of denoising reuse
relative to full-step inference.
Appendix~\ref{app:steps} provides measurement details.
}
\label{fig:warm}
\end{figure}

\begin{table}[t]
\centering
\footnotesize
\begin{tabular}{@{}lcc@{}}
\toprule
& Batch 1 (ms) & Batch 16 (ms) \\
\midrule
Vision encoder & 6.6 & 55.3 \\
Backbone & 17.9 & 252.4 \\
Action head, 10 steps & 26.3 & 53.0 \\
Action head, 1 step & 3.4 & 5.9 \\
\bottomrule
\end{tabular}
\vspace{-3.0pt}
\caption{Backbone computation dominates policy-call latency, while action-head latency barely grows with the batch. GPU time (ms) of each stage of $\pi_{0.5}$ on one RTX 4090 for one call at batch size 1 and for one batch of 16 calls.}
\label{tab:stages}
\end{table}

More surprisingly, reusing cached denoising states
does not necessarily outperform this naive baseline.
We evaluate the warm-start strategy underlying semantic denoising caches. At the same number of remaining steps on LIBERO-Long (Figure~\ref{fig:warm}), this nearest-entry warm start gains at most 0.8 points over fresh-noise sampling for Diffusion Policy, Fast-WAM, and Cosmos Policy, and even loses 7.0 points for $\pi_{0.5}$ and 13.4 for GR00T N1.5. A cached state can bias generation toward an action unsuited to the current observation, which a few refinement steps may not undo. Although more sophisticated caches may mitigate this error, these results show that denoising reuse offers no clear advantage.
% over the much simpler few-step baseline. 

\paragraph{Denoising reuse leaves a substantial compute floor.}
When the denoising model constitutes the entire policy,
as in Diffusion Policy and Cosmos Policy, plain step
reduction alone achieves 8.6$\times$ and 2.5$\times$
speedups ($100\!\to\!10$ and $5\!\to\!2$ steps).
For policies with separate backbones, such as
$\pi_{0.5}$, GR00T N1.5, and Fast-WAM, even perfect
denoising reuse leaves the vision encoder and backbone
computation intact. For example, ActionCache reports
up to 10.44$\times$ and 40.17$\times$ acceleration
of the action head
\cite{oi2026actioncachetrainingfreeaccelerationvisionlanguageaction},
yet end-to-end inference speedups for existing VLA
and WAM optimizations remain modest, generally around
1.2--2.5$\times$.
To compare these methods fairly despite differences
in hardware, baselines, and measurement settings,
we re-evaluate ten prior caching-based inference optimizations
under a common optimized implementation, alongside
their originally reported results
(Appendix~\ref{app:convert}). Surprisingly, 
as Table~\ref{tab:speedup} shows, these methods
offer no (clear) latency advantage over plain
step reduction.
% \jay{Sec 2.2 / Table 2 say "re-evaluate" and "reimplementations", but Appendix B describes analytic estimates that take the value most favorable to each method. Please make these consistent (label the column "estimated upper bound on our stack", or actually run the methods) and add success rates. Alternative: add one sentence in the Table 2 caption saying which methods were run and which were estimated.}

Beyond single-request latency, we further evaluate batched inference
to understand their serving throughput performance
(Table~\ref{tab:stages}). On an RTX 4090, the ten-step
action head of $\pi_{0.5}$ accounts for 52\% of GPU
time at batch size 1. At batch size 16, however,
the backbone dominates at 70\%, while the action
head accounts for only 15\%. This shift limits
the benefits of denoising-only acceleration under
batched serving. Hence, even eliminating the action head
entirely yields a maximum speedup of
$S_{\mathrm{head}}^{\max}=
(t_v+t_b+t_h)/(t_v+t_b)$,
where $t_v$, $t_b$, and $t_h$ denote the GPU times
of the vision encoder, backbone, and action head (\S\ref{subsec:robot}).
This bound drops from 2.07$\times$ at batch size 1
to just 1.17$\times$ at batch size 16.
Plain step reduction ($10\!\to\!1$) already achieves
1.82$\times$ and 1.15$\times$, respectively,
nearly saturating these limits.
Thus, further denoising optimization offers
diminishing returns, especially under batched
serving, motivating reuse that bypasses the backbone.

\begin{table}[t]
\centering
\scriptsize
\setlength{\tabcolsep}{4pt}
\begin{tabular}{@{}llrr@{}}
\toprule
Method & Policy & Reported & \begin{tabular}[b]{@{}r@{}}Re-Evaluated\\(ours)\end{tabular} \\
\midrule
\multicolumn{4}{@{}l}{\emph{VLA policies}} \\
ActionCache & $\pi_{0.5}$ & 2.12$\times$ & 1.62$\times$ \\
AdaVLA & $\pi_{0.5}$ & 1.87$\times$ & $\le$1.96$\times$ \\
EfficientVLA & CogACT & 1.93$\times$ & 1.80$\times$ \\
VLA-Rolling & $\pi_{0.5}$ & 1.30$\times$ & 1.19$\times$ \\
rMuscle & $\pi_{0.5}$ & 1.29$\times$ & 1.29$\times$ \\
Plain step reduction & $\pi_{0.5}$ & N/R & 1.62$\times$ \\
 & GR00T & N/R & 2.01$\times$ \\
\name (ours) & $\pi_{0.5}$, GR00T & N/R & 5.20, 6.56$\times$ \\
\midrule
\multicolumn{4}{@{}l}{\emph{World action models}} \\
C$^3$ache & Fast-WAM & 2.51$\times$ & 2.10$\times$ \\
WAMachine & Cosmos & 2.36$\times$ & 2.36$\times$ \\
Plain step reduction & Fast-WAM & N/R & 2.23$\times$ \\
 & Cosmos & N/R & 2.51$\times$ \\
\name (ours) & Fast-WAM, Cosmos & N/R & 5.91, 3.38$\times$ \\
\midrule
\multicolumn{4}{@{}l}{\emph{Diffusion Policy}} \\
Falcon & DP & 7.78$\times$\rlap{$^{*}$} & 12.69$\times$ \\
BAC & DP & 3.60$\times$ & 9.21$\times$ \\
RA-DP & DP & 3.6$\times$\rlap{$^{*}$} & 10.27$\times$ \\
Plain step reduction & DP, 10 steps & N/R & 8.62$\times$ \\
 & DP, 5 steps & N/R & 15.91$\times$ \\
\name (ours) & DP & N/R & 72.23$\times$ \\
\bottomrule
\end{tabular}
\vspace{-3.0pt}
\caption{Prior denoising-reuse methods show no (clear) efficiency
advantage over plain step reduction
(Appendix~\ref{app:convert}).
\emph{Reported} lists speedups from the original papers,
while \emph{Re-evaluated} shows results from our
reimplementations for controlled experiments in the same setting.
$^{*}$ denotes reported computation savings rather than
measured latency speedups. N/R denotes not reported. \name uses reduced steps and expanded horizons, as a geometric mean over benchmarks.}
\label{tab:speedup}
\vspace{-.4cm}
\end{table}

\paragraph{Opportunity: Reuse complete actions, not intermediate denoising states.}
In practical robot deployments, although individual executions differ in object placements and intermediate robot states, they often revisit similar situations requiring similar actions. This similarity creates an opportunity to reuse previous policy decisions, but directly replaying past actions is unreliable when the state differs. Our key insight is to cache \emph{complete action chunks} rather than intermediate denoising states, reusing and adapting them using features from the policy's vision encoder. On a cache hit, we can bypass both the backbone and action head but must correct action mismatches and recover from execution errors.

\section{\name Design}\label{sec:method}

\name is a lightweight serving layer that reduces inference latency and GPU demand without modifying the underlying robot policy\footnote{\name never trains or fine-tunes the policy and needs no GPU for fitting; its metric and corrector are able to be fitted on a CPU in minutes before deployment (Appendix~\ref{app:impl}).}. It supports diverse action-chunking policies, including VLA models, world-action models, and diffusion policies, by reusing complete action chunks from successful executions instead of invoking expensive action-generation networks. Given variations in object placements, robot
configurations, and intermediate execution outcomes,
\name does not assume that repeated tasks follow
identical trajectories. Instead, it exploits
recurring decision contexts across different
executions to reuse previously generated actions. This introduces two challenges: (i) \emph{Action mismatch}: similar observations may require different actions; and (ii) \emph{Execution drift}: small errors across successive cached decisions can accumulate and eventually stall the task.

\begin{figure*}[t]
\centering
\includegraphics[width=\textwidth]{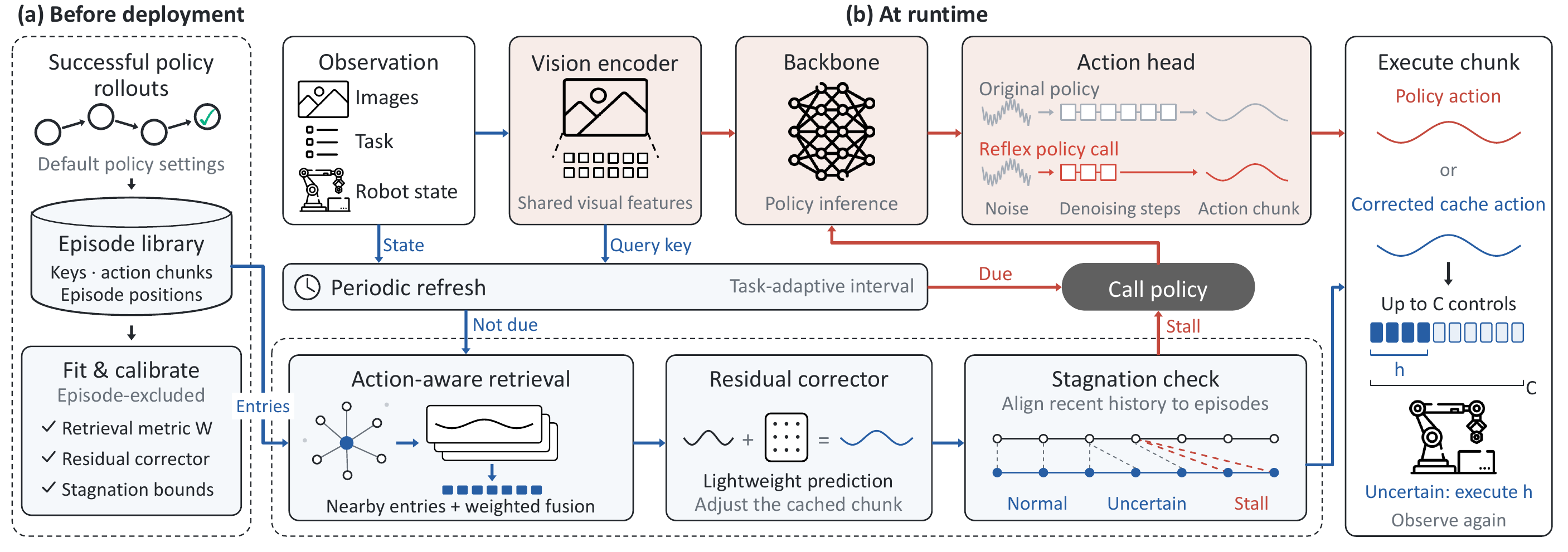}
\vspace{-.5cm}
\caption{\name serves most decisions without running the policy. \textcolor[RGB]{39,94,164}{Blue} marks the path of the cache, through action-aware retrieval (\S\ref{sec:reuse}), the residual corrector, and the stagnation check (\S\ref{sec:correct}). \textcolor[RGB]{196,73,61}{Red} marks the path of the policy itself, which runs only when a periodic refresh (\S\ref{sec:correct}) is due or a stall is detected.}
\label{fig:overview}
\end{figure*}

\noindent \textbf{Workflow.}
Figure~\ref{fig:overview} illustrates how \name addresses these challenges. Successful policy executions populate a cache of individual decisions and provide supervision for fitting an action-aware retrieval metric and a residual corrector. At runtime, the policy's vision encoder produces a cache query, and action-aware assembly combines compatible chunks from potentially different past episodes (\S\ref{sec:reuse}). Residual correction adapts the assembled chunk to the current state, while execution-level recovery selectively invokes the original policy when progress stalls (\S\ref{sec:correct}).

% \paragraph{Offline preparation and online workflow.}
% We collect successful episodes using the unmodified policy. Each decision contributes its task instruction, vision-derived representation, robot state, and predicted action chunk to a library. The same episodes also calibrate the components used at deployment. To avoid tuning on trivially similar frames from the same trajectory, we use \emph{leave-one-episode-out} (LOEO) calibration: whenever a stored decision acts as a query, all entries from its episode are excluded from the reference library. This simulates deployment, where the current episode has not been cached. At runtime, a decision first checks whether a scheduled policy refresh is due. If not, the system retrieves and fuses a cached chunk, applies the action-level corrector, and checks recent task progress before execution. Retrieval and correction run on the CPU; on a cache-served decision, the vision encoder is the only component of the original policy that runs.

\subsection{Action-Chunk Cache Assembly}\label{sec:reuse}

\name stores individual \emph{policy decisions}, not entire trajectories to replay. Entry $i$ contains the task, a compact state key $z_i\in\mathbb{R}^{D}$, its policy-generated action chunk $a_i$, and its position within a successful episode (trajectory). Each key concatenates the robot's proprioceptive state with pooled features from the policy's own vision encoder, projected onto a PCA basis fitted from cached observations. A runtime observation already produces a query key $q$ through the same encoder and projection. Thus, a new chunk may combine decisions from multiple successful episodes rather than following one recorded trajectory.

% \paragraph{Compact keys separate task selection from state matching.}
% For each camera, we spatially pool the vision encoder's output tokens and project them with a PCA basis fitted on the library. Concatenating the camera codes with the robot state produces a compact key $q\in\mathbb{R}^{D}$. The instruction is fixed within a task and cannot tell its states apart, so it selects the candidate pool instead of entering the key. 
% We search the entries of the stored task whose instruction embedding has the highest cosine similarity to the current one.

\paragraph{Learning similarity from action agreement.}
Visual similarity alone is an unreliable criterion for action reuse. A large change in irrelevant background features may leave the appropriate motion unchanged, while a slight change in object pose may alter the required grasp or end-effector trajectory. Therefore, conventional similarity metrics such as Euclidean distance over visual features can select states whose actions do not fit the current decision (\S\ref{sec:ablation}).

Our key insight is that the cached policy decisions themselves provide supervision for learning which observation differences matter for control. Specifically, for each recorded decision, we find its most action-similar chunks in \emph{other episodes} of the same task, comparing actions at corresponding positions in the chunk. The cross-episode design avoids learning mainly from adjacent, nearly duplicate frames within one execution. 

Let $S$ be the mean outer product of the standardized key differences $z_i-z_j$ over these action-matched pairs $(i,j)$. We rank candidate entries by the regularized Mahalanobis distance~\cite{NIPS2002_c3e4035a,3041838.3041840} $d(q,z_i)=\|W^\top(q-z_i)\|_2$, where $W$ denotes a matrix factor of the regularized inverse covariance.  Specifically,
\begin{equation}
\label{eq:metric}
WW^\top=\left(S+\lambda\frac{\operatorname{tr}(S)}{D}I\right)^{-1}.
\end{equation}
where $D$ represents the query key $q$'s dimension. As such, directions that vary substantially even between action-matched states are less informative for selecting reusable actions, so the inverse covariance downweights them. 
The regularization term ($\lambda=0.1$ by default) limits excessive weighting of poorly sampled directions. 
% Moreover, directions that hardly vary at all, such as a background that never changes, are already removed by the PCA of the key. 

The metric is fitted entirely from existing policy executions, without modifying the policy or collecting additional supervision. At runtime, \name precomputes the projected keys $W^\top z_i$ for cached decisions, projects the incoming query once, and computes its distance to each entry of the selected task. This search scales linearly with the number of cached decisions and takes only about 3.6\,ms for a 5{,}000-episode library with 228{,}160 cache entries ({Table~\ref{tab:cpu}}).

\paragraph{Fusing action-compatible chunks.}
Even the closest entry rarely matches the current robot and object states exactly, making direct replay error-prone. To address this issue, \name interpolates among a local set of action chunks. Because these neighbors correspond to similar motions in the same action coordinates, their stepwise weighted average can reduce errors specific to a single recorded execution (e.g., outliers). We further calibrate the potential action deviation due to reusing the action chunk in Section~\ref{sec:correct}.

For the $k$ closest entries, let $a_i$ denote their
action chunks and $d_1\le\cdots\le d_k$ their distances.
Rather than using a fixed bandwidth, we normalize
distances by the gap between the closest and
middle-ranked neighbors, capturing the local spread
without relying on the most distant entries. Specifically, 
the assembled chunk $\bar a$ is
\begin{equation}
\label{eq:fusion}
\begin{aligned}
u_i &= \exp\!\left[-\left(
\frac{d_i-d_1}{d_{k/2}-d_1}
\right)^2\right],\quad
\bar a &= \frac{\sum_{i=1}^{k}u_i a_i}
{\sum_{i=1}^{k}u_i}.
\end{aligned}
\end{equation}
The closest entry receives weight one, while the
$(k/2)$th receives $e^{-1}$, making the weighting
adaptive to local cache density. We use $k=16$ by
default and show its consistent effectiveness (Section~\ref{sec:ablation}). The resulting action candidate
$\bar a$ is then passed to the residual correction stage
(Section~\ref{sec:correction}).

\subsection{Residual Cache Correction and Recovery}\label{sec:correct}
\label{sec:correction}\label{sec:when}

Action-aware assembly produces a plausible action chunk, but it may still deviate from what the original policy model would generate. Moreover, even a good instantaneous match cannot guarantee that a sequence of cached chunks advances the task, as deviations accumulate across multiple reused decisions. \name corrects errors at two timescales: within a decision, the corrector adjusts each assembled chunk without the policy, and across decisions, the stagnation check and periodic refresh return control to the policy.

\paragraph{Correcting cache-induced action errors.} To reduce the deviation between cached chunks and
policy-generated actions, \name employs a lightweight
residual corrector. Our key insight is that recorded
policy decisions already provide supervision for this
correction. For each decision, we reconstruct a cache
candidate $\bar a_i$ using entries from other episodes
and compare it with the original policy-generated
chunk $a_i$. Their difference $g_i=a_i-\bar a_i$
captures the residual error introduced by cache reuse.
This yields correction targets without additional
policy inference.

We fit a residual model $M$ \emph{offline} to predict these
errors from the current state key, assembled action
chunk, and decision index. To capture nonlinear
relationships while keeping prediction inexpensive,
we transform these inputs with a Gaussian kernel \cite{NIPS2007_013a006f} and then feed the resulting input into a
linear-regression model. Residuals are standardized
by motion channel, and training examples are weighted
so that each episode contributes equally regardless
of its length. 

At runtime, $M$ predicts an adjustment to the
assembled chunk, which we convert back to action
units before applying it. Since predictions become
less reliable for states poorly represented in the
cache, we attenuate corrections as the distance to
the nearest cached state increases. Thus,
\name corrects systematic cache errors in familiar
states without making large adjustments in unfamiliar
ones. The entire correction requires only lightweight
feature computation and linear prediction, without
invoking the policy backbone or action head.

\paragraph{Controlling execution drift.}
Even after correcting individual chunks, small errors can
accumulate across cached decisions and eventually stall
the task. For example, a robot may repeatedly approach
an object without completing a grasp. Such failures
cannot be detected reliably from robot displacement,
since substantial motion need not advance the task.
\name combines progress-based stall detection with
periodic policy refresh to limit errors across decisions.

To detect stalls, we use successful cached episodes as
references for task progress. Inspired by demonstration
alignment \cite{zhang2026regilretrievalguidedimitationlearning,huey2025imitationlearningsingletemporally},
we align a short history of observation keys
$q_0,\ldots,q_m$ with the keys
$z_1^e,\ldots,z_{L_e}^e$ recorded in execution order
during each successful episode $e$. Using the
action-aware distance (\S\ref{sec:reuse}), we find
the minimum-cost monotone alignment:
\begin{equation}
\label{eq:progress-alignment}
j_{0:m}^e =
\arg\min_{1\le j_0\le\cdots\le j_m\le L_e}
\sum_{k=0}^{m} d(q_k,z_{j_k}^e).
\end{equation}
Here $j_k$ is the reference decision matched to
observation $q_k$. The monotone constraint preserves
execution order while allowing repeated or skipped
positions to accommodate different execution speeds.
For each reference episode, we estimate progress from
the advance $j_m^e-j_0^e$, normalized by episode
length and the control steps spanned. 

To distinguish stalls from naturally slow progress,
we calibrate a task-specific threshold $\delta_p$
as the $p$th percentile of the progress estimates attained above. 
At runtime, \name invokes the original
policy when $\hat\delta_t<\delta_p$; otherwise,
cached execution continues. We choose the threshold so that 90\% of
successful calibration windows exceed it (the 10th
percentile; Appendix~\ref{app:stagnation} gives the full rule), and show \name's consistent effectiveness across $\delta_p$ settings in
Appendix~\ref{app:params}.

Progress monitoring is reactive: errors may accumulate
before a stall becomes apparent. \name
also invokes the original policy periodically, after
approximately $\tau$ decisions since the last policy
call; a stall-triggered call resets this interval.
Since tasks differ in their tolerance for cached
execution, we adapt $\tau$ to each task.
We construct a ladder of target cache hit rates,
from pure inference to aggressive reuse, and estimate
the corresponding refresh intervals through library
replay. During deployment, each task starts from pure inference,
progressively increases cache use, and retains the
last setting whose measured success rate remains
within a specified tolerance of the pure-inference
baseline. This allocates policy computation according
to task difficulty while allowing more repeatable
tasks to benefit from greater cache reuse. Moreover, we also evaluated the threshold-based refresh design alternative and found that it underperforms our periodic design (\S\ref{sec:ablation}).

\section{Implementation}\label{sec:impl}

\paragraph{Policy-server integration.}
We implement \name as a reusable serving layer atop the
openpi policy server \cite{openpi2025}, supporting
PyTorch policy backends without modifying policy models. Integrating a new
policy only requires three lightweight hooks that expose its vision
encoder, conditioning prefix, and action-head sampler.
These hooks allow \name to obtain observation features of policy inference, while retaining
the original inference path for policy refreshes.
The same cache implementation already supports five mainstream policies (\S\ref{sec:eval}).

\paragraph{CPU--GPU execution.}
The GPU runs the policy's vision
encoder to extract the current observation features. Asynchronously on CPUs, 
\name performs cache selection, action fusion, residual
correction, and progress monitoring. We precompute
the keys for all cache entries, so
retrieval requires only projecting the query and
scanning the selected task's entries. If the decision
is served from the cache, the corrected action chunk
is returned directly; otherwise, the request follows
the original GPU inference path. 

% \fan{rewrite this part. e.g., it should have something like "xxx framework/library, atop xxx platform. xxx backends". }
% \ziyang{Reply to Fan: rewritten in this shape, atop the openpi policy server, with PyTorch policy backends.}
% We implemented \name atop the openpi policy server \todo{cite openpi (Physical Intelligence, github.com/Physical-Intelligence/openpi)}, and its backends serve PyTorch policies without modifying them.
% Any policy with its own vision encoder can use \name by exposing three hooks, for vision encoding, the conditioning prefix, and action-head sampling, and one cache implementation serves all five policies (Appendix~\ref{app:impl}).
% Retrieval, fusion, correction, and the stagnation check run on the CPU, so a cache hit adds no GPU work beyond the vision encoder.
% Even with a 500-episode library (22{,}816 entries), they take under 2~ms per decision on one CPU core (Table~\ref{tab:cpu}), a negligible serving overhead that the serving experiment in Section~\ref{sec:e2e} includes.
% We will release the code, the cache libraries, and our trained policy checkpoints on Hugging Face upon publication.

\begin{table}[t]
\centering
\footnotesize
\begin{tabular*}{\columnwidth}{@{\extracolsep{\fill}}lcccc@{}}
\toprule
& \multicolumn{4}{c}{Library (episodes)} \\
\cmidrule(l){2-5}
Stage & 50 & 200 & 500 & 5{,}000 \\
\midrule
Retrieval & 0.62 & 0.63 & 0.71 & 1.55 \\
Fusion & 0.13 & 0.13 & 0.14 & 0.15 \\
Correction & 0.36 & 0.36 & 0.38 & 0.50 \\
Stagnation check & 0.25 & 0.31 & 0.39 & 1.39 \\
\midrule
Total & 1.36 & 1.43 & 1.61 & 3.58 \\
\bottomrule
\end{tabular*}
\vspace{-8.0pt}
\caption{\name adds under 2~ms of CPU time per decision up to 500 episodes and 3.6~ms at 5{,}000. Time (ms) of each part for $\pi_{0.5}$ on LIBERO-Long, on one pinned core of an idle Xeon E5-2696 v4, averaged over 2{,}048 decisions.}
\label{tab:cpu}
\end{table}

% The library and its learned quantities, such as the metric $W$ and the corrector, are built offline from recorded successful episodes, holding out each episode when fitting; the remaining constants are fixed defaults, the refresh level of each task is chosen during deployment, and Appendix~\ref{app:params} lists each with its role, value, and sensitivity.

% Together, action-aware retrieval makes backbone-free decisions possible; two-level correction repairs the cache estimate at the current decision and escalates to the policy when the resulting behavior stalls; and task-adaptive budgeting sets how much additional policy inference to spend. The entire design uses the original policy unchanged.

\section{Evaluation}\label{sec:eval}

We evaluate \name across five robot policies,
three simulation benchmarks, concurrent multi-robot
serving, and physical robot execution.
Our key findings are:

\begin{denseitemize}
\item \name reduces amortized
inference time by 1.4--11$\times$ (3.3$\times$ on average), improves serving throughput by 2.2$\times$ while preserving task
success rates. On a real SO-101 arm, it reduces decision latency by 3.1$\times$ (\S\ref{sec:e2e}).

\item  \name achieves new efficiency--success-rate
frontiers, with each design component contributing
measurable improvements (\S\ref{sec:perf-breakdown}).

\item \name consistently outperforms existing advances across diverse settings and design alternatives (\S\ref{sec:ablation}).
\end{denseitemize}

\subsection{Experimental Setup}\label{sec:setup}

\paragraph{Benchmarks and workloads.}
We evaluate \name on recurring manipulation tasks in which
the workcell and task remain fixed while initial object
configurations vary. We also perform out-of-distribution (OOD) experiments on LIBERO-plus \cite{fei2025liberoplusindepthrobustnessanalysis}, which perturbs LIBERO scenes along seven dimensions (e.g., lighting, camera viewpoints, and object placements), demonstrating effectiveness and transferability across environments (\S\ref{sec:ablation}).
We use LIBERO \cite{liu2023liberobenchmarkingknowledgetransfer} and MetaWorld MT50
\cite{yu2021metaworldbenchmarkevaluationmultitask}, covering three levels of difficulty:
LIBERO-Spatial consists of short pick-and-place tasks;
LIBERO-Long contains ten multi-stage tasks in which
execution errors can accumulate; and MT50 spans 50
diverse manipulation skills, with pure-inference success
rates of only 56--76\% across the evaluated policies. 
We also include real-robot evaluations (\S\ref{sec:e2e}). 

Cache entries are collected from successful policy
executions with initial states disjoint from evaluation, thus not repeating the same object configurations and placements.
We evaluate all 500 official LIBERO test episodes per
suite and 2{,}500 MT50 test episodes, using three
independently collected libraries per policy.
Library sizes range from 50 to 500 episodes per LIBERO
suite and 250 to 2{,}500 across MT50's 50 tasks,
distributed evenly among tasks.
All methods use the same evaluation initial states.

\begin{table}[t]
\centering
\footnotesize
\setlength{\tabcolsep}{2.8pt}
\begin{tabular}{@{}lcccc@{}}
\toprule
& \multicolumn{2}{c}{LIBERO} & \multicolumn{2}{c}{MT50} \\
\cmidrule(lr){2-3}\cmidrule(l){4-5}
Policy & Steps & Chunk & Steps & Chunk \\
\midrule
$\pi_{0.5}$ & $10\!\to\!1$ & $10/5\!\to\!10$ & $10\!\to\!2$ & $5/5\!\to\!5$ \\
GR00T N1.5 & $8\!\to\!1$ & $16/5\!\to\!10$ & $4\!\to\!2$ & $16/5\!\to\!10$ \\
Diffusion Policy & $100\!\to\!10$ & $16/5\!\to\!10$ & $100\!\to\!10$ & $16/5\!\to\!10$ \\
Fast-WAM & $10\!\to\!1$ & $32/5\!\to\!10$ & $10\!\to\!3$ & $32/5\!\to\!10$ \\
Cosmos Policy & $5\!\to\!2$ & $16/16\!\to\!16$ & n/a & n/a \\
\bottomrule
\end{tabular}
\vspace{-8.0pt}
\caption{Policy settings. Steps gives the default and reduced numbers of denoising steps. Chunk gives chunk length / default execution horizon $\to$ expanded execution horizon.}
\label{tab:policies}
\end{table}

\paragraph{Policy models.}
We evaluate five state-of-the-art policies spanning
different model architectures: $\pi_{0.5}$
\cite{intelligence2025pi05visionlanguageactionmodelopenworld}, GR00T N1.5
\cite{nvidia2025gr00tn15}, Diffusion Policy
\cite{chi2024diffusionpolicyvisuomotorpolicy}, Fast-WAM
\cite{yuan2026fastwamworldactionmodels}, and Cosmos Policy
\cite{kim2026cosmospolicyfinetuningvideo}.
On LIBERO, we use released checkpoints except for
Diffusion Policy, which we fine-tune. On MT50,
we use a released $\pi_{0.5}$ checkpoint and train
GR00T N1.5, Diffusion Policy, and Fast-WAM.
\name does not modify or fine-tune any policy
during cache construction or serving.

\paragraph{Baselines.}
We compare \name against four baselines:
\begin{denseitemize}
\item \emph{Pure inference}, which invokes the original
policy at every decision with the default settings;
\item \emph{Reduced + expanded}, pure inference with
reduced denoising steps and expanded execution
horizons (Table~\ref{tab:policies});
\item \emph{Prior denoising-reuse advances}, whose speedups we already re-evaluated using our implementation
in Table~\ref{tab:speedup};
\item \emph{RT-Cache} \cite{kwon2025rtcachetrainingfreeretrievalrealtime},
a policy-bypassing action cache evaluated in
Table~\ref{tab:ladder}.
\end{denseitemize}
% Our primary baseline is pure inference, which
% invokes the original policy at every decision. We
% compare against both default inference and the
% reduced-step and expanded-horizon configurations
% in Table~\ref{tab:policies}. Plain step reduction
% provides a strong, architecture-agnostic no-cache
% baseline for denoising acceleration
% (Section~\ref{sec:motivation}), and with the expanded horizon it is at least as fast as the most optimistic estimate of every prior single-call acceleration that we reproduce. Table~\ref{tab:speedup}
% separately compares published acceleration methods
% on one optimized implementation.
% We additionally compare with RT-Cache
% \todo{cite Kwon 2025}, which can bypass policy
% inference, in the $\pi_{0.5}$ ablation
% (Section~\ref{sec:ablation}).

% \paragraph{Baselines.}
% We compare with pure inference, which calls the policy at every decision, under its default settings and with the reduced denoising steps and expanded execution horizon of Table~\ref{tab:policies}.
% Other training-free accelerations are built for one family of policies, and most shorten only the action head while the backbone still runs at every decision (Section~\ref{sec:related}, Table~\todo{}).
% Section~\ref{sec:motivation} shows that on our latency metric they are at best about as fast as reducing denoising steps (Table~\ref{tab:speedup}), which applies to every policy, so step reduction stands in for this class.
% We also compare with RT-Cache \todo{cite Kwon 2025}, the only prior cache that skips the policy, on $\pi_{0.5}$ (Section~\ref{sec:ablation}).

\paragraph{Metrics.}
We report task success rate, cache hit rate (the fraction
of decisions served without invoking the policy), and
amortized inference time. The latter is the total GPU
inference time divided by the number of decisions
required at the default execution horizon $h$.
This normalization enables fair comparison across
settings that execute different numbers of actions
per decision, capturing savings from both cheaper
and fewer policy calls. We compute it using
policy-stage latencies measured at batch size 1
on an RTX 4090. We separately measure throughput
and request latency under concurrent multi-robot
serving (\S\ref{sec:e2e}).

% \jay{Sec 5.1 Metrics: amortized inference time is computed from policy-stage GPU latencies, not measured wall-clock. Please state whether the CPU time of Table 3 is included or overlapped with GPU work, and report a measured end-to-end number for at least one policy. Alternative: report both the GPU-only and the CPU-inclusive number in Fig. 5 and Table 2.}

\begin{figure*}[t]
\centering
\includegraphics[width=\textwidth]{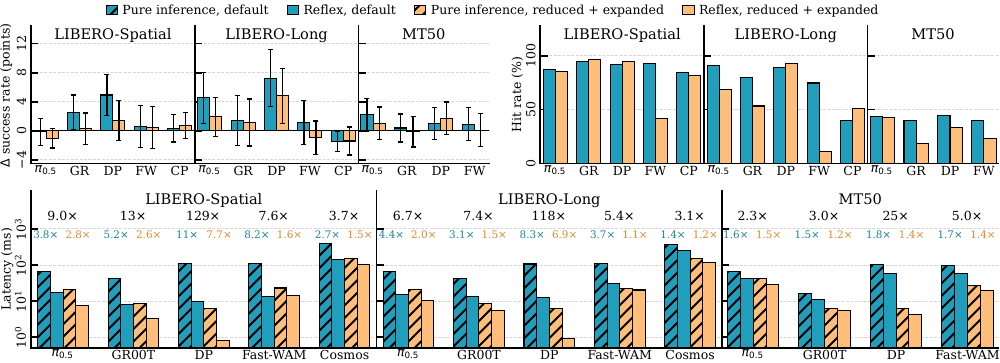}
\vspace{-.7cm}
\caption{\name reduces amortized inference time by
1.4--11$\times$ (3.3$\times$ on average) while
preserving task success, reaching 9.3$\times$ on
average when combined with reduced denoising steps
and expanded execution horizons (specified in Table~\ref{tab:policies}).
Each library contains 20 successful episodes per task
(200 on LIBERO and 1{,}000 on MT50). Results are
averaged over three independently collected libraries.
Success-rate changes are relative to pure inference
under the same settings, with error bars showing
95\% confidence intervals.
Colored annotations report speedups over pure inference
with matching settings; black annotations compare
the solid orange and hatched blue bars.
Cosmos Policy and $\pi_{0.5}$ on MT50 do not use
expanded execution horizons due to the high task complexity.}
\label{fig:main}
\end{figure*}

% We report task success rate, cache hit rate (the
% fraction of decisions served without the policy),
% and inference cost. To compare settings with
% different execution horizons, we normalize
% total GPU execution time by the number of decisions at the default execution horizon.
% These are measured at
% batch size 1 on an RTX 4090 ( Table~\ref{tab:speedup}). 
% We also evaluate concurrent multi-robot
% serving, measuring throughput and latency (\S\ref{sec:e2e}).

% \paragraph{Details and further results.}
% Appendices~\ref{app:library} and~\ref{app:training} describe how the libraries are collected and how the policies are trained, Appendices~\ref{app:latency} and~\ref{app:robot} describe the serving system and the robot, and Appendices~\ref{app:libsize}, \ref{app:knob}, \ref{app:threshold}, and~\ref{app:inside} report every library size, how the refresh level of each task is chosen, distance thresholds in place of periodic refresh, and one cached episode step by step.
\paragraph{Additional details and results.}
Appendices~\ref{app:stagnation}--\ref{app:robot} detail
the stagnation check, implementation, parameters, policy training,
cache construction, serving infrastructure, and real-robot setup.
Appendices~\ref{app:libsize}--\ref{app:ood} provide further analyses
of library size, task-adaptive refresh, a step-by-step example of cached execution,
and out-of-distribution scenes.

% \paragraph{Metrics.}
% We report the success rate, the latency, and the hit rate, which is the fraction of decisions served by the cache.
% Latency is the mean time per decision when one GPU serves one robot, the most common setup today \todo{cite}.
% Over $T$ executed control steps with $V$ vision encoder passes and $M$ policy calls, we compute it from measured stage latencies as
% \begin{equation}
% L = \frac{t_v V + t_m M}{T / h},
% \label{eq:latency}
% \end{equation}
% where $t_v$ and $t_m$ are the latencies of a vision encoder pass and of the rest of a policy call, measured at batch size 1 with compilation and CUDA graphs on an RTX 4090.
% The denominator is the number of decisions pure inference makes with the default execution horizon $h$, and pure inference has a latency of $t_v + t_m$.
% Normalizing by these decisions puts every setting on the same footing, so executing longer chunks lowers the latency in proportion to the decisions it saves.
% Section~\ref{sec:e2e} also serves many robots from one GPU and measures throughput and tail latency directly.

\subsection{End-to-End Results}\label{sec:e2e}

\paragraph{\name reduces inference time while preserving task success.}
Figure~\ref{fig:main} compares \name against pure inference
across five policies and three benchmarks. Under default
policy settings, \name reduces amortized inference time
by 1.4--11$\times$, averaging 3.3$\times$.
Task success remains comparable: no setting loses more
than 2 percentage points relative to pure inference,
while Diffusion Policy on LIBERO-Long gains 7.3 points.
These results show that reusing complete policy decisions
can substantially reduce inference cost without
compromising task reliability.

\begin{figure}[t]
\centering
\includegraphics[width=\columnwidth]{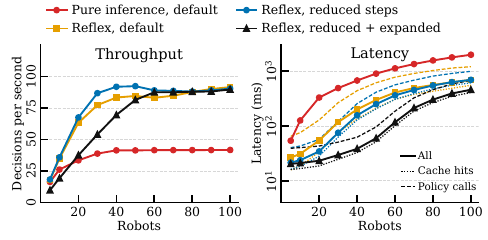}
\vspace{-26.3pt}
\caption{\name improves serving throughput (decisions per second) by about 2$\times$ over pure inference on one RTX 4090. Dashed lines split the latency into cache hits and policy calls.}
\label{fig:cluster}
% \vspace{-.2cm}
\end{figure}

\paragraph{\name improves serving throughput.}
We replay 500 LIBERO-Long episodes as requests from
up to 100 robots sharing a $\pi_{0.5}$ server on
one RTX 4090 (Figure~\ref{fig:cluster}).
Pure inference saturates at approximately
42.5 decisions/s, whereas \name sustains
87--97 decisions/s under default settings.
With at least 60 concurrent robots, \name completes
the same workload 2.0--2.2$\times$ faster;
reduced denoising steps and expanded execution
horizons increase this gain to 4.2--4.3$\times$.
At 20 robots, the p99 request latency drops
from 511~ms to 225~ms, and further to 61~ms
with both complementary optimizations.
These results confirm that backbone-free cache
reuse translates into higher GPU serving capacity
and lower tail latency under concurrent workloads.

% We wrote a concurrent server for $\pi_{0.5}$ and run it with and without \name on one RTX 4090 (Appendix~\todo{}).
% A replayer acts as up to \todo{100} robots that send the recorded requests of the \todo{500} LIBERO-Long test episodes, and each robot waits for its actions to execute before the next request.
% The server is not heavily tuned, so Figure~\ref{fig:cluster} gives a lower bound for a tuned deployment.
% Pure inference saturates the GPU at about \todo{40} robots and \todo{42.5} decisions per second.
% \name serves \todo{87} to \todo{97} decisions per second at the default settings, about twice as many, because cache hits skip the backbone.
% With \todo{60} or more robots, it finishes the \todo{500} episodes \todo{2.0$\times$} to \todo{2.2$\times$} faster, and \todo{4.2$\times$} to \todo{4.3$\times$} faster with reduced steps and the expanded horizon.
% Reduced steps alone add almost nothing to the default cache in Figure~\ref{fig:cluster}, because the action head takes only \todo{15\%} of a batched policy call (Table~\ref{tab:stages}), while the expanded horizon halves the decisions.
% At \todo{20} robots, \name lowers the \todo{99th} percentile latency from \todo{511}~ms to \todo{225}~ms, and to \todo{61}~ms with reduced steps and the expanded horizon.

\paragraph{\name improves real-robot performance.}
We run $\pi_{0.5}$ on an SO-101 arm for two tasks that place a block and a bottle into a frame in opposite orders.
Over 200 episodes, \name succeeds in 172 and pure inference in 174 (Table~\ref{tab:robot}).
\name serves 79.3\% of decisions from the cache and lowers the latency per decision from 66.5~ms to 21.8~ms, a 3.1$\times$ speedup.
This latency is measured on the robot, with one RTX 4090 serving it at batch size 1.

\begin{figure}[t]
\centering
\includegraphics[width=\columnwidth]{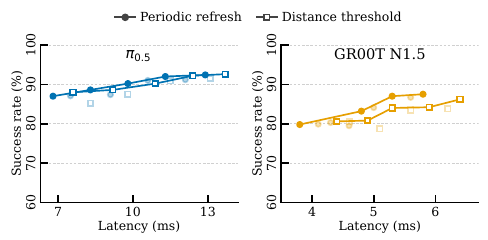}
\vspace{-25.1pt}
\caption{Periodic refresh achieves a better
latency--success-rate tradeoff than distance-based
refresh (LIBERO-Long).}
% \vspace{-.2cm}
\label{fig:threshold}
\end{figure}

\begin{table}[t]
\centering
\footnotesize
\begin{tabular}{@{}lccc@{}}
\toprule
Setting & Successes & Hit rate (\%) & Latency (ms) \\
\midrule
Pure inference & 174/200 & n/a & 66.5 \\
\name & 172/200 & 79.3 & 21.8 \\
\bottomrule
\end{tabular}
\vspace{-3.0pt}
\caption{On an SO-101 arm with $\pi_{0.5}$, \name lowers the latency per decision by 3.1$\times$ with nearly the same success rate.}
\label{tab:robot}
\end{table}

\paragraph{\name compounds gains from cheaper and fewer policy calls.}
Reducing denoising steps makes each policy call cheaper,
while expanding the execution horizon reduces the number
of decisions. Both optimizations complement cache reuse.
Even with these optimizations enabled, \name remains
2.0$\times$ faster on average than pure inference under
the same settings. On LIBERO-Long and MT50, the cache
hit rate decreases in 7 of 9 cases
(Figure~\ref{fig:main}, top right), as task-adaptive
refresh allocates more policy calls to preserve success.
Relative to pure inference at default settings,
combining all optimizations yields speedups of
2.3--129$\times$ (9.3$\times$ on average).
These gains compound because \name bypasses the
backbone on cache hits, beyond the savings from
cheaper and fewer policy calls.

\paragraph{\name adds little serving overhead.}
Table~\ref{tab:cpu} breaks down the overhead of cache
retrieval, fusion, correction, and stagnation detection.
The total overhead remains below
2~ms per decision for libraries of up to 500 episodes
and reaches only 3.6~ms at 5{,}000 episodes.
Retrieval and stagnation detection account for most
of the increase with library size, while fusion and
correction remain nearly constant.

\subsection{Performance Breakdown}
\label{sec:perf-breakdown}

\begin{table}[t]
\centering
\footnotesize
\setlength{\tabcolsep}{3.0pt}
\begin{tabular}{@{}lcc@{}}
\toprule
Setting & $\pi_{0.5}$ & Diffusion Policy \\
\midrule
RT-Cache & 25.6/100/17.9 & 25.0/100/18.8 \\
\midrule
Cache, Euclidean, nearest entry & 55.8/100/5.1 & 52.0/100/0.5 \\
+ Mahalanobis distance & 62.8/100/5.2 & 60.2/100/0.5 \\
\quad with CLIP keys & 40.8/100/0.8 & 39.6/100/0.8 \\
+ Fusion of 16 entries & 77.0/100/5.2 & 76.8/100/0.5 \\
+ Corrector & 84.6/100/5.2 & 78.6/100/0.5 \\
+ Stagnation check & 86.2/94/6.1 & 85.8/91/1.0 \\
+ Periodic refresh (full system) & 91.2/65/10.6 & 88.0/92/1.0 \\
\bottomrule
\end{tabular}
\vspace{-7.7pt}
\caption{Each component of \name raises the success rate on LIBERO-Long, with reduced steps, and the expanded horizon. Cells report success rate (\%) / hit rate (\%) / latency (ms).}
\vspace{-.4cm}
\label{tab:ladder}
\end{table}

\paragraph{Performance breakdown by design components.}
Table~\ref{tab:ladder} isolates the contributions of
\name's components. Without policy calls, action-aware
retrieval, chunk fusion, and residual correction raise
success from 55.8\% to 84.6\% for $\pi_{0.5}$ and from
52.0\% to 78.6\% for Diffusion Policy, with fusion
providing the largest individual gain.
Adding progress-based stall detection and periodic
refresh further improves success by 6.6 and 9.4
percentage points, respectively, reaching 91.2\%
and 88.0\%, compared with 91.8\% and 82.4\% for
pure inference under the same settings.
Task-adaptive refresh is also important: using a
single interval across tasks reduces success by
3.9 and 3.0 points for GR00T N1.5 and Cosmos Policy
on LIBERO-Long at comparable latency, and by
7.4--12.1 points on MT50 even with the best
shared interval.

\paragraph{Performance breakdown by inference stages.}
A cache hit bypasses the backbone and action head but
still runs the vision encoder. Consequently, achievable
speedup depends on both the fraction of decisions served
from the cache and the encoder's share of inference time.
For Diffusion Policy, vision encoding takes only 1~ms
of a 106~ms decision, allowing an 11$\times$ speedup
at a 92\% hit rate on LIBERO-Spatial. In contrast,
$\pi_{0.5}$ spends about 10~ms of its 68~ms decision
on vision encoding, limiting the maximum cache-only
speedup to roughly 6.6$\times$. At a 91\% hit rate,
\name achieves 4.4$\times$ on LIBERO-Long.
For more error-sensitive tasks, such as MT50 and
Cosmos Policy on LIBERO-Long, task-adaptive refresh
trades some cache hits for higher success rates,
yielding smaller but reliable speedups.

\paragraph{Design alternatives of the cache key.}
We examine whether \name can replace the policy's own
vision features with general-purpose image embeddings.
Using CLIP features \cite{radford2021learningtransferablevisualmodels} while
keeping the retrieval method unchanged reduces success
from 62.8\% to 40.8\% for $\pi_{0.5}$ and from
60.2\% to 39.6\% for Diffusion Policy
(Table~\ref{tab:ladder}).
We also compare against RT-Cache \cite{kwon2025rtcachetrainingfreeretrievalrealtime},
a policy-bypassing cache that retrieves actions using
general-purpose DINOv2 and SigLIP features.
It achieves only 25.6\% and 25.0\% success,
respectively, on the same benchmark.
These results highlight the importance of representation
choice: general-purpose visual features capture semantic
similarity, but may overlook subtle state differences
that determine the appropriate action.
Using the policy's own vision features better aligns
retrieval with the observations on which its actions
depend.

\subsection{Ablation Study}\label{sec:ablation}

\begin{figure}[t]
\centering
\includegraphics[width=\columnwidth]{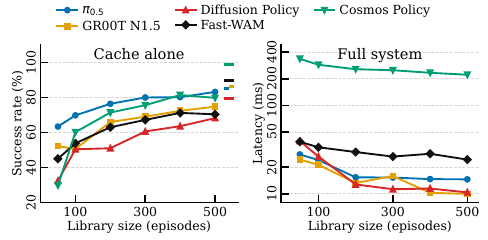}
\vspace{-.4cm}
\caption{Cache alone, \name without the stagnation check and periodic refresh (100\% hit rate), gains with library size but stays below pure inference (short dashes, left), while in the full system a larger library lowers the latency (right), on LIBERO-Long.}
\label{fig:libsize}
% \vspace{-.2cm}
\end{figure}

\paragraph{Moderate cache libraries capture most reuse benefits.}
Figure~\ref{fig:libsize} reports the impact of library
size on \name. Without policy calls, increasing the
library from 50 to 500 episodes improves success by
20--50 percentage points on LIBERO-Long, while
increasing it from 250 to 2{,}500 episodes yields
12--21-point gains on MT50. Improvements begin to
level off near 500 episodes on LIBERO-Long, suggesting
that \name does not require extensive experience
collection. Nevertheless, cache-only success remains
below pure inference even with the largest libraries
(75.3\% versus 87.8\% on average), motivating
policy feedback. In the full system (\name), progress-based
stall detection and task-adaptive refresh preserve
success while larger libraries enable more cache hits.
From 50 to 500 episodes, amortized inference time
decreases by 1.5--3.8$\times$, achieving
1.7--10.2$\times$ speedups over pure inference
(Figure~\ref{fig:libsize}, right).

\paragraph{Periodic Refresh vs. Distance Threshold-based Refresh.}
\name periodically invokes the policy to limit execution
drift. An alternative is distance-based refresh, which
invokes the policy when the nearest cached state exceeds
a distance threshold. We compare both strategies across
refresh settings on LIBERO-Long (Figure~\ref{fig:threshold}).
Periodic refresh consistently achieves a better
latency--success-rate tradeoff, requiring approximately
1.2$\times$ less inference time at comparable
success rates.

\paragraph{Efficient adaptation to new scenes.}
We evaluate \name under scene perturbations using
LIBERO-Plus \cite{fei2025liberoplusindepthrobustnessanalysis}.
An unchanged cache
achieves 62.6\% success, versus 79.5\% for pure
$\pi_{0.5}$ inference (Table~\ref{tab:ood-main}).
To adapt, we collect three policy episodes per new
scene, add their decisions to the library, 
and refit the action-aware metric, all within seconds
on a CPU. This raises success to 74.9\%, or 77.1\%
with stagnation detection, within 2.4 percentage
points of pure inference while requiring policy calls
for only 8.8\% of decisions. On LIBERO-Spatial,
\name slightly exceeds pure inference (89.5\% vs.\
88.6\%). The learned metric is critical: replacing
it with Euclidean distance reduces adapted success
from 74.9\% to 66.8\%.

% A library built in the original scenes does not transfer unchanged, and on LIBERO-plus \todo{cite Fei 2025 (LIBERO-Plus)} the cache succeeds in 62.6\% against 79.5\% for $\pi_{0.5}$ (Table~\ref{tab:ood-main}).
% We adapt the cache to each new scene with three episodes of $\pi_{0.5}$, a one-time cost that pays off when the scene is served for many episodes.
% Their decisions join the library, a ridge map aligns the third-person features of the new scene with those of the library when that view has shifted more than the wrist view, and the metric of Equation~\ref{eq:metric} is refit, all in seconds on a CPU.
% With the stagnation check, \name then comes within 2.5 points of $\pi_{0.5}$ while calling it for 8.8\% of the decisions, and on LIBERO-Spatial it succeeds more often than $\pi_{0.5}$.
% The learned metric is what makes adaptation work, since with the Euclidean distance adaptation reaches only 66.8\% instead of 74.9\%.

% In Figure~\ref{fig:libsize}, only the cache and the corrector run.
% Growing the library from \todo{50} to \todo{500} episodes raises the success rate by \todo{20} to \todo{50} points on LIBERO-Long, and from \todo{250} to \todo{2{,}500} episodes by \todo{12} to \todo{21} points on MT50, but every curve stays below pure inference.
% \name therefore keeps execution-level correction.

\begin{table}[t]
\centering
\footnotesize
\setlength{\tabcolsep}{1.6pt}
\begin{tabular}{@{}lccccc@{}}
\toprule
LIBERO-Plus & $\pi_{0.5}$ & Unchanged & Adapted & + Stagnation & Calls (\%) \\
\midrule
Spatial & 88.6 & 74.5 & 88.9 & 89.5 & 6.6 \\
Long & 71.3 & 51.7 & 62.1 & 65.7 & 9.5 \\
Both & 79.5 & 62.6 & 74.9 & 77.1 & 8.8 \\
\bottomrule
\end{tabular}
\vspace{-8.0pt}
\caption{Three policy episodes per new scene recover
most of the success rate on LIBERO-Plus.
Values are success rates (\%); Calls denotes the
fraction of decisions invoking the policy with
stagnation detection.}
\label{tab:ood-main}
\vspace{-.2cm}
\end{table}

\section{Related Work}
\label{sec:related}

\paragraph{Accelerating robot policies.}
Existing approaches reduce the computation required by
robot policies through denoising-state reuse
\cite{oi2026actioncachetrainingfreeaccelerationvisionlanguageaction,chen2025falconfastvisuomotorpolicies,zhao2026c3acheacceleratingworldaction},
token and feature reuse or pruning
\cite{xu2025vlacacheefficientvisionlanguageactionmanipulation,yang2025efficientvlatrainingfreeaccelerationcompression},
and computation reuse across repeated task executions
\cite{zhou2026rmuscleroboticmusclememory,zheng2025leveragingoslevelprimitivesrobotic}.
Distillation trains policies that require fewer
denoising steps
\cite{prasad2024consistencypolicyacceleratedvisuomotor,wang2024onestepdiffusionpolicyfast}.
These approaches primarily reduce the cost of policy
inference, whereas \name avoids both backbone and
action-head computation on cache hits. The two are
complementary.
% : per-call accelerations can further
% reduce the cost of \name's remaining policy invocations.

\paragraph{Retrieval for robot control.}
Retrieval-based policies select actions from demonstrations
using pretrained visual representations
\cite{pari2021surprisingeffectivenessrepresentationlearning,kwon2025rtcachetrainingfreeretrievalrealtime}.
Other approaches retrieve past experiences to guide or
correct a frozen policy during inference
\cite{zhao2026retrievethensteeronlinesuccessmemory}.
RT-Cache also bypasses policy inference, but retrieves
actions using general-purpose visual features without
periodic policy feedback. In contrast, \name retrieves
the original policy's decisions using its own visual
features, learns similarity from action agreement, and
corrects cached actions while monitoring execution
progress. 

\paragraph{Caching in ML serving.}
LLM serving systems cache attention states to avoid
recomputing shared prefixes
\cite{kwon2023efficientmemorymanagementlarge,zheng2024sglangefficientexecutionstructured,gim2024promptcachemodularattention,Yu_2025}.
Approximate caches extend reuse to similar requests,
including semantic response caching
\cite{Fu2023GPTCacheAO} and denoising-state
reuse in image generation \cite{agarwal2023approximatecachingefficientlyserving}.
Robot control poses an additional challenge: cached
actions alter the physical state and thus affect
subsequent observations and decisions (\S\ref{sec:ablation}). 

\paragraph{Serving robot policies.}
Recent systems improve shared robot-policy serving
through batching, request scheduling, action-chunk
management, and multi-model GPU allocation
\cite{jiang2026rosaroboticsfoundationmodel,kairos-arxiv26,bansal2026actionchunkschedulingbatched,adamopoulos2026efficientvisionlanguageactionmanagementserving}.
These systems optimize the execution of policy requests.
\name instead reduces the number of requests requiring
expensive backbone inference. 

\section{Conclusion}
\label{sec:conclusion}

We presented \name, a training-free cache that exploits recurring states across robot task executions.
By reusing complete action chunks from successful
executions, \name bypasses expensive policy inference
while addressing action mismatch and execution drift
through action-aware retrieval, residual correction,
and task-adaptive policy refresh. Across five policies
and three benchmarks, \name reduces inference
time by 3.3$\times$ while preserving task
success.

\bibliography{example_paper}

@misc{kairos-arxiv26,
      title={{Kairos: A Scalable Serving System for Physical AI}}, 
      author={Yinwei Dai and Ganesh Ananthanarayanan and Landon Cox and Xenofon Foukas and Bozidar Radunovic and Ravi Netravali},
      year={2026},
      eprint={2605.11381},
      archivePrefix={arXiv},
      primaryClass={cs.RO},
      url={https://arxiv.org/abs/2605.11381}, 
}

@inproceedings{chi2024diffusionpolicyvisuomotorpolicy,
  title     = {{{Diffusion Policy}: Visuomotor Policy Learning via Action Diffusion}},
  author    = {Cheng Chi and Siyuan Feng and Yilun Du and Zhenjia Xu and Eric Cousineau and Benjamin Burchfiel and Shuran Song},
  booktitle = {Proceedings of Robotics: Science and Systems (RSS)},
  address   = {Daegu, Republic of Korea},
  month     = jul,
  year      = {2023},
  doi       = {10.15607/RSS.2023.XIX.026},
}

@misc{black2026pi0visionlanguageactionflowmodel,
      title={{$\pi_0$: A Vision-Language-Action Flow Model for General Robot Control}}, 
      author={Kevin Black and Noah Brown and Danny Driess and Adnan Esmail and Michael Equi and Chelsea Finn and Niccolo Fusai and Lachy Groom and Karol Hausman and Brian Ichter and Szymon Jakubczak and Tim Jones and Liyiming Ke and Sergey Levine and Adrian Li-Bell and Mohith Mothukuri and Suraj Nair and Karl Pertsch and Lucy Xiaoyang Shi and James Tanner and Quan Vuong and Anna Walling and Haohuan Wang and Ury Zhilinsky},
      year={2024},
      eprint={2410.24164},
      archivePrefix={arXiv},
      primaryClass={cs.LG},
      url={https://arxiv.org/abs/2410.24164}, 
}

@inproceedings{intelligence2025pi05visionlanguageactionmodelopenworld,
  title     = {{$\pi_{0.5}$: a Vision-Language-Action Model with Open-World Generalization}},
  author    = {Kevin Black and Noah Brown and James Darpinian and Karan Dhabalia and Danny Driess and Adnan Esmail and Michael Robert Equi and Chelsea Finn and Niccolo Fusai and Manuel Y. Galliker and Dibya Ghosh and Lachy Groom and Karol Hausman and Brian Ichter and Szymon Jakubczak and Tim Jones and Liyiming Ke and Devin LeBlanc and Sergey Levine and Adrian Li-Bell and Mohith Mothukuri and Suraj Nair and Karl Pertsch and Allen Z. Ren and Lucy Xiaoyang Shi and Laura Smith and Jost Tobias Springenberg and Kyle Stachowicz and James Tanner and Quan Vuong and Homer Walke and Anna Walling and Haohuan Wang and Lili Yu and Ury Zhilinsky},
  booktitle = {Proceedings of The 9th Conference on Robot Learning},
  series    = {Proceedings of Machine Learning Research},
  volume    = {305},
  pages     = {17--40},
  publisher = {PMLR},
  year      = {2025},
}

@inproceedings{kim2024openvlaopensourcevisionlanguageactionmodel,
  title     = {{{OpenVLA}: An Open-Source Vision-Language-Action Model}},
  author    = {Moo Jin Kim and Karl Pertsch and Siddharth Karamcheti and Ted Xiao and Ashwin Balakrishna and Suraj Nair and Rafael Rafailov and Ethan P. Foster and Pannag R. Sanketi and Quan Vuong and Thomas Kollar and Benjamin Burchfiel and Russ Tedrake and Dorsa Sadigh and Sergey Levine and Percy Liang and Chelsea Finn},
  booktitle = {Proceedings of The 8th Conference on Robot Learning},
  series    = {Proceedings of Machine Learning Research},
  volume    = {270},
  pages     = {2679--2713},
  publisher = {PMLR},
  year      = {2025},
}

@misc{nvidia2025gr00tn1openfoundation,
      title={{GR00T N1: An Open Foundation Model for Generalist Humanoid Robots}}, 
      author={{NVIDIA} and Johan Bjorck and Fernando Castañeda and Nikita Cherniadev and Xingye Da and Runyu Ding and Linxi "Jim" Fan and Yu Fang and Dieter Fox and Fengyuan Hu and Spencer Huang and Joel Jang and Zhenyu Jiang and Jan Kautz and Kaushil Kundalia and Lawrence Lao and Zhiqi Li and Zongyu Lin and Kevin Lin and Guilin Liu and Edith Llontop and Loic Magne and Ajay Mandlekar and Avnish Narayan and Soroush Nasiriany and Scott Reed and You Liang Tan and Guanzhi Wang and Zu Wang and Jing Wang and Qi Wang and Jiannan Xiang and Yuqi Xie and Yinzhen Xu and Zhenjia Xu and Seonghyeon Ye and Zhiding Yu and Ao Zhang and Hao Zhang and Yizhou Zhao and Ruijie Zheng and Yuke Zhu},
      year={2025},
      eprint={2503.14734},
      archivePrefix={arXiv},
      primaryClass={cs.RO},
      url={https://arxiv.org/abs/2503.14734}, 
}

@misc{nvidia2025gr00tn15,
      title={{{GR00T N1.5}: An Improved Open Foundation Model for Generalist Humanoid Robots}},
      author={{NVIDIA}},
      year={2025},
      howpublished={\url{https://research.nvidia.com/labs/gear/gr00t-n1_5/}},
      note={Model weights: \url{https://huggingface.co/nvidia/GR00T-N1.5-3B}},
}

@misc{yuan2026fastwamworldactionmodels,
      title={{Fast-WAM: Do World Action Models Need Test-time Future Imagination?}}, 
      author={Tianyuan Yuan and Zibin Dong and Yicheng Liu and Hang Zhao},
      year={2026},
      eprint={2603.16666},
      archivePrefix={arXiv},
      primaryClass={cs.CV},
      url={https://arxiv.org/abs/2603.16666}, 
}

@inproceedings{kim2026cosmospolicyfinetuningvideo,
  title     = {{{Cosmos Policy}: Fine-Tuning Video Models for Visuomotor Control and Planning}},
  author    = {Moo Jin Kim and Yihuai Gao and Tsung-Yi Lin and Yen-Chen Lin and Yunhao Ge and Grace Lam and Percy Liang and Shuran Song and Ming-Yu Liu and Chelsea Finn and Jinwei Gu},
  booktitle = {International Conference on Learning Representations (ICLR)},
  year      = {2026},
  url       = {https://openreview.net/forum?id=wPEIStHxYH},
}

@misc{guo2026unified4dworldaction,
      title={{Unified 4D World Action Modeling from Video Priors with Asynchronous Denoising}}, 
      author={Jun Guo and Qiwei Li and Peiyan Li and Zilong Chen and Nan Sun and Yifei Su and Heyun Wang and Yuan Zhang and Xinghang Li and Huaping Liu},
      year={2026},
      eprint={2604.26694},
      archivePrefix={arXiv},
      primaryClass={cs.RO},
      url={https://arxiv.org/abs/2604.26694}, 
}

@misc{ye2026worldactionmodelszeroshot,
      title={{World Action Models are Zero-shot Policies}}, 
      author={Seonghyeon Ye and Yunhao Ge and Kaiyuan Zheng and Shenyuan Gao and Sihyun Yu and George Kurian and Suneel Indupuru and You Liang Tan and Chuning Zhu and Jiannan Xiang and Ayaan Malik and Kyungmin Lee and William Liang and Nadun Ranawaka and Jiasheng Gu and Yinzhen Xu and Guanzhi Wang and Fengyuan Hu and Avnish Narayan and Johan Bjorck and Jing Wang and Gwanghyun Kim and Dantong Niu and Ruijie Zheng and Yuqi Xie and Jimmy Wu and Qi Wang and Ryan Julian and Danfei Xu and Yilun Du and Yevgen Chebotar and Scott Reed and Jan Kautz and Yuke Zhu and Linxi "Jim" Fan and Joel Jang},
      year={2026},
      eprint={2602.15922},
      archivePrefix={arXiv},
      primaryClass={cs.RO},
      url={https://arxiv.org/abs/2602.15922}, 
}

@misc{geminiroboticsteam2025geminiroboticsbringingai,
      title={{Gemini Robotics: Bringing AI into the Physical World}}, 
      author={{Gemini Robotics Team}},
      year={2025},
      eprint={2503.20020},
      archivePrefix={arXiv},
      primaryClass={cs.RO},
      url={https://arxiv.org/abs/2503.20020}, 
}

@inproceedings{radford2021learningtransferablevisualmodels,
  title     = {{Learning Transferable Visual Models From Natural Language Supervision}},
  author    = {Alec Radford and Jong Wook Kim and Chris Hallacy and Aditya Ramesh and Gabriel Goh and Sandhini Agarwal and Girish Sastry and Amanda Askell and Pamela Mishkin and Jack Clark and Gretchen Krueger and Ilya Sutskever},
  booktitle = {Proceedings of the 38th International Conference on Machine Learning},
  series    = {Proceedings of Machine Learning Research},
  volume    = {139},
  pages     = {8748--8763},
  publisher = {PMLR},
  year      = {2021},
}

@misc{shukor2025smolvlavisionlanguageactionmodelaffordable,
      title={{SmolVLA: A Vision-Language-Action Model for Affordable and Efficient Robotics}}, 
      author={Mustafa Shukor and Dana Aubakirova and Francesco Capuano and Pepijn Kooijmans and Steven Palma and Adil Zouitine and Michel Aractingi and Caroline Pascal and Martino Russi and Andres Marafioti and Simon Alibert and Matthieu Cord and Thomas Wolf and Remi Cadene},
      year={2025},
      eprint={2506.01844},
      archivePrefix={arXiv},
      primaryClass={cs.LG},
      url={https://arxiv.org/abs/2506.01844}, 
}

@inproceedings{liu2023liberobenchmarkingknowledgetransfer,
  title     = {{{LIBERO}: Benchmarking Knowledge Transfer for Lifelong Robot Learning}},
  author    = {Bo Liu and Yifeng Zhu and Chongkai Gao and Yihao Feng and Qiang Liu and Yuke Zhu and Peter Stone},
  booktitle = {Advances in Neural Information Processing Systems (NeurIPS), Datasets and Benchmarks Track},
  volume    = {36},
  pages     = {44776--44791},
  year      = {2023},
}

@inproceedings{yu2021metaworldbenchmarkevaluationmultitask,
  title     = {{{Meta-World}: A Benchmark and Evaluation for Multi-Task and Meta Reinforcement Learning}},
  author    = {Tianhe Yu and Deirdre Quillen and Zhanpeng He and Ryan Julian and Karol Hausman and Chelsea Finn and Sergey Levine},
  booktitle = {Proceedings of the Conference on Robot Learning},
  series    = {Proceedings of Machine Learning Research},
  volume    = {100},
  pages     = {1094--1100},
  publisher = {PMLR},
  year      = {2020},
}

@inproceedings{nasiriany2024robocasalargescalesimulationeveryday,
  title     = {{{RoboCasa}: Large-Scale Simulation of Household Tasks for Generalist Robots}},
  author    = {Soroush Nasiriany and Abhiram Maddukuri and Lance Zhang and Adeet Parikh and Aaron Lo and Abhishek Joshi and Ajay Mandlekar and Yuke Zhu},
  booktitle = {Proceedings of Robotics: Science and Systems (RSS)},
  address   = {Delft, Netherlands},
  month     = jul,
  year      = {2024},
  doi       = {10.15607/RSS.2024.XX.050},
}

@inproceedings{chen2025robotwin20scalabledata,
  title     = {{{RoboTwin 2.0}: A Scalable Data Generator and Benchmark with Strong Domain Randomization for Robust Bimanual Robotic Manipulation}},
  author    = {Tianxing Chen and Zanxin Chen and Baijun Chen and Zijian Cai and Yibin Liu and Zixuan Li and Qiwei Liang and Xianliang Lin and Yiheng Ge and Zhenyu Gu and Weiliang Deng and Yubin Guo and Tian Nian and Xuanbing Xie and Qiangyu Chen and Kailun Su and Tianling Xu and Guodong Liu and Mengkang Hu and Huan-Ang Gao and Kaixuan Wang and Zhixuan Liang and Yusen Qin and Xiaokang Yang and Ping Luo and Yao Mu},
  booktitle = {Proceedings of the 43rd International Conference on Machine Learning},
  series    = {Proceedings of Machine Learning Research},
  volume    = {306},
  pages     = {13673--13699},
  publisher = {PMLR},
  year      = {2026},
}

@misc{fei2025liberoplusindepthrobustnessanalysis,
      title={{LIBERO-Plus: In-depth Robustness Analysis of Vision-Language-Action Models}}, 
      author={Senyu Fei and Siyin Wang and Junhao Shi and Zihao Dai and Jikun Cai and Pengfang Qian and Li Ji and Xinzhe He and Shiduo Zhang and Zhaoye Fei and Jinlan Fu and Jingjing Gong and Xipeng Qiu},
      year={2025},
      eprint={2510.13626},
      archivePrefix={arXiv},
      primaryClass={cs.RO},
      url={https://arxiv.org/abs/2510.13626}, 
}

@misc{so101-github,
  title        = {{Standard Open {SO-100} \& {SO-101} Arms}},
  author       = {Rob Knight and Pepijn Kooijmans and Remi Cadene and Simon Alibert and Michel Aractingi and Dana Aubakirova and Adil Zouitine and Martino Russi and Steven Palma and Caroline Pascal and Thomas Wolf},
  howpublished = {GitHub repository, \url{https://github.com/TheRobotStudio/SO-ARM100}},
  year         = {2024},
}

@inproceedings{cadene2026lerobotopensourcelibraryendtoend,
  title     = {{{LeRobot}: An Open-Source Library for End-to-End Robot Learning}},
  author    = {Remi Cadene and Simon Alibert and Francesco Capuano and Michel Aractingi and Adil Zouitine and Pepijn Kooijmans and Jade Choghari and Martino Russi and Caroline Pascal and Steven Palma and Dana Aubakirova and Mustafa Shukor and Jess Moss and Alexander Soare and Quentin Lhoest and Quentin Gallou{\'e}dec and Thomas Wolf},
  booktitle = {International Conference on Learning Representations (ICLR)},
  year      = {2026},
  url       = {https://openreview.net/forum?id=CiZMMAFQR3},
}

@misc{oi2026actioncachetrainingfreeaccelerationvisionlanguageaction,
      title={{ActionCache: Training-Free Acceleration for Vision-Language-Action Models with Action Caching and Refinement}}, 
      author={Ryuji Oi and Hikari Otsuka and Kosuke Matsushima and Yuki Ichikawa and Masato Motomura and Tatsuya Kaneko and Daichi Fujiki},
      year={2026},
      eprint={2607.06370v2},
      archivePrefix={arXiv},
      primaryClass={cs.RO},
      url={https://arxiv.org/abs/2607.06370v2}, 
}

@inproceedings{chen2025falconfastvisuomotorpolicies,
  title     = {{{Falcon}: Fast Visuomotor Policies via Partial Denoising}},
  author    = {Haojun Chen and Minghao Liu and Chengdong Ma and Xiaojian Ma and Zailin Ma and Huimin Wu and Yuanpei Chen and Yifan Zhong and Mingzhi Wang and Qing Li and Yaodong Yang},
  booktitle = {Proceedings of the 42nd International Conference on Machine Learning},
  series    = {Proceedings of Machine Learning Research},
  volume    = {267},
  pages     = {8552--8573},
  publisher = {PMLR},
  year      = {2025},
}

@inproceedings{Duan_2025,
   title={{Real-time Iteration Scheme for Diffusion Policy}},
   url={http://dx.doi.org/10.1109/IROS60139.2025.11247391},
   DOI={10.1109/iros60139.2025.11247391},
   booktitle={2025 IEEE/RSJ International Conference on Intelligent Robots and Systems (IROS)},
   publisher={IEEE},
   author={Duan, Yufei and Yin, Hang and Kragic, Danica},
   year={2025},
   month=Oct, pages={11758–11764} }

@misc{he2026vlarollingfewerdenoisingsteps,
      title={{VLA-Rolling: Fewer Denoising Steps, Better Control via Adaptive Action Reuse}}, 
      author={Sicheng He and Yi Pan and Qiuting Yu and Jiaming Huang and Qi Lu and Shuyan Wu and Hongyan Feng and Xuanyu Liu and Jun Zhang and Jie Zhang and Wenqi Zhang and Xuhong Zhang and Jianwei Yin},
      year={2026},
      eprint={2610.25043},
      archivePrefix={arXiv},
      primaryClass={cs.AI},
      url={https://arxiv.org/abs/2610.25043}, 
}

@inproceedings{han2026adavlaadaptivestepflow,
  title     = {{AdaVLA: Adaptive Step Flow Matching for Training-free Acceleration of Vision-Language-Action Models}},
  author    = {Sunghwan Han and Youngtae Han and Youngmin Yi},
  booktitle = {2026 IEEE/RSJ International Conference on Intelligent Robots and Systems (IROS)},
  year      = {2026},
  note      = {arXiv:2608.29208}
}

@inproceedings{yang2025efficientvlatrainingfreeaccelerationcompression,
  title     = {{EfficientVLA: Training-Free Acceleration and Compression for Vision-Language-Action Models}},
  author    = {Yantai Yang and Yuhao Wang and Zichen Wen and Luo Zhongwei and Chang Zou and Zhipeng Zhang and Chuan Wen and Linfeng Zhang},
  booktitle = {Advances in Neural Information Processing Systems},
  volume    = {38},
  year      = {2025}
}

@misc{zhou2026rmuscleroboticmusclememory,
      title={{rMuscle: Robotic Muscle Memory for Efficient Vision-Language-Action Model Inference}}, 
      author={Kaijun Zhou and Zhiyang Li and Le Chen and Jinyu Gu},
      year={2026},
      eprint={2609.19104},
      archivePrefix={arXiv},
      primaryClass={cs.RO},
      url={https://arxiv.org/abs/2609.19104}, 
}

@misc{zhao2026c3acheacceleratingworldaction,
      title={{C$^3$ache: Accelerating World Action Models with Cross Inference Chunk Cache}}, 
      author={Weisen Zhao and Lam Nguyen and Zhicong Lu and Yuzhang Shang},
      year={2026},
      eprint={2606.08962},
      archivePrefix={arXiv},
      primaryClass={cs.LG},
      url={https://arxiv.org/abs/2606.08962}, 
}

@misc{liu2026efficientworldactionmodel,
  title         = {{Efficient World Action Model Inference with Adaptive Intermediate States}},
  author        = {Zhinan Liu and Haozhi Han and Ruge Zhang and Teng Ma and Tao Ma and Zheng Liu and Yifeng Chen and Yunquan Zhang and Ting Cao and Yunxin Liu and Kun Li},
  year          = {2026},
  eprint        = {2609.34608},
  archivePrefix = {arXiv},
  primaryClass  = {cs.RO},
  url           = {https://arxiv.org/abs/2609.34608}
}

@inproceedings{ji2026blockwiseadaptivecachingaccelerating,
  title     = {{Block-wise Adaptive Caching for Accelerating Diffusion Policy}},
  author    = {Kangye Ji and Yuan Meng and Hanyun Cui and Ye Li and Jianbo Zhou and Shengjia Hua and Lei Chen and Zhi Wang},
  booktitle = {International Conference on Learning Representations (ICLR)},
  year      = {2026},
  url       = {https://openreview.net/forum?id=c6ZWfQLOWD}
}

@inproceedings{ye2025radprapidadaptivediffusion,
  title     = {{RA-DP: Rapid Adaptive Diffusion Policy for Training-Free High-frequency Robotics Replanning}},
  author    = {Xi Ye and Rui Heng Yang and Jun Jin and Yinchuan Li and Amir Rasouli},
  booktitle = {2025 IEEE/RSJ International Conference on Intelligent Robots and Systems (IROS)},
  pages     = {6227--6234},
  year      = {2025},
  doi       = {10.1109/IROS60139.2025.11246593}
}

@inproceedings{xu2025vlacacheefficientvisionlanguageactionmanipulation,
  title     = {{VLA-Cache: Efficient Vision-Language-Action Manipulation via Adaptive Token Caching}},
  author    = {Siyu Xu and Yunke Wang and Chenghao Xia and Dihao Zhu and Tao Huang and Chang Xu},
  booktitle = {Advances in Neural Information Processing Systems},
  volume    = {38},
  year      = {2025}
}

@misc{zheng2025leveragingoslevelprimitivesrobotic,
      title={{Leveraging OS-Level Primitives for Robotic Action Management}}, 
      author={Wenxin Zheng and Boyang Li and Bin Xu and Erhu Feng and Jinyu Gu and Haibo Chen},
      year={2025},
      eprint={2508.10259},
      archivePrefix={arXiv},
      primaryClass={cs.OS},
      url={https://arxiv.org/abs/2508.10259}, 
}

@inproceedings{prasad2024consistencypolicyacceleratedvisuomotor,
  title     = {{Consistency Policy: Accelerated Visuomotor Policies via Consistency Distillation}},
  author    = {Aaditya Prasad and Kevin Lin and Jimmy Wu and Linqi Zhou and Jeannette Bohg},
  booktitle = {Robotics: Science and Systems XX},
  year      = {2024},
  doi       = {10.15607/RSS.2024.XX.071}
}

@inproceedings{wang2024onestepdiffusionpolicyfast,
  title     = {{One-Step Diffusion Policy: Fast Visuomotor Policies via Diffusion Distillation}},
  author    = {Wang, Zhendong and Li, Max and Mandlekar, Ajay and Xu, Zhenjia and Fan, Jiaojiao and Narang, Yashraj and Fan, Linxi and Zhu, Yuke and Balaji, Yogesh and Zhou, Mingyuan and Liu, Ming-Yu and Zeng, Yu},
  booktitle = {Proceedings of the 42nd International Conference on Machine Learning},
  series    = {Proceedings of Machine Learning Research},
  volume    = {267},
  pages     = {63399--63416},
  year      = {2025},
  publisher = {PMLR},
  url       = {https://proceedings.mlr.press/v267/wang25ba.html}
}

@inproceedings{kwon2025rtcachetrainingfreeretrievalrealtime,
  title     = {{RT-Cache: Training-Free Retrieval for Real-Time Manipulation}},
  author    = {Owen Kwon and Abraham George and Alison Bartsch and Amir Barati Farimani},
  booktitle = {2025 IEEE-RAS 24th International Conference on Humanoid Robots (Humanoids)},
  pages     = {1--8},
  year      = {2025},
  doi       = {10.1109/Humanoids65713.2025.11203198}
}

@inproceedings{pari2021surprisingeffectivenessrepresentationlearning,
  title     = {{The Surprising Effectiveness of Representation Learning for Visual Imitation}},
  author    = {Jyothish Pari and Nur Muhammad Shafiullah and Sridhar Pandian Arunachalam and Lerrel Pinto},
  booktitle = {Robotics: Science and Systems XVIII},
  year      = {2022},
  doi       = {10.15607/RSS.2022.XVIII.010}
}

@inproceedings{zhao2026retrievethensteeronlinesuccessmemory,
  title     = {{Retrieve-then-Steer: Online Success Memory for Test-Time Adaptation of Generative VLAs}},
  author    = {Jianchao Zhao and Huoren Yang and Yusong Hu and Yuyang Gao and Qiguan Ou and Cong Wan and SongLin Dong and Zhiheng Ma and Yihong Gong},
  booktitle = {Advances in Neural Information Processing Systems},
  volume    = {39},
  year      = {2026},
  note      = {arXiv:2605.10094}
}

@inproceedings{agarwal2023approximatecachingefficientlyserving,
  author    = {Shubham Agarwal and Subrata Mitra and Sarthak Chakraborty and Srikrishna Karanam and Koyel Mukherjee and Shiv Kumar Saini},
  title     = {{Approximate Caching for Efficiently Serving {Text-to-Image} Diffusion Models}},
  booktitle = {21st USENIX Symposium on Networked Systems Design and Implementation (NSDI 24)},
  year      = {2024},
  pages     = {1173--1189},
  address   = {Santa Clara, CA},
  publisher = {USENIX Association},
  isbn      = {978-1-939133-39-7},
  url       = {https://www.usenix.org/conference/nsdi24/presentation/agarwal-shubham}
}

@inproceedings{ma2023deepcacheacceleratingdiffusionmodels,
  title     = {{DeepCache: Accelerating Diffusion Models for Free}},
  author    = {Xinyin Ma and Gongfan Fang and Xinchao Wang},
  booktitle = {Proceedings of the IEEE/CVF Conference on Computer Vision and Pattern Recognition (CVPR)},
  pages     = {15762--15772},
  year      = {2024},
  doi       = {10.1109/CVPR52733.2024.01492}
}

@inproceedings{liu2025timestepembeddingtellsits,
  title     = {{Timestep Embedding Tells: It's Time to Cache for Video Diffusion Model}},
  author    = {Feng Liu and Shiwei Zhang and Xiaofeng Wang and Yujie Wei and Haonan Qiu and Yuzhong Zhao and Yingya Zhang and Qixiang Ye and Fang Wan},
  booktitle = {Proceedings of the IEEE/CVF Conference on Computer Vision and Pattern Recognition (CVPR)},
  pages     = {7353--7363},
  year      = {2025},
  doi       = {10.1109/CVPR52734.2025.00689}
}

@inproceedings{zhao2025realtimevideogenerationpyramid,
  title     = {{Real-Time Video Generation with Pyramid Attention Broadcast}},
  author    = {Xuanlei Zhao and Xiaolong Jin and Kai Wang and Yang You},
  booktitle = {International Conference on Learning Representations (ICLR)},
  year      = {2025}
}

@misc{barik2026soundweavercompositionalwarmstartingtexttoaudio,
      title={{SoundWeaver: Compositional Warm-Starting for Text-to-Audio Diffusion Serving}}, 
      author={Ayush Barik and Sofia Stoica and Nikhil Sarda and Arnav Kethana and Abhinav Khanduja and Muchen Xu and Fan Lai},
      year={2026},
      eprint={2603.07865},
      archivePrefix={arXiv},
      primaryClass={cs.SD},
      url={https://arxiv.org/abs/2603.07865}, 
}

@inproceedings{kwon2023efficientmemorymanagementlarge,
  author    = {Kwon, Woosuk and Li, Zhuohan and Zhuang, Siyuan and Sheng, Ying and Zheng, Lianmin and Yu, Cody Hao and Gonzalez, Joseph E. and Zhang, Hao and Stoica, Ion},
  title     = {{Efficient Memory Management for Large Language Model Serving with {PagedAttention}}},
  booktitle = {Proceedings of the 29th Symposium on Operating Systems Principles (SOSP '23)},
  pages     = {611--626},
  publisher = {ACM},
  year      = {2023},
  doi       = {10.1145/3600006.3613165}
}

@inproceedings{zheng2024sglangefficientexecutionstructured,
  author    = {Zheng, Lianmin and Yin, Liangsheng and Xie, Zhiqiang and Sun, Chuyue and Huang, Jeff and Yu, Cody Hao and Cao, Shiyi and Kozyrakis, Christos and Stoica, Ion and Gonzalez, Joseph E. and Barrett, Clark and Sheng, Ying},
  title     = {{{SGLang}: Efficient Execution of Structured Language Model Programs}},
  booktitle = {Advances in Neural Information Processing Systems},
  volume    = {37},
  pages     = {62557--62583},
  publisher = {Curran Associates, Inc.},
  year      = {2024},
  doi       = {10.52202/079017-2000}
}

@inproceedings{gim2024promptcachemodularattention,
  author    = {Gim, In and Chen, Guojun and Lee, Seung-seob and Sarda, Nikhil and Khandelwal, Anurag and Zhong, Lin},
  title     = {{Prompt Cache: Modular Attention Reuse for Low-Latency Inference}},
  booktitle = {Proceedings of Machine Learning and Systems},
  volume    = {6},
  pages     = {325--338},
  year      = {2024}
}

@inproceedings{Fu2023GPTCacheAO,
  author    = {Fu, Bang and Feng, Di},
  title     = {{{GPTCache}: An Open-Source Semantic Cache for {LLM} Applications Enabling Faster Answers and Cost Savings}},
  booktitle = {Proceedings of the 3rd Workshop for Natural Language Processing Open Source Software (NLP-OSS 2023)},
  pages     = {212--218},
  address   = {Singapore},
  publisher = {Association for Computational Linguistics},
  month     = dec,
  year      = {2023},
  doi       = {10.18653/v1/2023.nlposs-1.24},
  url       = {https://aclanthology.org/2023.nlposs-1.24/}
}

@inproceedings{Yu_2025, series={SOSP ’25},
   title={{IC-Cache: Efficient Large Language Model Serving via In-context Caching}},
   url={http://dx.doi.org/10.1145/3731569.3764829},
   DOI={10.1145/3731569.3764829},
   booktitle={Proceedings of the ACM SIGOPS 31st Symposium on Operating Systems Principles},
   publisher={ACM},
   author={Yu, Yifan and Gan, Yu and Sarda, Nikhil and Tsai, Lillian and Shen, Jiaming and Zhou, Yanqi and Krishnamurthy, Arvind and Lai, Fan and Levy, Hank and Culler, David},
   year={2025},
   month=Oct, pages={375–398},
   collection={SOSP ’25} }

@misc{jiang2026rosaroboticsfoundationmodel,
      title={{ROSA: A Robotics Foundation Model Serving System for Robot Factories}}, 
      author={Wenqi Jiang and Jason Clemons and Rowland O'Flaherty and Hugo Hadfield and Alperen Degirmenci and Shuran Song and Yashraj Narang and Christos Kozyrakis},
      year={2026},
      eprint={2607.01088},
      archivePrefix={arXiv},
      primaryClass={cs.RO},
      url={https://arxiv.org/abs/2607.01088}, 
}

@misc{bansal2026actionchunkschedulingbatched,
      title={{Action Chunk Scheduling for Batched Robot Policy Serving}}, 
      author={Rohan Bansal and David He and Nadun Ranawaka Arachchige and Zhenyang Chen and Soobum Kim and Kexin Rong and Danfei Xu},
      year={2026},
      eprint={2608.00337},
      archivePrefix={arXiv},
      primaryClass={cs.RO},
      url={https://arxiv.org/abs/2608.00337}, 
}

@misc{adamopoulos2026efficientvisionlanguageactionmanagementserving,
      title={{Efficient Vision-Language-Action Management and Serving for Robot Factories}}, 
      author={Dionysios Adamopoulos and Nattapol Chanpaisit and Basel Fakhri and Christina Giannoula},
      year={2026},
      eprint={2609.12075},
      archivePrefix={arXiv},
      primaryClass={cs.DC},
      url={https://arxiv.org/abs/2609.12075}, 
}

@inproceedings{NIPS2002_c3e4035a,
 author = {Xing, Eric P. and Ng, Andrew Y. and Jordan, Michael I. and Russell, Stuart},
 booktitle = {Advances in Neural Information Processing Systems},
 editor = {S. Becker and S. Thrun and K. Obermayer},
 pages = {},
 publisher = {MIT Press},
 title = {{Distance Metric Learning with Application to Clustering with Side-Information}},
 url = {https://proceedings.neurips.cc/paper_files/paper/2002/file/c3e4035af2a1cde9f21e1ae1951ac80b-Paper.pdf},
 volume = {15},
 year = {2002}
}

@inproceedings{3041838.3041840,
  author    = {Bar-Hillel, Aharon and Hertz, Tomer and Shental, Noam and Weinshall, Daphna},
  title     = {{Learning Distance Functions using Equivalence Relations}},
  booktitle = {Proceedings of the Twentieth International Conference on Machine Learning (ICML 2003)},
  editor    = {Fawcett, Tom and Mishra, Nina},
  pages     = {11--18},
  publisher = {AAAI Press},
  address   = {Washington, DC, USA},
  year      = {2003}
}

@inproceedings{NIPS2007_013a006f,
 author = {Rahimi, Ali and Recht, Benjamin},
 booktitle = {Advances in Neural Information Processing Systems},
 editor = {J. Platt and D. Koller and Y. Singer and S. Roweis},
 pages = {},
 publisher = {Curran Associates, Inc.},
 title = {{Random Features for Large-Scale Kernel Machines}},
 url = {https://proceedings.neurips.cc/paper_files/paper/2007/file/013a006f03dbc5392effeb8f18fda755-Paper.pdf},
 volume = {20},
 year = {2007}
}

@misc{zhang2026regilretrievalguidedimitationlearning,
      title={{ReGIL: Retrieval-Guided Imitation Learning from a Single Demonstration}}, 
      author={Yuying Zhang and Francesco Verdoja and Wenyan Yang and Ville Kyrki},
      year={2026},
      eprint={2606.09381},
      archivePrefix={arXiv},
      primaryClass={cs.RO},
      url={https://arxiv.org/abs/2606.09381}, 
}

@inproceedings{huey2025imitationlearningsingletemporally,
  author    = {Huey, William and Wang, Huaxiaoyue and Wu, Anne and Artzi, Yoav and Choudhury, Sanjiban},
  title     = {{Imitation Learning from a Single Temporally Misaligned Video}},
  booktitle = {Proceedings of the 42nd International Conference on Machine Learning},
  series    = {Proceedings of Machine Learning Research},
  volume    = {267},
  pages     = {26283--26310},
  publisher = {PMLR},
  year      = {2025}
}

@misc{openpi2025,
      title={{openpi: Open-Source Models and Packages for Robotics}},
      author={{Physical Intelligence}},
      year={2025},
      howpublished={\url{https://github.com/Physical-Intelligence/openpi}},
}

@inproceedings{Ansel2024PyTorch2F,
  author    = {Ansel, Jason and Yang, Edward and He, Horace and Gimelshein, Natalia and Jain, Animesh and Voznesensky, Michael and Bao, Bin and Bell, Peter and Berard, David and Burovski, Evgeni and Chauhan, Geeta and Chourdia, Anjali and Constable, Will and Desmaison, Alban and DeVito, Zachary and Ellison, Elias and Feng, Will and Gong, Jiong and Gschwind, Michael and Hirsh, Brian and Huang, Sherlock and Kalambarkar, Kshiteej and Kirsch, Laurent and Lazos, Michael and Lezcano, Mario and Liang, Yanbo and Liang, Jason and Lu, Yinghai and Luk, C. K. and Maher, Bert and Pan, Yunjie and Puhrsch, Christian and Reso, Matthias and Saroufim, Mark and Siraichi, Marcos Yukio and Suk, Helen and Zhang, Shunting and Suo, Michael and Tillet, Phil and Zhao, Xu and Wang, Eikan and Zhou, Keren and Zou, Richard and Wang, Xiaodong and Mathews, Ajit and Wen, William and Chanan, Gregory and Wu, Peng and Chintala, Soumith},
  title     = {{{PyTorch} 2: Faster Machine Learning Through Dynamic {Python} Bytecode Transformation and Graph Compilation}},
  booktitle = {Proceedings of the 29th ACM International Conference on Architectural Support for Programming Languages and Operating Systems, Volume 2 (ASPLOS '24)},
  pages     = {929--947},
  publisher = {ACM},
  year      = {2024},
  doi       = {10.1145/3620665.3640366}
}

@inproceedings{yu2025rlinfflexibleefficientlargescale,
  author    = {Yu, Chao and Wang, Yuanqing and Guo, Zhen and Lin, Hao and Xu, Si and Zang, Hongzhi and Zhang, Quanlu and Wu, Yongji and Zhu, Chunyang and Hu, Junhao and Huang, Zixiao and Wei, Mingjie and Xie, Yuqing and Yang, Ke and Dai, Bo and Xu, Zhexuan and Du, Jiakun and Wang, Xiangyuan and Fu, Xu and Shi, Letong and Liu, Zhihao and Chen, Kang and Liu, Weilin and Liu, Gang and Li, Boxun and Yang, Jianlei and Yang, Zhi and Dai, Guohao and Wang, Yu},
  title     = {{{RLinf}: Flexible and Efficient Large-Scale Reinforcement Learning via Macro-to-Micro Flow Transformation}},
  booktitle = {20th USENIX Symposium on Operating Systems Design and Implementation (OSDI 26)},
  pages     = {829--846},
  address   = {Seattle, WA},
  publisher = {USENIX Association},
  month     = jul,
  year      = {2026},
  isbn      = {978-1-939133-55-7},
  url       = {https://www.usenix.org/conference/osdi26/presentation/yu-chao}
}

@misc{chen2026pitextttrlonlinerlfinetuning,
      title={{$\pi_\texttt{RL}$: Online RL Fine-tuning for Flow-based Vision-Language-Action Models}}, 
      author={Kang Chen and Zhihao Liu and Tonghe Zhang and Zhen Guo and Si Xu and Hao Lin and Hongzhi Zang and Xiang Li and Quanlu Zhang and Zhaofei Yu and Guoliang Fan and Tiejun Huang and Yu Wang and Chao Yu},
      year={2025},
      eprint={2510.25889},
      archivePrefix={arXiv},
      primaryClass={cs.LG},
      url={https://arxiv.org/abs/2510.25889}, 
}

@misc{yang2026dpliberocheckpoints,
      title={{dp-libero-checkpoints}},
      author={Brandon Yang},
      year={2026},
      howpublished={\url{https://huggingface.co/brandonyang/dp-libero-checkpoints}},
      note={Hugging Face model repository},
}
\bibliographystyle{mlsys2025}

\clearpage
\appendix
\onecolumn
\section{Step Reduction, Warm Start, and Prior Accelerations}\label{app:steps}
Table~\ref{tab:app-steps} lists every policy, benchmark, and number of denoising steps behind Figure~\ref{fig:steps}, and Table~\ref{tab:app-warm} gives warm start on both LIBERO suites.

\begin{table*}[!htb]
\centering
\footnotesize
\setlength{\tabcolsep}{5pt}
\begin{tabular}{@{}llcccl@{}}
\toprule
Benchmark & Policy & Default & 1 step & 2 steps & More steps \\
\midrule
LIBERO-Spatial & $\pi_{0.5}$ & 98.8 (10) & 98.8 & 98.2 & \\
 & GR00T N1.5 & 94.6 (8) & 93.2 & 91.8 & 3: 94.6 \\
 & Diffusion Policy & 91.2 (100) & & & 5: 90.6, 10: 90.0 \\
 & Fast-WAM & 90.8 (10) & 90.4 & 91.8 & \\
 & Cosmos Policy & 98.6 (5) & 97.4 & 97.4 & 3: 98.4 \\
\midrule
LIBERO-Long & $\pi_{0.5}$ & 83.6 (10) & 82.4 & 82.8 & 4: 85.0 \\
 & GR00T N1.5 & 86.8 (8) & 85.6 & 87.2 & 3 to 7: 84.2 to 88.6 \\
 & Diffusion Policy & 79.6 (100) & & & 5: 75.4, 10: 76.4 \\
 & Fast-WAM & 91.6 (10) & 88.6 & 89.8 & \\
 & Cosmos Policy & 98.0 (5) & & 97.2 & \\
\midrule
LIBERO-Object & $\pi_{0.5}$ & 97.6 (10) & 98.4 & 98.8 & 4: 97.8 \\
LIBERO-Goal & $\pi_{0.5}$ & 95.2 (10) & 93.2 & 95.6 & 4: 94.4 \\
\midrule
MT50 & $\pi_{0.5}$ & 56.3 (10) & 50.0 & 59.6 & 3: 61.3, 4: 58.9, 5: 53.8 \\
 & GR00T N1.5 & 75.8 (4) & 72.2 & 75.2 & 3: 75.2 \\
 & Diffusion Policy & 61.2 (100) & & & 10: 59.4 \\
 & Fast-WAM & 57.2 (10) & 44.6 & 57.0 & 3: 59.1, 4: 58.5, 5: 58.7 \\
\midrule
RoboCasa & Cosmos Policy & 67.1 (5) & 67.3 & & \\
 & X-WAM & 79.2 (10) & 79.3 & 76.8 & 3: 78.6, 5: 78.1 \\
RoboTwin 2.0 & X-WAM & 90.8 (10) & 86.4 & & \\
\bottomrule
\end{tabular}
\vspace{-3.0pt}
\caption{Success rate (\%) of plain step reduction. Default gives the default number of steps in parentheses, and More steps gives the number of steps before each success rate. Diffusion Policy reduces steps with DDIM, and the default of Cosmos Policy on RoboCasa is the value reported by its authors.}
\label{tab:app-steps}
\end{table*}

\begin{table}[!htb]
\centering
\scriptsize
\setlength{\tabcolsep}{2.2pt}
\begin{tabular}{@{}lcccccc@{}}
\toprule
& \multicolumn{3}{c}{LIBERO-Spatial} & \multicolumn{3}{c}{LIBERO-Long} \\
\cmidrule(lr){2-4}\cmidrule(l){5-7}
Policy (steps) & Default & Plain & Warm & Default & Plain & Warm \\
\midrule
$\pi_{0.5}$ (2) & 98.6 & 99.6 & 92.6 & 84.4 & 84.8 & 77.8 \\
GR00T N1.5 (1) & 94.4 & 93.4 & 93.8 & 88.0 & 84.6 & 71.2 \\
GR00T N1.5 (2) & 94.4 & 93.6 & 93.8 & 88.0 & 84.2 & 81.8 \\
Diffusion Policy (10) & & & & 79.6 & 79.2 & 79.6 \\
Fast-WAM (1) & & & & 91.6 & 88.2 & 88.8 \\
Cosmos Policy (2) & & & & 98.6 & 96.6 & 97.4 \\
\bottomrule
\end{tabular}
\vspace{-3.0pt}
\caption{At the same number of steps, warm start is never clearly better than plain step reduction and loses up to 13.4 points. Success rate (\%) over 500 episodes per cell. These runs are separate from those of Table~\ref{tab:app-steps}.}
\label{tab:app-warm}
\end{table}

\begin{table}[!htb]
\centering
\footnotesize
\setlength{\tabcolsep}{5pt}
\begin{tabular}{@{}llcccl@{}}
\toprule
& & Cross-episode & \multicolumn{2}{c}{Work skipped} & \\
\cmidrule(lr){4-5}
Method & Policies & key & Backbone & Head & Policy call \\
\midrule
ActionCache & VLA & VLM output & \hbnone & \hbhalf & Every decision \\
VLA-Rolling & VLA & \xmark & \hbnone & \hbhalf & Every decision \\
AdaVLA & VLA & \xmark & \hbhalf & \hbhalf & Every decision \\
EfficientVLA & VLA & \xmark & \hbhalf & \hbhalf & Every decision \\
rMuscle & VLA & Image, joints & \hbhalf & \hbhalf & Every decision \\
C$^3$ache & WAM & \xmark & \hbnone & \hbhalf & Every decision \\
WAMachine & WAM & \xmark & \hbnone & \hbhalf & Every decision \\
Falcon & DP, VLA & \xmark & \hbnone & \hbhalf & Every decision \\
BAC & DP, VLA & \xmark & \hbnone & \hbhalf & Every decision \\
RA-DP & DP & \xmark & \hbnone & \hbhalf & Every decision \\
RT-Cache & VLA & DINOv2\,\&\,SigLIP & \hbfull & \hbfull & Never \\
\midrule
\textbf{\name} & All & \textbf{Policy encoder} & \hbfull & \hbfull & \textbf{On stall or refresh} \\
\bottomrule
\end{tabular}
\vspace{-3.0pt}
\caption{\name reuses complete decisions from earlier episodes, so a cache hit skips both the backbone and the action head, for all three families of robot policies. Cross-episode key is what a reused decision is looked up by, work skipped is \hbfull{} all, \hbhalf{} part, or \hbnone{} none of the computation at a decision, and policy call gives when the policy runs. \cite{oi2026actioncachetrainingfreeaccelerationvisionlanguageaction,he2026vlarollingfewerdenoisingsteps,han2026adavlaadaptivestepflow,yang2025efficientvlatrainingfreeaccelerationcompression,zhou2026rmuscleroboticmusclememory,zhao2026c3acheacceleratingworldaction,liu2026efficientworldactionmodel,chen2025falconfastvisuomotorpolicies,ji2026blockwiseadaptivecachingaccelerating,ye2025radprapidadaptivediffusion,kwon2025rtcachetrainingfreeretrievalrealtime}}
\label{tab:diff}
\end{table}

\paragraph{Prior accelerations.}
Table~\ref{tab:diff} compares the work each prior acceleration skips at a decision with that of RT-Cache and \name, and Appendix~\ref{app:convert} recomputes their speedups on our optimized implementation, since the papers differ in policy, benchmark, hardware, and what a speedup counts.
C$^3$ache, rMuscle, EfficientVLA, and VLA-Rolling report no success rate for plain step reduction, and Falcon compares with DPMSolver but not with DDIM at the same number of steps.
To recover the success rate of full inference, ActionCache refines a hit with extra steps, Falcon filters cached states by a one-step estimate, and VLA-Rolling trains a confidence model, whereas plain step reduction changes one parameter.
Where the papers do measure plain step reduction on LIBERO, it holds up.
In the table of ActionCache, one plain step of $\pi_{0.5}$ succeeds in 96.9\% of episodes with 5.84~ms of action-head time, while ActionCache succeeds in 92.1\% to 97.0\% with 6.57 to 14.25~ms.
In the table of AdaVLA, one plain step of $\pi_{0.5}$ succeeds in 97.05\% of episodes at 447~ms per chunk, against 99.45\% at 513~ms for AdaVLA.
Headline speedups also often count only part of the work: the 10.44$\times$ and 40.17$\times$ of ActionCache count only the action head, and its whole call speeds up by 2.12$\times$; Falcon counts denoising steps against 100-step DDPM; and the 2.51$\times$ of C$^3$ache comes from a setting that never refreshes its cache after the first chunk.
Independent ports agree with Figure~\ref{fig:warm}: rMuscle ports Nirvana to $\pi_{0.5}$ and succeeds in 26.9\% of RoboTwin episodes, against 35.2\% for the original policy.

\FloatBarrier

\section{Estimating Prior Speedups on Our Implementation}\label{app:convert}
Each paper reports speedup against its own baseline, on its own hardware, and sometimes as a saving in computation instead of time.
A speedup over an unoptimized baseline can shrink on an optimized one, where kernel launches and framework overheads no longer hide the cost of the stages that a method leaves unchanged.
To put them on one scale, we profile every stage of each policy on our compiled implementation (Section~\ref{sec:setup}) and convert the work each method does at a decision into the time it would take there.
Each conversion takes the most optimistic gain that the paper allows, and wherever a paper leaves a quantity open, we take the value most favorable to the method.
Each estimate is therefore an upper bound on what the method would gain on our implementation.
A full-step decision takes 67.5~ms for $\pi_{0.5}$ (vision encoder 10.26~ms, backbone 27.69~ms, 2.96~ms per denoising step), 108.1~ms for Fast-WAM (VAE 6.65~ms, video prefill 30.63~ms, 7.09~ms per step), 356.6~ms for Cosmos Policy (VAE 88.04~ms, 268.61~ms for five steps and 54.02~ms for two), and 105.6~ms for Diffusion Policy (observation encoder 1.01~ms, 1.05~ms per DDPM step, 11.25~ms for ten DDIM steps).
Plain step reduction is our own measurement, with the reduced steps of Table~\ref{tab:policies} for VLAs and world action models and 10, 5, or 4 DDIM steps for Diffusion Policy.
Table~\ref{tab:app-proxy} lists every estimate, and Table~\ref{tab:app-fastest} compares the fastest method for each policy with plain inference.

\paragraph{VLA policies.}
ActionCache reports hit rates of 92.6\%, 91.9\%, and 87.7\% on LIBERO with 2, 1, and 0 refinement steps after a hit.
Its key is the VLM output, so every decision still runs the vision encoder and the backbone, and a miss runs all ten steps; a decision therefore costs $t_v + t_b + (hk + (1-h)\cdot 10)\,t_s$ for hit rate $h$ and $k$ refinement steps, which gives 1.48$\times$ to 1.62$\times$ with the lookup ignored.
VLA-Rolling runs 6.33 denoising steps per chunk on average on LIBERO and leaves the backbone unchanged, which gives 1.19$\times$ with its 1.6~ms confidence model ignored.
AdaVLA reports neither its average number of steps nor how much of the VLM it prunes.
Its latencies on a Jetson AGX Orin (447~ms with one step, 678~ms with five, 290~ms for the VLM forward pass, of which 18 MLP blocks take 12.5~ms each) tie the two together through its 513~ms per chunk; the more of the saving comes from pruning, the faster it is on our implementation, so we take every MLP as pruned with about six steps left, which gives at most 1.96$\times$.
EfficientVLA lowers the measured latency of CogACT from 134.5 to 58.9~ms in the language model and from 51.5 to 26.2~ms in the action module, with the vision encoder unchanged, and scaling the backbone and the action head of $\pi_{0.5}$ by these ratios gives 1.80$\times$.
rMuscle is measured end to end against FlashRT, an optimized engine, on an RTX 4090, so we use its 1.29$\times$ for $\pi_{0.5}$ as is.

\paragraph{World action models.}
In the 2.51$\times$ setting of C$^3$ache, steps 0 to 7 of 10 reuse a cached residual instead of running the DiT blocks; taking those steps as free and the VAE and video prefill as unchanged gives 2.10$\times$ on Fast-WAM.
WAMachine continues a replan from the previous one with three instead of five steps on Cosmos Policy and skips 38.4\% of its layers; taking every replan as a continuation and pricing three steps between our two- and five-step measurements gives 2.16$\times$, below the 2.36$\times$ of its own GPU time, which we keep.

\paragraph{Diffusion Policy.}
Falcon reports the average number of denoising steps per Robomimic task, 12.9 to 48.0 on top of 100-step DDPM and 6.5 to 10.5 on top of 16-step DDIM.
Pricing these steps and ignoring the one-step estimates it runs over its buffer gives 2.06$\times$ to 7.28$\times$, and up to 12.69$\times$ when the speedup of DDIM itself is credited to Falcon.
BAC recomputes the transformer blocks at 10 of 100 DDPM steps in its main setting, which gives 9.21$\times$ with the other steps free.
RA-DP reports replanning frequency, calling the network once per control step on a queue of partly denoised actions; pricing each call as one denoising step plus one pass of the observation encoder and ignoring its guidance gradients, a decision of five actions gives 10.27$\times$.
Plain step reduction to five DDIM steps already gives 15.91$\times$, above every prior method for Diffusion Policy.

\begin{table*}[!htb]
\centering
\footnotesize
\setlength{\tabcolsep}{5pt}
\begin{tabular}{@{}llll@{}}
\toprule
Method & Policy, benchmark & In paper & Est.\ speedup \\
\midrule
\multicolumn{4}{@{}l}{\emph{VLA policies}} \\
ActionCache & $\pi_{0.5}$, VLABench & 2.12$\times$ & 1.48 to 1.62$\times$ \\
VLA-Rolling & $\pi_{0.5}$, LIBERO & 1.30$\times$ & 1.19$\times$ \\
AdaVLA & $\pi_{0.5}$, LIBERO & 1.87$\times$ & at most 1.96$\times$ \\
EfficientVLA & CogACT, SIMPLER & 1.93$\times$ & 1.80$\times$ \\
rMuscle & $\pi_{0.5}$, LIBERO & 1.29$\times$ & 1.29$\times$ \\
Plain step reduction & $\pi_{0.5}$, $10\!\to\!1$ steps & & 1.65$\times$ \\
\name & $\pi_{0.5}$ & & 9.01 / 6.70 / 2.33$\times$ \\
\midrule
\multicolumn{4}{@{}l}{\emph{World action models}} \\
C$^3$ache & Fast-WAM, LIBERO & 2.51$\times$ & 2.10$\times$ \\
WAMachine & Cosmos Policy, LIBERO & 2.36$\times$ & 2.36$\times$ \\
Plain step reduction & Fast-WAM 1 step; Cosmos Policy 2 steps & & 2.44$\times$; 2.51$\times$ \\
\name & Fast-WAM; Cosmos Policy & & 7.63 / 5.38 / 5.02$\times$; 3.66 / 3.12$\times$ \\
\midrule
\multicolumn{4}{@{}l}{\emph{Diffusion Policy}} \\
Falcon & Diffusion Policy, Robomimic & 2 to 7.78$\times$ (steps) & 2.06 to 7.28$\times$ (12.69$\times$ with DDIM) \\
BAC & Diffusion Policy, Robomimic & 3.40 to 3.60$\times$ & 9.21$\times$ \\
RA-DP & Diffusion Policy, MetaWorld & 3.6$\times$ (replanning) & 10.27$\times$ \\
Plain step reduction & DDIM 10 / 5 / 4 steps & & 8.61 / 15.91 / 19.16$\times$ \\
\name & Diffusion Policy & & 129.47 / 118.18 / 24.63$\times$ \\
\bottomrule
\end{tabular}
\vspace{-3.0pt}
\caption{Every estimate of Appendix~\ref{app:convert}, next to plain step reduction and \name with fewer steps and a longer horizon. In paper is the speedup each paper reports on its own baseline, with its unit in parentheses where it is not end-to-end time, and Est.\ speedup is our upper-bound estimate on our implementation. \name gives LIBERO-Spatial / LIBERO-Long / MT50.}
\label{tab:app-proxy}
\end{table*}

\begin{table*}[!htb]
\centering
\footnotesize
\setlength{\tabcolsep}{6pt}
\begin{tabular}{@{}llrrrr@{}}
\toprule
& \multicolumn{2}{c}{Fastest prior acceleration} & \multicolumn{2}{c}{Plain inference} & \\
\cmidrule(lr){2-3}\cmidrule(lr){4-5}
Policy & Method & Est.\ speedup & Fewer steps & Fewer steps, longer horizon & \name \\
\midrule
$\pi_{0.5}$ & AdaVLA & $\le$1.96$\times$ & 1.62$\times$ & 2.54$\times$ & 5.20$\times$ \\
Fast-WAM & C$^3$ache & 2.10$\times$ & 2.23$\times$ & 4.40$\times$ & 5.91$\times$ \\
Cosmos Policy & WAMachine & 2.36$\times$ & 2.51$\times$ & 2.51$\times$ & 3.38$\times$ \\
Diffusion Policy & Falcon & 12.69$\times$ & 8.62$\times$ & 17.01$\times$ & 72.23$\times$ \\
\bottomrule
\end{tabular}
\vspace{-3.0pt}
\caption{Estimated on our implementation, the fastest prior acceleration of each policy is no faster than plain inference with fewer steps and a longer horizon, while \name is faster than both. Est.\ speedup applies the reported setting of each method to our stage latencies with every unknown resolved in its favor (Appendix~\ref{app:convert}). Our columns are geometric means over the benchmarks, and \name uses fewer steps and a longer horizon; Cosmos Policy keeps its execution horizon.}
\label{tab:app-fastest}
\end{table*}

\FloatBarrier

\section{Stagnation Detection}\label{app:stagnation}

This appendix describes the stagnation check of Section~\ref{sec:correct} for one task.
All distances are Euclidean distances between keys after the map $W^\top$ of Equation~\ref{eq:metric}, so they equal the Mahalanobis distance of Section~\ref{sec:reuse}.

\paragraph{Templates.}
The templates are the successful library episodes of the task.
We sample each template every $C$ control steps, where $C$ is the expanded execution horizon, and always keep its last frame, so template frames are spaced like the decisions of the runtime workflow.
A frame at control step $s$ of a template with $T_e$ control steps has phase $\psi = s/T_e$, which runs from $0$ to $1$.
With $E$ templates and $\bar T_e$ the median template length counted in units of $C$ control steps, the window covers $m = \max(2, \lceil 0.05\,\bar T_e \rceil)$ intervals and the estimate uses $K = \min(5, E-1)$ templates.

\paragraph{Alignment.}
At each decision the check keeps the keys $q_0, \dots, q_m$ of the last $m+1$ decisions, which span $\ell$ control steps.
For a template with frames $f_1, \dots, f_n$, it finds the nondecreasing indices $j_0 \le j_1 \le \dots \le j_m$ that minimize the mean distance
\begin{equation}
u = \frac{1}{m+1} \sum_{k=0}^{m} \lVert q_k - f_{j_k} \rVert_2 .
\label{eq:alignment}
\end{equation}
Indices may repeat or skip frames.
Dynamic programming over the window gives the exact minimum, and ties go to earlier indices.
The advance of the window under this template is
\begin{equation}
\delta = \bigl(\psi_{j_m} - \psi_{j_0}\bigr)\, \frac{mC}{\ell},
\label{eq:advance}
\end{equation}
where the factor $mC/\ell$ rescales a window that spans more or fewer than $mC$ control steps.

\paragraph{Estimate.}
The check keeps the $K$ templates with the smallest cost $u$.
Its estimate $\hat\delta$ is the median of their advances.
It also records the median start phase $\hat\psi$, the median cost $\hat u$, and the spread $\hat v$, the difference between the $90$th and $10$th percentiles of the advances.
Every percentile, the median included, takes the lower order statistic and never interpolates.

\paragraph{Calibration with the episode left out.}
Before deployment we slide the window over every successful library episode and compute the estimate with that episode removed from the templates.
Each window gives its true advance $\delta^\star$, read from the episode's own phases and rescaled as in Equation~\ref{eq:advance}, and the residual $\delta^\star - \hat\delta$.
We describe each window by the context $\bigl(\hat\psi, F_u(\hat u), F_v(\hat v)\bigr)$, where $F_u$ and $F_v$ are the empirical distribution functions of $\hat u$ and $\hat v$ over all windows of the task, with every episode given equal weight.

\paragraph{Bounds at runtime.}
At a decision, the check computes the context of the current window and takes from every successful library episode its one window whose context is nearest in $L_1$ distance, with ties going to the earlier window.
Over these windows, $\bar\epsilon$ is the $90$th percentile of the residuals and $\underline\delta$ is the $10$th percentile of the true advances.
The check needs windows from at least four library episodes.

\paragraph{Decision.}
The check maps $\hat\delta$, $\bar\epsilon$, and $\underline\delta$ to one of three outcomes:
\begin{equation}
\label{eq:stagnation}
\begin{cases}
\text{call policy}, & \hat\delta+\bar\epsilon<\underline\delta,\\
\text{observe sooner}, & \hat\delta<\underline\delta\le\hat\delta+\bar\epsilon,\\
\text{execute cached chunk}, & \text{otherwise}.
\end{cases}
\end{equation}
The first case is a one-sided test.
Because $\bar\epsilon$ bounds how much the estimate understates the true advance, $\hat\delta+\bar\epsilon$ is an optimistic estimate, and the check calls the policy only when even this estimate falls below $\underline\delta$, the advance of the slower successful episodes.
Every false alarm costs a policy call, so the test prefers missing a mild slowdown to calling the policy too often.
The second case is an uncertainty band, in which the check gathers more evidence at the cost of one extra decision.
A call happens unless the previous decision itself triggered a stagnation call, so two stagnation calls never come back to back.
A request to observe sooner makes the next decision come after the default execution horizon $h$, so the robot skips the rest of the cached chunk.
This happens only when the decision did not call the policy, and at most once every $mC$ control steps.
The check stays inactive while fewer than $m+1$ decisions have accumulated, when the task has fewer than four calibrated library episodes, and after an invalid observation, which also clears the window.
It uses no threshold on robot displacement, gripper state, contact, or task name.

\FloatBarrier

\section{Implementation Details}\label{app:impl}
\paragraph{Integration.}
A policy exposes three hooks.
The first encodes the camera images, the second builds the conditioning prefix from the encoded images, the instruction, and the state, and the third samples an action chunk from that prefix with a given number of denoising steps.
\name calls the first hook at every decision and the other two only when it calls the policy, so a policy call is the unchanged forward pass of the policy.
The same cache code serves all five policies, and adding a policy means wrapping its forward pass in these hooks and choosing where to read the key.

\paragraph{Keys.}
Table~\ref{tab:keys} lists where each policy's key is read.
Every key is a tensor that the first hook computes anyway, so reading it adds only a pooling step.
PCA then reduces each camera to 64 dimensions (Section~\ref{sec:reuse}), and the state is appended.

\begin{table}[!htb]
\centering
\footnotesize
\begin{tabular*}{\columnwidth}{@{\extracolsep{\fill}}llr@{}}
\toprule
Policy & Key read from & Dimensions per camera \\
\midrule
$\pi_{0.5}$ & SigLIP image tokens, $16\times16$ grid pooled to $4\times4$ & 32{,}768 \\
GR00T N1.5 & image tokens entering the language model, pooled to $4\times4$ & 32{,}768 \\
Diffusion Policy & output of the ResNet-18 observation encoder & 64 \\
Fast-WAM & video tokens entering the transformer, pooled to $4\times4$ & 49{,}152$^{*}$ \\
Cosmos Policy & VAE latent of the camera image, pooled to $4\times4$ & 256 \\
\bottomrule
\end{tabular*}
\vspace{-8pt}
\caption{Where each policy's key is read, before PCA. $^{*}$Fast-WAM tiles its two cameras into one image, so it has one key for both.}
\label{tab:keys}
\end{table}

\paragraph{Decision loop.}
Algorithm~\ref{alg:loop} shows how \name serves one robot.
Each decision first checks the refresh interval, and once $\tau$ decisions have passed since the last policy call, it calls the policy directly.
Otherwise retrieval, fusion, correction, and the stagnation check run on the CPU, the GPU runs only the first hook, and the policy is called only when the check reports a stall.
The check may also ask to observe sooner (Appendix~\ref{app:stagnation}).
$\textsc{Round}$ rounds a fractional $\tau$ up or down at random so that the mean interval is $\tau$.

\begin{algorithm}[!htb]
\caption{Decision loop of \name for one robot.}
\label{alg:loop}
\footnotesize
\DontPrintSemicolon
\LinesNotNumbered
\KwIn{library $L$ of the instructed task, refresh interval $\tau$, horizons $h$ and $C$, denoising steps $K$}
$next \gets \textsc{Round}(\tau)$ \tcp*{first refresh}
\For{each decision $t = 0, 1, \dots$}{
  $H \gets C$ \tcp*{$C = h$ without the expanded horizon}
  \eIf{$t \ge next$}{
    $a \gets \textsc{Policy}(o_t, \ell, s_t, K)$ \tcp*{periodic refresh, all three hooks, GPU}
    $next \gets t + \textsc{Round}(\tau)$
  }{
    $z \gets \textsc{EncodeImages}(o_t)$ \tcp*{first hook, GPU}
    $q \gets \textsc{Key}(z, s_t)$ \tcp*{pooling, PCA, state}
    $a \gets \textsc{Correct}(\textsc{Fuse}(\textsc{Nearest}(q, L)))$ \tcp*{CPU}
    $status \gets \textsc{StagnationCheck}(q, L)$ \tcp*{CPU}
    \uIf{$status$ is a stall}{
      $a \gets \textsc{SampleActions}(\textsc{Prefix}(z, \ell, s_t), K)$ \tcp*{second and third hooks, GPU}
      $next \gets t + \textsc{Round}(\tau)$
    }
    \ElseIf{the check asks to observe sooner}{
      $H \gets h$
    }
  }
  execute the first $H$ actions of $a$
}
\end{algorithm}

\paragraph{Serving.}
The server keeps one connection per robot.
Each connection holds its own execution cursor, the recent keys of the stagnation check, and the refresh counter, while the libraries, the metric, and the corrector are read-only and shared by all connections.
Policy calls from different robots are batched on the GPU (Appendix~\ref{app:latency}).
The metric, the corrector, and the bounds of the stagnation check are fitted once on the CPU before deployment.

\paragraph{Fitting with the episode left out.}
\name fits all of its learned quantities on the library before deployment, and Table~\ref{tab:fit} lists what each one is fitted on.
At runtime, a new episode never finds itself in the library.
Within one library episode, neighboring decisions look almost the same, so a fit that let a decision find its own episode would see a cache far more accurate than the one a new episode meets.
Every fit that compares a library decision with the rest of the library leaves that decision's episode out.
PCA uses no actions, so it is fitted on all library keys.
The corrector solves the weighted ridge regression
\begin{equation}
\label{eq:ridge}
\min_{B,c}\ \sum_i\omega_i
\|\tilde g_i-B\phi(x_i)-c\|_2^2
+\alpha\|B\|_F^2,
\end{equation}
where $\tilde g_i$ is the residual of library decision $i$ on the executed motion channels, standardized per channel, the input $x_i$ combines the standardized key, the assembled chunk, and the decision index, $\phi(x)=\bigl[x,\ \sqrt{2}\cos(\Omega x+\theta)\bigr]$ appends random Fourier features with $\Omega$ and $\theta$ sampled once, $\alpha=100$, and $\omega_i=1/n_{e(i)}$ with $n_{e(i)}$ the number of training entries from the episode of decision $i$, so every episode contributes the same total weight.
At runtime the corrector predicts $\hat g=B\phi(x)+c$, converts it back to action units, and adds $\beta\,\kappa(r)\,\hat g$ to the assembled chunk, where $\beta=0.5$ and $\kappa(r)$ falls linearly from 1 at $r_{\min}$ to 0 at $r_{\max}$ of the distance $r$ to the nearest entry.

\begin{table}[!htb]
\centering
\small
\begin{tabular*}{\columnwidth}{@{\extracolsep{\fill}}lll@{}}
\toprule
Quantity & Fitted on & Episode left out \\
\midrule
PCA bases & Keys of all library decisions & None \\
Metric $W$ & Key differences of action-matched pairs & The two entries of a pair come from different episodes \\
Corrector $B$, $c$ & Residual of each library chunk from its cache estimate & The chunk's own episode \\
Correction scale & Nearest-neighbor distances of library entries & The entry's own episode \\
Bounds $\bar\epsilon$, $\underline\delta$ & Windows of successful library episodes & The window's own episode (Appendix~\ref{app:stagnation}) \\
Refresh interval $\tau$ & Replay of library episodes & The replayed episode (Appendix~\ref{app:params}) \\
\bottomrule
\end{tabular*}
\vspace{-8.0pt}
\caption{What each learned quantity of \name is fitted on, and which episode the fit leaves out.}
\label{tab:fit}
\end{table}

\paragraph{Overhead and library size.}
The memory of \name grows linearly with the number of library entries.
A library of 2{,}500 MT50 episodes takes at most 203~MB and is fitted in about 14 minutes, and the CPU time per decision on LIBERO-Long stays under 2~ms up to 1{,}000 episodes and reaches 3.6~ms at 5{,}000.
LIBERO has initial states for libraries of about 500 episodes, so we measure larger libraries by replicating the 500-episode library, which leaves the computation per entry unchanged.
Table~\ref{tab:cpu-scale} breaks the CPU time per decision into its parts.
Retrieval, which projects the query and scans the entries of its task, takes 0.71~ms at 500 episodes and 1.55~ms at 5{,}000, and the scan itself grows from 0.18~ms to 0.97~ms.
The stagnation check grows the most, since it compares recent keys with more successful episodes.
All fitting times are measured on a single CPU core.
The fits of different tasks are independent of each other, so on a machine with more cores they run in parallel and finish much sooner.

\begin{table}[!htb]
\centering
\footnotesize
\begin{tabular}{@{}lcccccc@{}}
\toprule
& \multicolumn{6}{c}{Library (episodes)} \\
\cmidrule(l){2-7}
Part & 50 & 200 & 500 & 1{,}000 & 2{,}000 & 5{,}000 \\
\midrule
Retrieval & 0.62 & 0.63 & 0.71 & 0.78 & 0.94 & 1.55 \\
\quad PCA projection & 0.52 & 0.50 & 0.53 & 0.53 & 0.54 & 0.59 \\
\quad Mahalanobis distance & 0.09 & 0.11 & 0.15 & 0.21 & 0.34 & 0.82 \\
\quad Nearest entries & 0.01 & 0.02 & 0.03 & 0.04 & 0.07 & 0.15 \\
Fusion & 0.13 & 0.13 & 0.14 & 0.14 & 0.14 & 0.15 \\
Correction & 0.36 & 0.36 & 0.38 & 0.38 & 0.41 & 0.50 \\
\quad Confidence and bookkeeping & 0.19 & 0.18 & 0.20 & 0.21 & 0.23 & 0.30 \\
\quad Corrector & 0.17 & 0.17 & 0.18 & 0.18 & 0.18 & 0.19 \\
Stagnation check & 0.25 & 0.31 & 0.39 & 0.50 & 0.72 & 1.39 \\
\midrule
Total & 1.36 & 1.43 & 1.61 & 1.80 & 2.21 & 3.58 \\
\bottomrule
\end{tabular}
\vspace{-3.0pt}
\caption{Retrieval stays under 2~ms per decision even with 5{,}000 episodes. Each part of Table~\ref{tab:cpu} is followed by its steps. CPU time (ms) of each part of \name for $\pi_{0.5}$ on LIBERO-Long on one pinned core of an idle Xeon E5-2696 v4, averaged over 2{,}048 decisions. Libraries above 500 episodes replicate the 500-episode library. A library of 500 episodes holds 22{,}816 entries from its 436 successful episodes, and one of 5{,}000 episodes holds 228{,}160.}
\label{tab:cpu-scale}
\end{table}

\FloatBarrier

\section{Parameters and Their Sensitivity}\label{app:params}

\name combines standard learning tools, each with its usual constants: a regularized Mahalanobis metric, kernel-weighted averaging of neighbors, ridge regression on random Fourier features, percentile bounds on a progress estimate, and a ladder of operating points.
Three design choices keep these constants from becoming a tuning burden.
First, every quantity that depends on the data, such as the PCA bases, the metric $W$, the corrector, and the bounds of the stagnation check, is fitted on the library alone (Table~\ref{tab:fit}).
Second, most remaining constants are relative by construction: $\lambda$ is scaled by $\operatorname{tr}(S)/D$, $r_{\min}$ and $r_{\max}$ are in units of the median nearest-neighbor distance, the corrector works on standardized inputs and normalized targets, $\beta$ is a share, the bounds are percentiles, and the window is a fraction of the median episode length.
Third, the one setting that must differ between tasks, how often the policy runs, is chosen by the ladder of Section~\ref{sec:correct}.

\begin{figure}[!t]
\centering
\includegraphics[width=\textwidth]{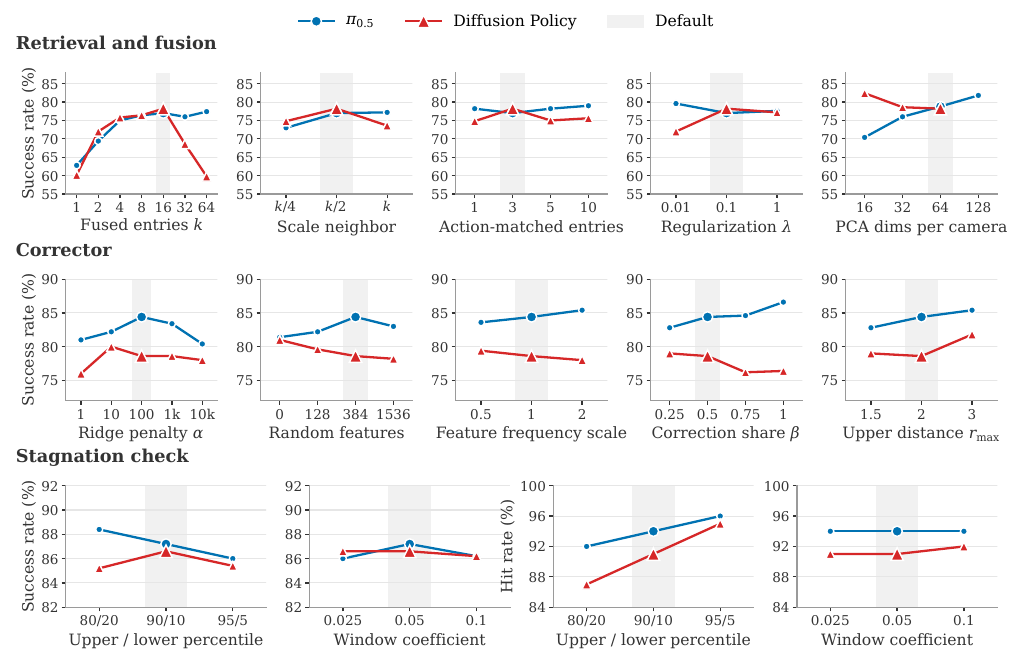}
\vspace{-27.0pt}
\caption{Success rate, and for the stagnation check also the hit rate, when one constant changes, on LIBERO-Long with libraries of 200 episodes. Gray bands mark the defaults, and each point averages 500 episodes. Diffusion Policy has no 128-dimensional setting because its keys have 64 dimensions per camera.}
\label{fig:sensitivity}
\end{figure}

Table~\ref{tab:params} lists the constants, and Figure~\ref{fig:sensitivity} changes one at a time for $\pi_{0.5}$ and Diffusion Policy on LIBERO-Long with libraries of 200 episodes, keeping the 8th entry as the scale when the number of fused entries changes.
Each row of the figure starts from the matching row of Table~\ref{tab:ladder}: fusion without the corrector and the corrector without the stagnation check, which never call the policy, and the stagnation check without periodic refresh.

\begin{table}[!htb]
\centering
\small
\setlength{\tabcolsep}{3pt}
\begin{tabular}{@{}llll@{}}
\toprule
Component & Constant & Default & Role \\
\midrule
Retrieval and fusion & Fused entries $k$ & 16 & entries averaged into the assembled chunk \\
& Scale neighbor & $k/2=8$ & its distance sets the width of the fusion weights \\
& Action-matched entries & 3 & entries per decision that form $S$ \\
& Regularization $\lambda$ & 0.1 & keeps poorly sampled directions from dominating \\
& PCA dimensions & 64 per camera & size of the visual part of the key \\
\midrule
Corrector & Ridge penalty $\alpha$ & 100 & shrinks the corrector weights $B$ \\
& Random features & 384 & number of rows of $\Omega$ \\
& Frequency scale & $\Omega_{ij}\sim\mathcal{N}(0, 1/p)$ & width of the approximated Gaussian kernel \\
& Correction share $\beta$ & 0.5 & share of the predicted gap that is added \\
& Distance bounds $r_{\min}$, $r_{\max}$ & 0.75, 2 & where the share starts to fall and reaches zero \\
\midrule
Stagnation check & Percentiles of $\bar\epsilon$, $\underline\delta$ & 90th, 10th & how strong the evidence of a stall must be \\
& Window coefficient & 0.05 & window length $m=\max(2,\lceil 0.05\,\bar T_e\rceil)$ (Appendix~\ref{app:stagnation}) \\
\midrule
Budgeting & Target hit rates & 50\% to 90\% & levels of the ladder in Section~\ref{sec:correct} \\
\bottomrule
\end{tabular}
\vspace{-3.0pt}
\caption{Constants of \name and their defaults.}
\label{tab:params}
\end{table}

\paragraph{Budgeting.}
The ladder of Section~\ref{sec:correct} has levels with target hit rates between 50\% and 90\%, followed by the level where only the stagnation check calls the policy.
To map a target hit rate to the refresh interval $\tau$, we replay the library episodes with the cache, holding out each episode from the library it queries, and search for the $\tau$ whose replayed hit rate meets the target.
A level passes a task if its success rate is at most 4 points below that of pure inference.

\paragraph{Overall.}
Outside the number of fused entries, 43 of the 54 single changes stay within 3 points of the default and none loses more than 8.4 points, and the number of action-matched entries and the frequency scale of the random features change the success rate by at most 3.4 and 1.0 points.

\paragraph{Fusion.}
Fusing one or two entries loses 6 to 18 points, the gain of fusion in Table~\ref{tab:ladder}, and equal weights for the 16 entries lose 7 points for $\pi_{0.5}$ and 9 for Diffusion Policy, so averaging helps only when closer entries count more.
Diffusion Policy, trained to reproduce several modes of the demonstrations \cite{chi2024diffusionpolicyvisuomotorpolicy}, also loses 10 to 18 points with 32 or 64 entries and 4.6 points when the 16th entry sets the scale, since distant entries can hold a different motion, while $\pi_{0.5}$ is unaffected by both.
Taking the scale from the 4th entry loses 3 to 4 points for both, so sixteen entries with the $(k/2)$th setting the scale is the widest averaging that is safe for both policies.

\paragraph{Key.}
$\pi_{0.5}$ gains 3.0 points from 128 PCA dimensions per camera and 2.6 from a weaker regularization and loses 8.4 points with 16 dimensions, whereas Diffusion Policy gains 4.2 points from 16 dimensions and loses 6.2 with the weaker regularization.
The large pretrained encoder of $\pi_{0.5}$ carries useful detail in many directions, while the 64 dimensions per camera of the encoder of Diffusion Policy, trained with the policy, gain from compression and stronger regularization that suppress noise.

\paragraph{Corrector and stagnation check.}
For $\pi_{0.5}$, the ridge penalty has a clear optimum at the default, with 3 to 4 points lost at both extremes, the random features add 3 points over a linear corrector, and a larger share $\beta$ adds up to 2.2 points, whereas for Diffusion Policy every corrector constant stays within 3.2 points of the default and a share above the default loses about 2 points.
Looser percentiles (80th and 20th) call the policy more often and lower the hit rate by 2 to 4 points, gaining 1.2 points for $\pi_{0.5}$ and losing 1.4 for Diffusion Policy, while tighter ones (95th and 5th) call it less often and lose about one point for both, and the window coefficient changes the success rate by at most 1.2 points.

\paragraph{One setting for both policies.}
Where the two policies disagree, the defaults, 64 dimensions with $\lambda=0.1$, $\beta=0.5$, and the stagnation percentiles, lie between their preferences rather than at either optimum, so one setting serves both a VLA with a large pretrained encoder and a diffusion policy with a small one, and the settings that cause large losses, too few fused entries, equal weights, and a key compressed too far for a large encoder, all lie far from the defaults.

\FloatBarrier

\section{Policy Training}\label{app:training}
On LIBERO we use the released checkpoints of $\pi_{0.5}$, GR00T N1.5, Fast-WAM, and Cosmos Policy, and on MT50 the $\pi_{0.5}$ checkpoint released by RLinf \cite{yu2025rlinfflexibleefficientlargescale,chen2026pitextttrlonlinerlfinetuning}.
We train the other four policies with the settings of Table~\ref{tab:training}.
On LIBERO, Diffusion Policy is trained on the demonstrations of all 40 tasks.
On MT50, every policy is trained on the 2{,}500 demonstrations of the 50 tasks, 50 per task \cite{shukor2025smolvlavisionlanguageactionmodelaffordable}.

Diffusion Policy is the Transformer variant with a ResNet-18 image encoder and frozen CLIP text embeddings \cite{chi2024diffusionpolicyvisuomotorpolicy,radford2021learningtransferablevisualmodels}.
On LIBERO it starts from released language-conditioned weights \cite{yang2026dpliberocheckpoints}, and on MT50 from random initialization.
GR00T N1.5 starts from its base model, keeps the vision-language model frozen, and trains its action head, following NVIDIA's fine-tuning script \cite{nvidia2025gr00tn15}.
Fast-WAM starts from its released LIBERO checkpoint and trains the action expert, with new input and output layers for the four-dimensional state and action of MetaWorld.

\begin{table}[!htb]
\centering
\footnotesize
\begin{tabular*}{\columnwidth}{@{\extracolsep{\fill}}lcccc@{}}
\toprule
 & Diffusion Policy & Diffusion Policy & GR00T N1.5 & Fast-WAM \\
Benchmark & LIBERO & MT50 & MT50 & MT50 \\
\midrule
Initialization & LIBERO release & random & base model & LIBERO release \\
Trained parameters & all but text encoder & all but text encoder & action head & action expert \\
Updates & 100k & 100k & 60k & 16k \\
Batch size & 256 & 256 & 64 & 128 \\
Peak / final learning rate & $10^{-4}$ / $10^{-5}$ & $10^{-4}$ / $10^{-5}$ & $10^{-4}$ / 0 & $10^{-4}$ / $10^{-6}$ \\
Warmup updates & 500 & 500 & 3{,}000 & 800 \\
Weight decay & $10^{-6}$ & $10^{-6}$ & $10^{-5}$ & 0.01 \\
Cameras & 2 & 1 & 1 & 1 \\
Images & $96^2$, $84^2$ crop & $96^2$, $84^2$ crop & $224^2$, crop, color jitter & $224^2$ \\
\bottomrule
\end{tabular*}
\vspace{-8pt}
\caption{Settings of the four training configurations. All use AdamW with a linear warmup, a cosine decay of the learning rate, and bfloat16 activations.}
\label{tab:training}
\end{table}

\FloatBarrier

\section{Building the Library}\label{app:library}
The library is the only data that \name needs besides the policy.
This appendix asks which episodes to keep when the library must stay small, and whether the episodes must come from the policy that is served.

\paragraph{Collection and evaluation.}
LIBERO provides 1{,}000 initial states per suite, and 500 of them form the official test set.
We run every test episode and collect library episodes only on the other 500 initial states.
On MT50 we run 2{,}500 test episodes and collect libraries with other seeds, so no library initial state appears in testing.
We size a library by the episodes collected, split evenly across tasks, and the three libraries of a policy use different seeds.

\paragraph{A small library should keep only successful episodes.}
We fix the library at 50 episodes of LIBERO-Long, 5 per task, drawn from one pool of collected episodes, and serve with the cache and the corrector only, so the library alone sets the success rate (Table~\ref{tab:library-select}).
A random draw of the collected episodes includes failures, about one in nine for $\pi_{0.5}$ and one in five for Diffusion Policy.
Drawing only from the successful episodes raises the success rate by 2.6 to 3.3 points for $\pi_{0.5}$ and by 14.9 to 18.2 points for Diffusion Policy.

\paragraph{Among successful episodes, representative scenes help both policies.}
We group the first observations of the successful episodes of each task into five clusters by their keys and keep the episode nearest the center of each cluster.
Over successful episodes drawn at random, this adds 3.7 and 5.1 points for $\pi_{0.5}$ and 2.8 and 1.4 points for Diffusion Policy.
With the default horizon, the 50 episodes of $\pi_{0.5}$ chosen this way succeed as often as about 90 episodes drawn at random.
Keeping the five scenes that lie farthest apart instead, which maximizes coverage, adds 4.8 and 1.7 points for $\pi_{0.5}$ but loses 4.8 and 7.9 points for Diffusion Policy, since the farthest scenes are often the least typical ones.

\begin{table}[!htb]
\centering
\footnotesize
\begin{tabular}{@{}lcccc@{}}
\toprule
& \multicolumn{2}{c}{$\pi_{0.5}$} & \multicolumn{2}{c}{Diffusion Policy} \\
\cmidrule(lr){2-3}\cmidrule(l){4-5}
Library of 50 episodes & Default & Expanded & Default & Expanded \\
\midrule
Random & 61.4 & 67.3 & 37.2 & 57.1 \\
Random, successful only & 64.7 & 69.9 & 52.1 & 75.3 \\
Farthest apart & 69.5 & 71.6 & 47.3 & 67.4 \\
Representative & 68.3 & 75.0 & 54.9 & 76.7 \\
\bottomrule
\end{tabular}
\vspace{-3.0pt}
\caption{Representative successful episodes help both policies, and far-apart ones help only $\pi_{0.5}$. Success rate (\%) on LIBERO-Long with the cache and the corrector only, with the default and the expanded horizon, averaged over three runs of 500 episodes. The last two rows choose 5 successful episodes per task from their first observations.}
\label{tab:library-select}
\end{table}

\paragraph{Human demonstrations make a better library.}
The library can also hold the demonstrations that trained the policy.
On LIBERO-Long with 200 demonstrations, $\pi_{0.5}$ succeeds in 94.8\% with only the stagnation check, against 87.2\% with 200 episodes it collected itself ($p<10^{-5}$), and in 94.8\% against 90.4\% with the full system ($p=0.006$).
Diffusion Policy rises from 86.6\% to 88.8\%.

\paragraph{A library borrowed from a stronger policy helps.}
On MT50, a policy can also serve from the library of another policy, with keys from its own vision encoder (Table~\ref{tab:library-borrow}).
With the library of GR00T N1.5, the strongest policy here, $\pi_{0.5}$ rises from 58.6\% to 61.8\% with the full system ($p<10^{-4}$) and from 48.7\% to 60.2\% with only the stagnation check.
With the library of Diffusion Policy, $\pi_{0.5}$ gains 6.3 points with only the stagnation check and keeps its success rate with the full system.
Libraries of weaker policies cost 2.8 to 5.4 points with the full system, so a library is only as good as the policy that collected it.

\begin{table}[!htb]
\centering
\footnotesize
\begin{tabular}{@{}llcc@{}}
\toprule
Served policy & Library from & Stagnation check only & Full system \\
\midrule
$\pi_{0.5}$ (56.3) & GR00T N1.5 (75.6) & 48.7 $\to$ 60.2 & 58.6 $\to$ 61.8 \\
$\pi_{0.5}$ (56.3) & Diffusion Policy (61.2) & 48.7 $\to$ 55.0 & 58.6 $\to$ 58.8 \\
Fast-WAM (56.5) & GR00T N1.5 (75.6) & 46.2 $\to$ 48.3 & 56.9 $\to$ 57.0 \\
Diffusion Policy (61.2) & $\pi_{0.5}$ (56.3) & 53.0 $\to$ 48.0 & 61.4 $\to$ 56.0 \\
Fast-WAM (56.5) & $\pi_{0.5}$ (56.3) & 46.2 $\to$ 44.7 & 56.9 $\to$ 54.1 \\
GR00T N1.5 (75.6) & $\pi_{0.5}$ (56.3) & 56.0 $\to$ 48.8 & 76.0 $\to$ 70.7 \\
\bottomrule
\end{tabular}
\vspace{-3.0pt}
\caption{The library of a stronger policy keeps or raises the success rate, and that of a weaker policy lowers it. Mean success rate over the 50 tasks of MT50 (\%) with the own library $\to$ the borrowed one, with libraries of 20 successful episodes per task. Parentheses give the success rate of pure inference. A task without a successful episode of the lender calls the policy.}
\label{tab:library-borrow}
\end{table}

\paragraph{Why the main experiments use the policy's own episodes.}
Demonstrations give the best library, but whether they are available is uncertain.
A deployed policy may come without its training data, or may have been fine-tuned by someone else.
The successful episodes of the policy itself are available in every deployment, so the experiments on LIBERO and MT50 build the library from them alone.

\FloatBarrier

\section{Batched Serving System}\label{app:latency}
The serving experiment of Section~\ref{sec:e2e} and Table~\ref{tab:stages} use a concurrent server that we wrote for $\pi_{0.5}$.
Pure inference is the same server with the cache turned off, so it runs the same inference code and every optimization of this appendix except the cache-specific ones benefits both settings.

\paragraph{Server.}
One copy of $\pi_{0.5}$ serves all robots from one RTX 4090.
The server splits inference into the vision encoder, the backbone, and the action head, and gives each stage its own queue.
A stage takes every queued request up to its batch limit as soon as it can run, without waiting for more requests to arrive.
The backbone and the action head run as one batch, so a policy call is handed between stages only once.

\paragraph{Shared optimizations.}
All optimizations keep the model in bfloat16 and leave its weights unchanged.
Each stage is traced into a static graph, compiled with TorchInductor \cite{Ansel2024PyTorch2F}, and captured as CUDA graphs at startup for a fixed set of batch sizes, and a smaller batch is padded to the next captured size.
The backbone and the whole denoising loop form one graph.
The prefix skips the slot of the unused third camera and pads the instruction to 80 tokens, which shortens it from 968 to 592 tokens.
The skipped tokens are masked in the original model, so the actions are unchanged up to bfloat16 rounding.
The images of a batch arrive in one pinned buffer and reach the GPU in one copy.
On the CPU, each request runs in its own thread from a pool of 512, each BLAS call uses four threads, and the Python garbage collector is frozen after warmup so that no full collection pauses serving.
These changes lower a full-step call from 67.5~ms to 50.8~ms at batch size 1 (Table~\ref{tab:twolat}) and to 22.5~ms per request at batch size 16 (Table~\ref{tab:stages}).
Pure inference saturates at 42.5 decisions/s, close to the 44 decisions/s that the GPU time of a batched policy call allows, so the baseline uses the GPU well.

\paragraph{Cache-specific optimizations.}
The vision encoder computes the keys of a whole batch on the GPU and copies them to the CPU in one transfer.
It keeps two batches in flight at a higher stream priority, so cache hits do not wait behind policy calls.
Retrieval, correction, and the stagnation check use matrix products that release the Python global interpreter lock, so request threads run them in parallel, and starting an episode reuses the state of its task instead of rebuilding it.
Table~\ref{tab:serveconfig} lists the configuration of both settings.

\begin{table}[!htb]
\centering
\footnotesize
\begin{tabular*}{\columnwidth}{@{\extracolsep{\fill}}lcc@{}}
\toprule
Setting & Pure inference & \name \\
\midrule
Model copies on the GPU & 1 & 1 \\
Precision & bfloat16 & bfloat16 \\
Compilation & TorchInductor, CUDA graphs & TorchInductor, CUDA graphs \\
Prefix tokens & 592 & 592 \\
Batch limit of the vision encoder & 16 & 32 \\
Batch limit of the backbone and action head & 16 & 16 \\
Wait for a fuller batch & none & none \\
Vision encoder batches in flight, stream priority & 1, normal & 2, high \\
Request threads, BLAS threads per call & 512, 4 & 512, 4 \\
Denoising steps of a policy call & 10 & 10, or 1 with reduced steps \\
Library & none & 500 episodes \\
\bottomrule
\end{tabular*}
\vspace{-8pt}
\caption{Server configuration in the serving experiment of Section~\ref{sec:e2e}. Each setting keeps one configuration at every robot count.}
\label{tab:serveconfig}
\end{table}

\begin{figure}[!htb]
\centering
\includegraphics[width=0.6\textwidth]{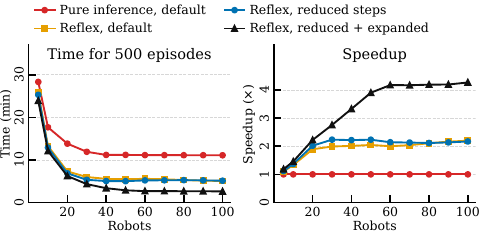}
\vspace{-11.0pt}
\caption{With 60 or more robots, \name finishes the 500 episodes 2.0$\times$ to 2.2$\times$ faster than pure inference, and 4.2$\times$ to 4.3$\times$ faster with reduced steps and the expanded horizon. Time to finish the replay and speedup over pure inference with the default settings, for the runs of Figure~\ref{fig:cluster}.}
\label{fig:cluster-app}
\end{figure}

\paragraph{Workload.}
We record the requests of the 500 LIBERO-Long test episodes by serving them closed-loop with \name, 27{,}747 requests in total, each with two camera images, the state, and the instruction.
A replayer on the same machine acts as $N$ robots, with $N$ from 5 to 100.
Each robot replays recorded episodes one after another from a shuffled queue until all 500 are done, and the robots start evenly spread over one request interval.
After it receives a chunk, a robot waits for five control steps at 20~Hz plus a 2~ms network round trip, 252~ms in total, before it sends the next request.
Pure inference replays the same requests.
For the expanded horizon, a robot sends every other recorded request and waits for ten control steps.
Before each run, 40 episodes at 20 robots warm up the server and are not counted.
We report the latency of each request from its arrival at the server to the reply, the throughput while all robots are active, and the time to finish all 500 episodes.
The hit rate is 78.3\% at every robot count, and 81.4\% with the expanded horizon.
The replay measures serving performance, and Section~\ref{sec:e2e} reports the success rates.
Figure~\ref{fig:cluster-app} gives the time to finish the 500 episodes and the speedup over pure inference at each robot count.

\paragraph{Pricing.}
Section~\ref{sec:setup} prices all five policies with the evaluation code, which is the conservative choice: pricing $\pi_{0.5}$ with the serving system would raise its speedups, for example from 4.4$\times$ to 4.9$\times$ at the default settings on LIBERO-Long.

\begin{table}[!htb]
\centering
\footnotesize
\begin{tabular*}{\columnwidth}{@{\extracolsep{\fill}}lcc@{}}
\toprule
Stage & Evaluation & Serving \\
\midrule
Vision encoder & 10.3 & 6.6 \\
Backbone & 27.7 & 17.9 \\
Action head, 10 steps & 29.6 & 26.3 \\
Action head, 1 step & 3.0 & 3.4 \\
\midrule
Full-step call & 67.5 & 50.8 \\
\bottomrule
\end{tabular*}
\vspace{-8pt}
\caption{GPU time (ms) of each stage of $\pi_{0.5}$ at batch size 1 on one RTX 4090, in the evaluation code of Section~\ref{sec:setup} and in the serving server of Section~\ref{sec:e2e}.}
\label{tab:twolat}
\end{table}

\FloatBarrier

\section{Real-Robot Experiment}\label{app:robot}
\begin{figure}[!h]
\centering
\includegraphics[width=0.25\textwidth]{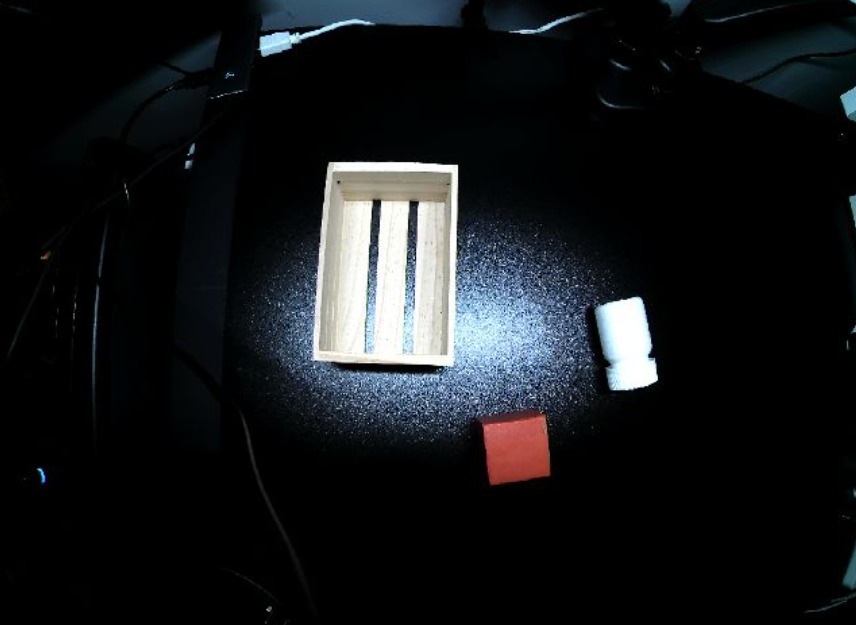}\hspace{12pt}%
\includegraphics[width=0.25\textwidth]{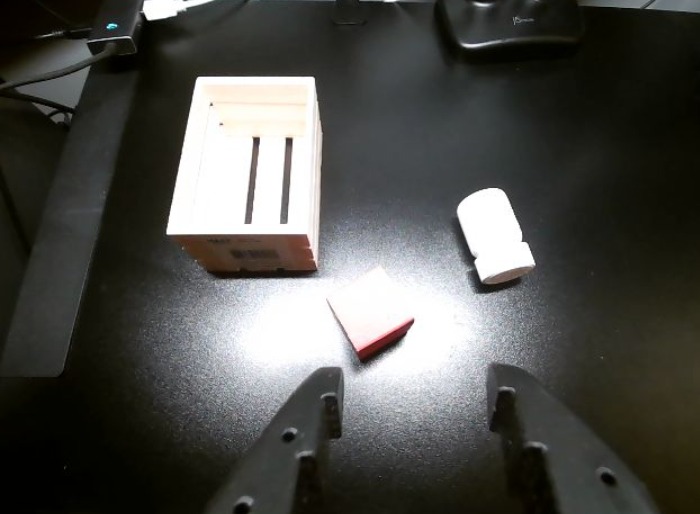}
\vspace{-9.7pt}
\caption{Views of the camera above the table (left) and the wrist camera (right) on the SO-101 arm.}
\label{fig:robot-cams}
\includegraphics[width=0.192\textwidth]{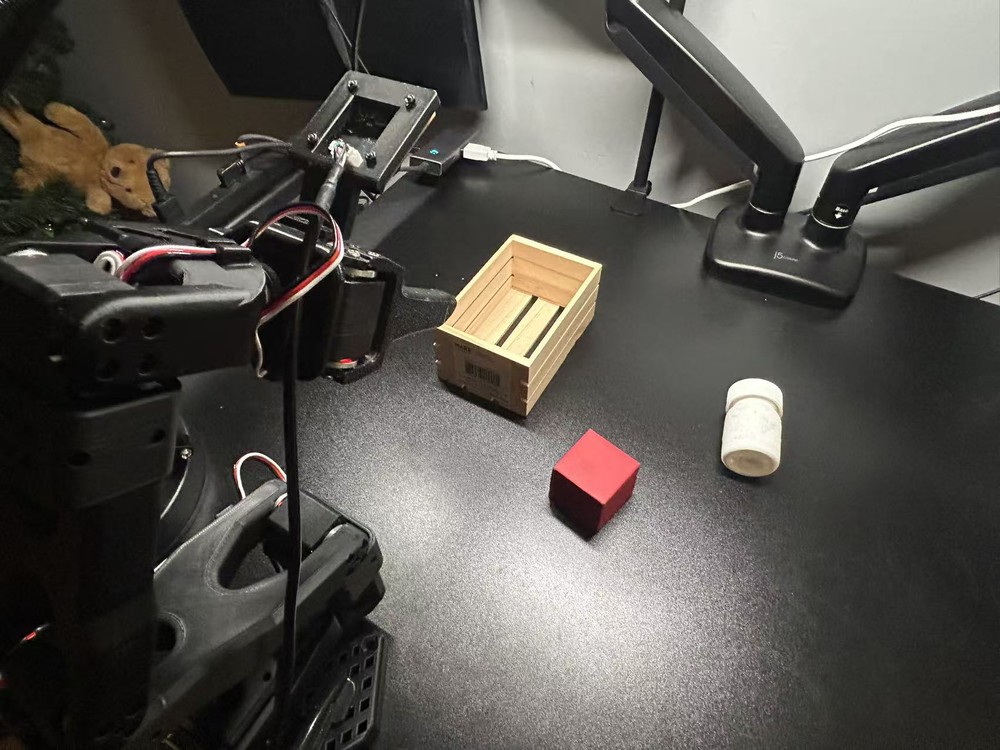}\hspace{0.01\textwidth}\includegraphics[width=0.192\textwidth]{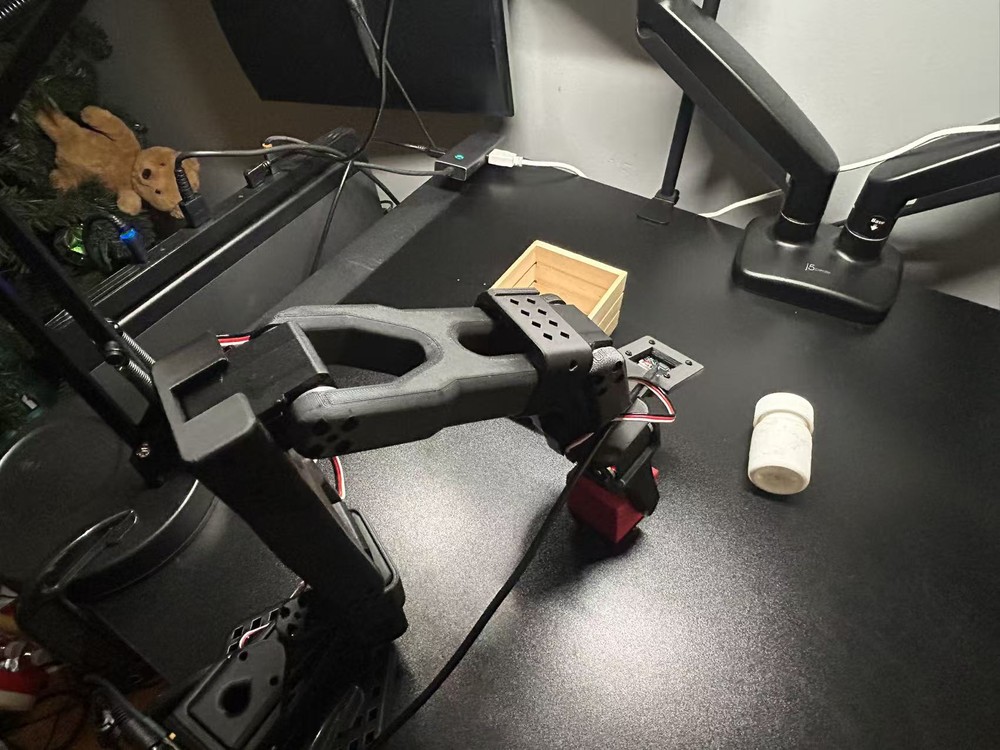}\hspace{0.01\textwidth}\includegraphics[width=0.192\textwidth]{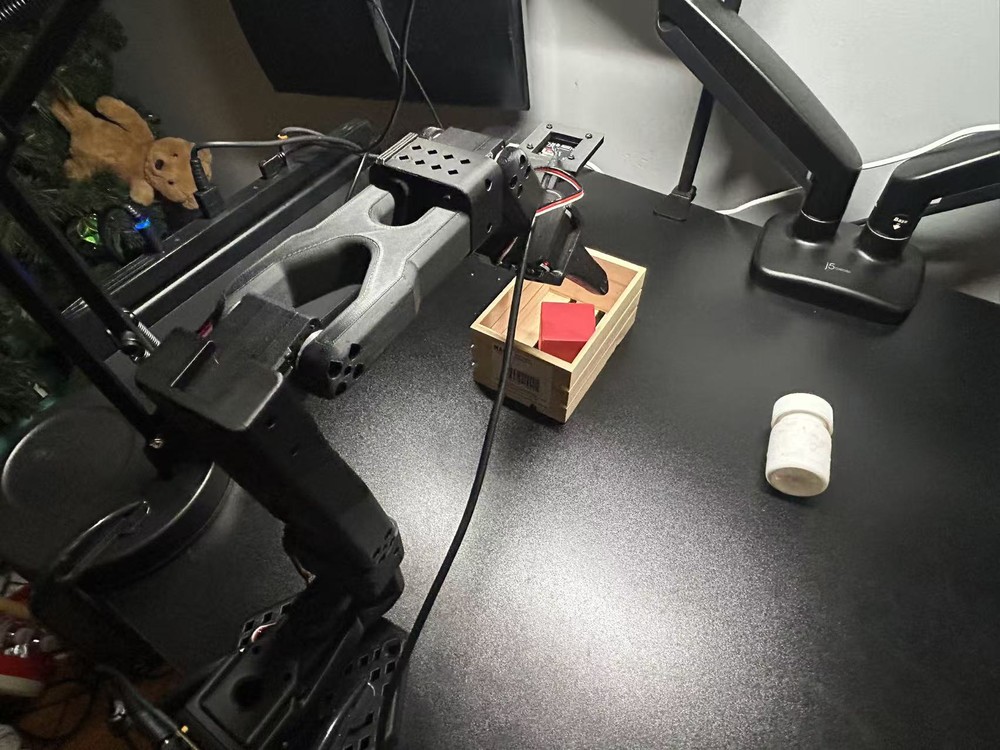}\hspace{0.01\textwidth}\includegraphics[width=0.192\textwidth]{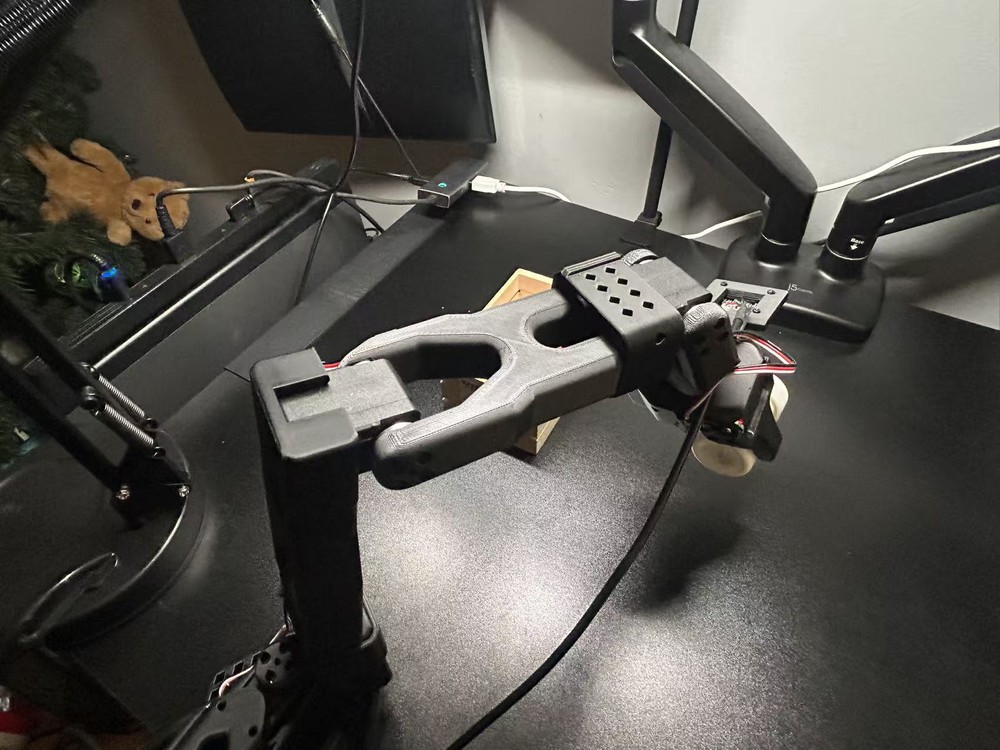}\hspace{0.01\textwidth}\includegraphics[width=0.192\textwidth]{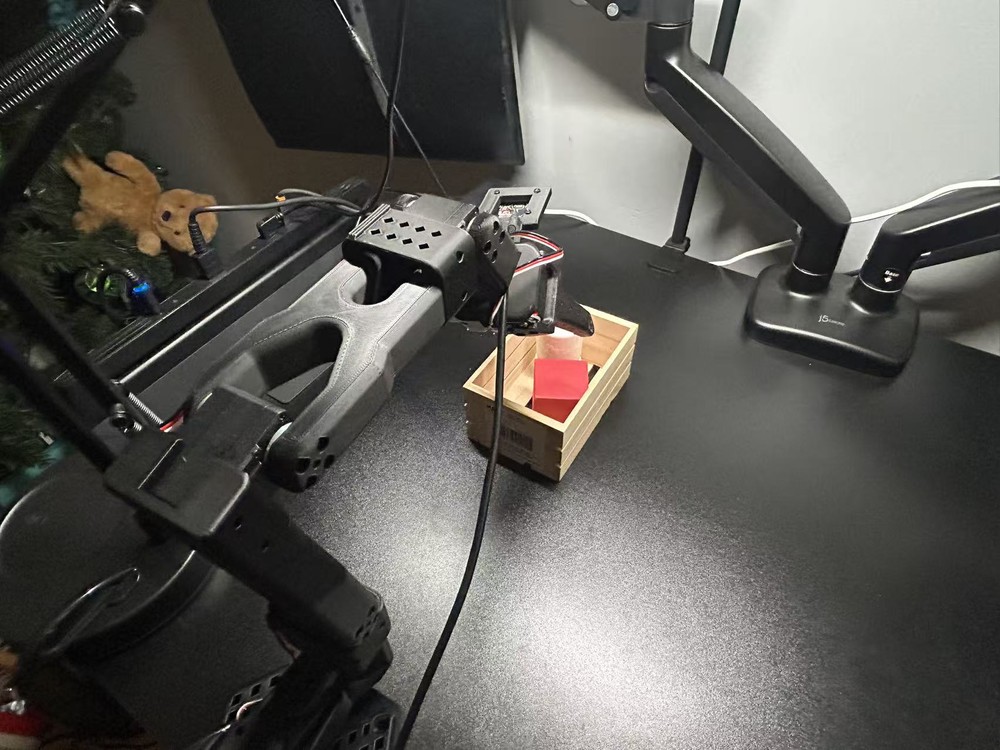}\\[3pt]
\includegraphics[width=0.192\textwidth]{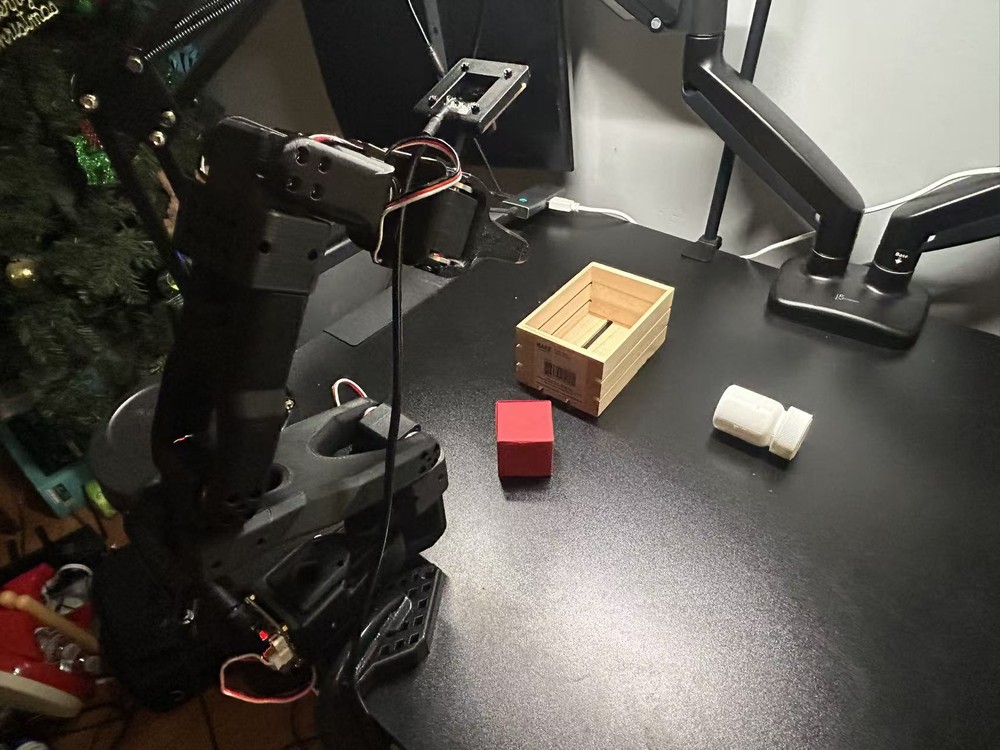}\hspace{0.01\textwidth}\includegraphics[width=0.192\textwidth]{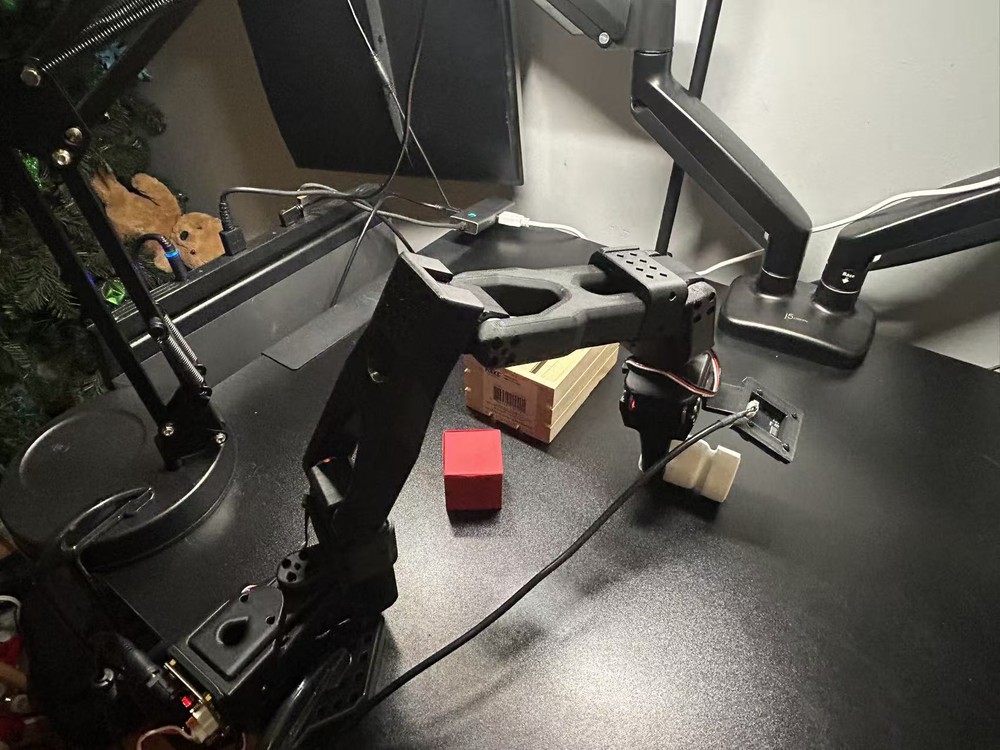}\hspace{0.01\textwidth}\includegraphics[width=0.192\textwidth]{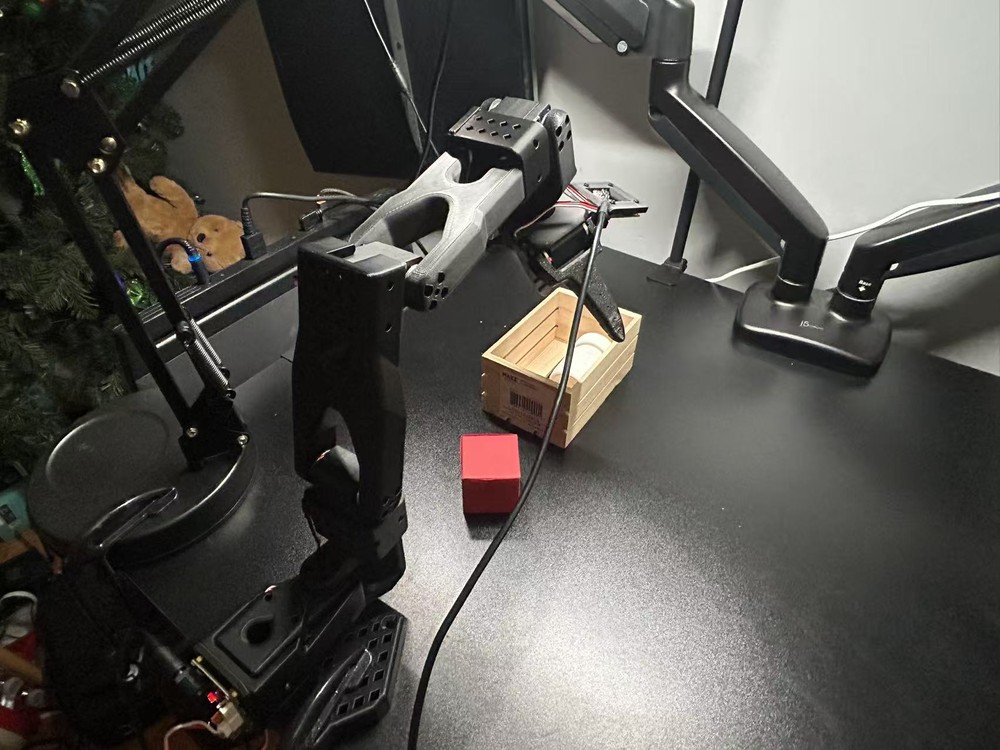}\hspace{0.01\textwidth}\includegraphics[width=0.192\textwidth]{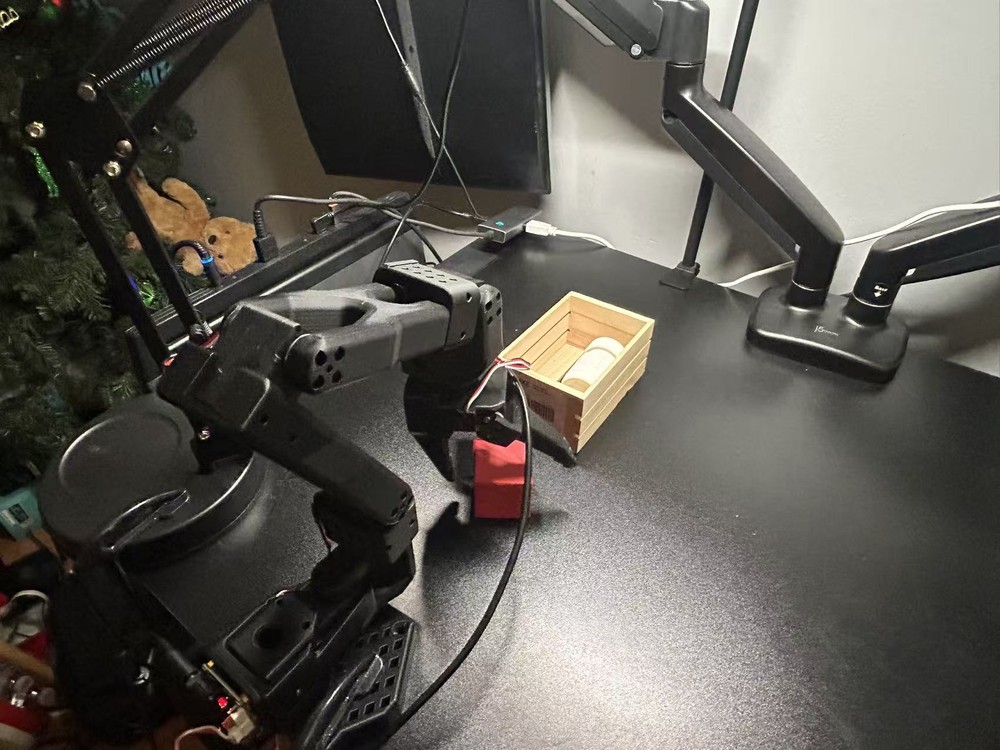}\hspace{0.01\textwidth}\includegraphics[width=0.192\textwidth]{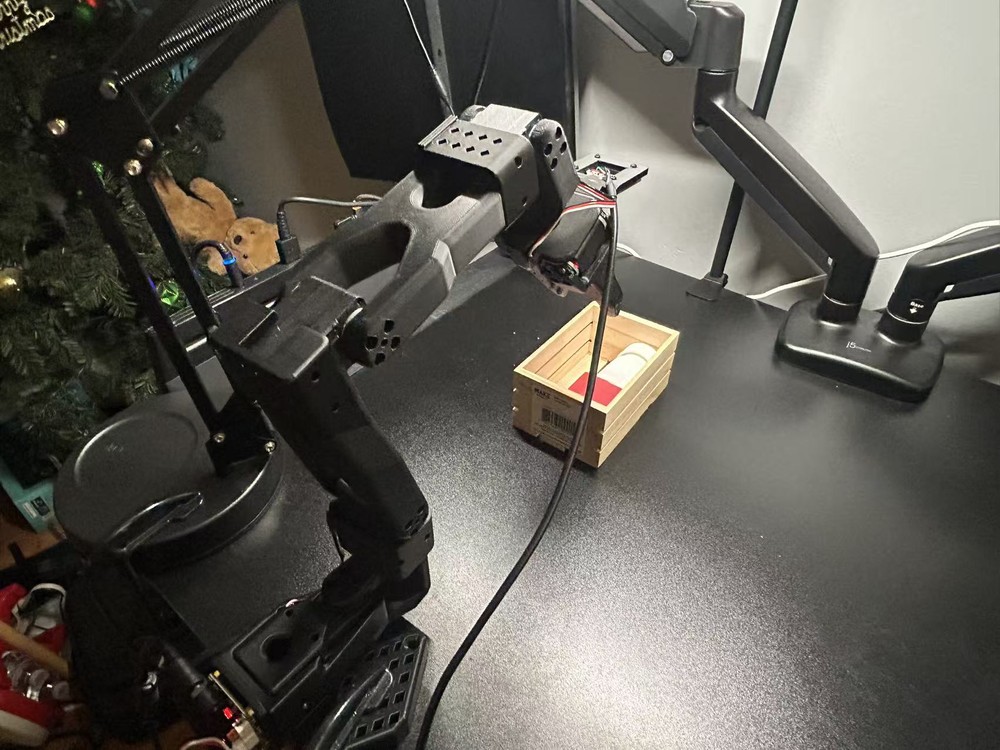}
\vspace{-21.7pt}
\caption{One episode of each task, from left to right. The top row places the red block first, and the bottom row places the white pill bottle first. The bottle always lies on its side and is never stood upright.}
\label{fig:robot-tasks}
\end{figure}

We use an SO-101 arm \cite{so101-github,cadene2026lerobotopensourcelibraryendtoend} with two cameras, one fixed above the table and one on the wrist (Figure~\ref{fig:robot-cams}).
Both tasks place a red block and a white pill bottle into a wooden frame and differ only in the order (Figure~\ref{fig:robot-tasks}).
Their instructions are ``First place the red block into the wooden frame, then place the white pill bottle into the wooden frame'' and ``First place the white pill bottle into the wooden frame, then place the red block into the wooden frame''.

We fine-tune all parameters of the $\pi_{0.5}$ base model \cite{intelligence2025pi05visionlanguageactionmodelopenworld} on 50 demonstrations per task with its flow-matching loss.
Training uses AdamW with a batch size of 32 for 5{,}000 steps in bfloat16, and the learning rate warms up over 166 steps to $2.5\times10^{-5}$ and then follows a cosine decay to $2.5\times10^{-6}$.
Images are $224\times224$ without augmentation, and states and actions are normalized by their 1st and 99th percentiles.
The chunk length is 50, each policy call runs 10 denoising steps, and the default execution horizon is $h=10$.

The cache of each task holds 100 successful episodes, the 50 demonstrations used for fine-tuning and 50 more.
\name uses its default settings, without reduced steps or the expanded horizon.
Each task runs 100 episodes with each setting, and the two settings alternate.
After each episode of pure inference, we put the objects back where it started and run \name from the same placement.
An episode succeeds if both objects are in the frame, placed in the instructed order, within one minute.
On the first task \name succeeds in 93 episodes and pure inference in 91, and on the second in 79 and 83.

\FloatBarrier

\section{Library Size and Component Ablations}\label{app:libsize}\label{app:ablation}
This appendix repeats the results of Section~\ref{sec:eval} for every library size and separates the contribution of each component.

\paragraph{Full system.}
In 294 of the 300 settings of Table~\ref{tab:main-all}, the success rate of \name falls at most 2 points below pure inference with the same settings, and at most 4.3 points in all of them.
At the default settings, growing the library from the smallest to the largest size raises the mean hit rate from 58\% to 93\% on LIBERO-Spatial, from 51\% to 81\% on LIBERO-Long, and from 25\% to 56\% on MT50.

\begin{table*}[p]
\centering
\scriptsize
\renewcommand{\arraystretch}{0.78}
\setlength{\tabcolsep}{3pt}
\setlength{\aboverulesep}{0.5pt}\setlength{\belowrulesep}{0.8pt}
\begin{tabular*}{\textwidth}{@{\extracolsep{\fill}}lllcccccccc@{}}
\toprule
& & & \multicolumn{4}{c}{Pure inference} & \multicolumn{4}{c}{\name} \\
\cmidrule(lr){4-7}\cmidrule(l){8-11}
& Policy & Library & Default & \begin{tabular}[b]{@{}c@{}}Expanded\\horizon\end{tabular} & \begin{tabular}[b]{@{}c@{}}Reduced\\steps\end{tabular} & Both & Default & \begin{tabular}[b]{@{}c@{}}Expanded\\horizon\end{tabular} & \begin{tabular}[b]{@{}c@{}}Reduced\\steps\end{tabular} & Both \\
\midrule
\multirow{30}{*}{\rotatebox[origin=c]{90}{LIBERO-Spatial}} & \multirow{6}{*}{$\pi_{0.5}$} & 50 & \multirow{6}{*}{96.6/67.5} & \multirow{6}{*}{97.8/34.5} & \multirow{6}{*}{98.0/40.9} & \multirow{6}{*}{98.4/20.9} & 96.6/76/23.7 & 96.5/68/14.6 & 97.4/71/19.1 & 96.9/72/9.6 \\
 &  & 100 &  &  &  &  & 97.4/83/20.2 & 97.7/82/10.5 & 96.9/86/14.4 & 97.2/84/7.7 \\
 &  & 200 &  &  &  &  & 96.5/87/17.5 & 97.5/83/10.3 & 97.2/84/15.2 & 97.4/86/7.5 \\
 &  & 300 &  &  &  &  & 96.7/86/18.4 & 97.7/81/10.9 & 97.5/80/16.3 & 97.7/91/6.7 \\
 &  & 400 &  &  &  &  & 96.2/89/16.8 & 98.0/94/7.0 & 96.6/86/14.6 & 97.7/95/6.1 \\
 &  & 500 &  &  &  &  & 97.0/96/12.7 & 98.0/86/9.3 & 96.2/86/14.5 & 97.8/90/6.8 \\
\cmidrule(l){2-11}
 & \multirow{6}{*}{GR00T N1.5} & 50 & \multirow{6}{*}{93.4/41.4} & \multirow{6}{*}{95.4/21.2} & \multirow{6}{*}{93.8/16.8} & \multirow{6}{*}{95.0/8.6} & 94.1/60/20.2 & 96.2/47/12.8 & 94.2/72/9.1 & 95.2/52/5.7 \\
 &  & 100 &  &  &  &  & 94.1/76/14.5 & 96.2/76/7.5 & 94.4/85/7.7 & 95.9/97/3.3 \\
 &  & 200 &  &  &  &  & 95.9/95/8.0 & 95.9/77/7.3 & 96.1/95/6.7 & 95.3/97/3.3 \\
 &  & 300 &  &  &  &  & 95.7/95/7.9 & 96.0/70/8.6 & 95.0/95/6.7 & 95.9/97/3.3 \\
 &  & 400 &  &  &  &  & 96.2/95/7.8 & 96.8/70/8.5 & 96.4/95/6.6 & 96.4/98/3.3 \\
 &  & 500 &  &  &  &  & 97.8/96/7.6 & 97.0/98/3.5 & 96.4/95/6.6 & 97.0/98/3.2 \\
\cmidrule(l){2-11}
 & \multirow{6}{*}{Diffusion Policy} & 50 & \multirow{6}{*}{90.0/107.3} & \multirow{6}{*}{91.4/54.9} & \multirow{6}{*}{91.2/12.5} & \multirow{6}{*}{91.4/6.3} & 92.9/58/45.6 & 90.5/7/51.0 & 92.4/60/5.6 & 90.8/6/6.0 \\
 &  & 100 &  &  &  &  & 93.5/91/11.0 & 91.7/67/18.4 & 93.9/91/2.1 & 90.5/62/2.7 \\
 &  & 200 &  &  &  &  & 94.9/92/9.6 & 92.1/95/3.5 & 95.1/92/1.9 & 92.8/95/0.8 \\
 &  & 300 &  &  &  &  & 95.2/93/8.5 & 93.9/95/3.0 & 94.9/93/1.9 & 93.9/96/0.8 \\
 &  & 400 &  &  &  &  & 95.2/93/8.3 & 94.5/96/2.7 & 94.9/93/1.8 & 93.9/96/0.8 \\
 &  & 500 &  &  &  &  & 95.2/93/8.2 & 94.0/96/2.5 & 95.6/93/1.8 & 94.0/96/0.8 \\
\cmidrule(l){2-11}
 & \multirow{6}{*}{Fast-WAM} & 50 & \multirow{6}{*}{91.0/109.8} & \multirow{6}{*}{92.2/56.0} & \multirow{6}{*}{91.8/45.1} & \multirow{6}{*}{91.0/23.0} & 91.5/73/31.1 & 91.7/22/42.9 & 91.9/66/17.2 & 90.7/16/18.7 \\
 &  & 100 &  &  &  &  & 91.1/82/23.7 & 92.3/9/50.8 & 90.9/82/11.4 & 90.9/14/19.6 \\
 &  & 200 &  &  &  &  & 91.6/93/13.4 & 91.5/15/47.4 & 93.7/84/10.9 & 91.5/42/14.4 \\
 &  & 300 &  &  &  &  & 91.3/92/14.4 & 91.9/11/50.0 & 93.5/83/11.5 & 91.7/39/14.7 \\
 &  & 400 &  &  &  &  & 91.9/92/14.2 & 92.0/10/50.1 & 92.7/90/9.3 & 90.9/11/20.1 \\
 &  & 500 &  &  &  &  & 91.0/91/14.4 & 92.0/10/49.7 & 91.8/93/8.6 & 89.8/16/18.8 \\
\cmidrule(l){2-11}
 & \multirow{6}{*}{Cosmos Policy} & 50 & \multirow{6}{*}{96.4/381.0} & \multirow{6}{*}{n/a} & \multirow{6}{*}{96.6/151.3} & \multirow{6}{*}{n/a} & 96.5/25/310.4 & n/a & 97.3/26/136.4 & n/a \\
 &  & 100 &  &  &  &  & 96.0/71/176.5 & n/a & 96.5/66/113.2 & n/a \\
 &  & 200 &  &  &  &  & 96.7/84/138.9 & n/a & 97.3/82/104.0 & n/a \\
 &  & 300 &  &  &  &  & 96.2/90/122.2 & n/a & 96.6/92/98.0 & n/a \\
 &  & 400 &  &  &  &  & 96.7/84/139.7 & n/a & 96.5/92/98.1 & n/a \\
 &  & 500 &  &  &  &  & 97.2/90/120.9 & n/a & 97.4/91/99.0 & n/a \\
\midrule
\multirow{30}{*}{\rotatebox[origin=c]{90}{LIBERO-Long}} & \multirow{6}{*}{$\pi_{0.5}$} & 50 & \multirow{6}{*}{85.0/67.5} & \multirow{6}{*}{90.8/34.0} & \multirow{6}{*}{82.2/40.9} & \multirow{6}{*}{91.8/20.6} & 90.0/69/27.9 & 92.0/45/21.2 & 89.5/65/21.1 & 92.6/53/12.4 \\
 &  & 100 &  &  &  &  & 87.3/76/24.1 & 91.5/48/20.1 & 84.3/90/13.4 & 92.2/69/10.0 \\
 &  & 200 &  &  &  &  & 89.5/91/15.4 & 92.7/57/17.5 & 89.3/91/13.0 & 93.7/68/10.1 \\
 &  & 300 &  &  &  &  & 87.5/91/15.3 & 92.1/61/16.5 & 87.2/91/13.0 & 94.1/78/8.6 \\
 &  & 400 &  &  &  &  & 90.1/92/14.7 & 92.1/86/9.2 & 89.3/92/12.8 & 93.5/79/8.5 \\
 &  & 500 &  &  &  &  & 89.6/92/14.6 & 93.2/69/14.2 & 88.8/92/12.8 & 93.6/81/8.1 \\
\cmidrule(l){2-11}
 & \multirow{6}{*}{GR00T N1.5} & 50 & \multirow{6}{*}{86.2/41.4} & \multirow{6}{*}{88.0/20.9} & \multirow{6}{*}{87.6/16.8} & \multirow{6}{*}{88.2/8.5} & 86.5/48/24.4 & 88.1/38/14.1 & 87.5/45/12.0 & 87.4/37/6.5 \\
 &  & 100 &  &  &  &  & 86.7/57/21.3 & 88.7/52/11.6 & 87.7/43/12.3 & 88.3/47/6.0 \\
 &  & 200 &  &  &  &  & 87.6/80/13.3 & 88.2/64/9.5 & 88.2/77/8.6 & 89.3/53/5.6 \\
 &  & 300 &  &  &  &  & 87.9/72/15.9 & 87.9/69/8.6 & 89.5/70/9.3 & 89.6/63/5.1 \\
 &  & 400 &  &  &  &  & 88.4/88/10.3 & 88.7/77/7.1 & 89.7/76/8.7 & 91.5/62/5.1 \\
 &  & 500 &  &  &  &  & 88.8/89/10.0 & 88.2/93/4.3 & 87.2/53/11.2 & 90.6/70/4.7 \\
\cmidrule(l){2-11}
 & \multirow{6}{*}{Diffusion Policy} & 50 & \multirow{6}{*}{79.6/106.2} & \multirow{6}{*}{87.8/53.5} & \multirow{6}{*}{76.4/12.3} & \multirow{6}{*}{82.4/6.2} & 83.5/63/39.5 & 83.5/29/38.2 & 81.1/72/4.2 & 81.9/34/4.3 \\
 &  & 100 &  &  &  &  & 83.8/76/26.4 & 84.6/72/15.2 & 79.6/88/2.4 & 80.9/92/1.0 \\
 &  & 200 &  &  &  &  & 86.9/89/12.8 & 86.6/93/4.4 & 86.3/88/2.3 & 87.2/93/0.9 \\
 &  & 300 &  &  &  &  & 87.0/90/11.3 & 89.4/95/3.4 & 86.3/90/2.1 & 89.3/95/0.8 \\
 &  & 400 &  &  &  &  & 89.6/90/11.5 & 90.9/95/3.2 & 89.9/90/2.1 & 90.9/95/0.8 \\
 &  & 500 &  &  &  &  & 90.8/91/10.4 & 91.2/95/3.0 & 91.2/91/2.1 & 89.6/95/0.8 \\
\cmidrule(l){2-11}
 & \multirow{6}{*}{Fast-WAM} & 50 & \multirow{6}{*}{89.8/108.8} & \multirow{6}{*}{92.4/54.9} & \multirow{6}{*}{90.2/44.7} & \multirow{6}{*}{94.8/22.5} & 90.1/65/38.8 & 91.5/19/44.4 & 89.6/60/18.4 & 92.0/21/17.6 \\
 &  & 100 &  &  &  &  & 89.9/71/33.6 & 91.5/22/42.5 & 90.3/68/16.1 & 93.5/17/18.7 \\
 &  & 200 &  &  &  &  & 90.9/75/29.7 & 91.3/21/43.5 & 90.1/71/14.7 & 93.9/11/20.2 \\
 &  & 300 &  &  &  &  & 90.7/78/26.4 & 90.9/21/43.6 & 91.3/70/15.2 & 92.6/25/17.0 \\
 &  & 400 &  &  &  &  & 90.3/76/28.4 & 91.8/17/45.7 & 91.2/72/14.5 & 93.3/20/18.0 \\
 &  & 500 &  &  &  &  & 90.4/80/24.4 & 89.0/32/37.1 & 90.8/77/12.4 & 92.4/29/16.1 \\
\cmidrule(l){2-11}
 & \multirow{6}{*}{Cosmos Policy} & 50 & \multirow{6}{*}{98.6/367.2} & \multirow{6}{*}{n/a} & \multirow{6}{*}{96.6/146.2} & \multirow{6}{*}{n/a} & 98.0/12/335.1 & n/a & 95.7/19/135.4 & n/a \\
 &  & 100 &  &  &  &  & 97.7/29/287.0 & n/a & 94.8/45/121.2 & n/a \\
 &  & 200 &  &  &  &  & 97.1/40/255.8 & n/a & 95.2/51/117.7 & n/a \\
 &  & 300 &  &  &  &  & 97.7/43/249.2 & n/a & 95.7/53/116.4 & n/a \\
 &  & 400 &  &  &  &  & 97.5/48/233.0 & n/a & 95.9/55/115.6 & n/a \\
 &  & 500 &  &  &  &  & 97.6/53/221.5 & n/a & 96.0/66/109.7 & n/a \\
\midrule
\multirow{24}{*}{\rotatebox[origin=c]{90}{MT50}} & \multirow{6}{*}{$\pi_{0.5}$} & 250 & \multirow{6}{*}{56.3/66.3} & \multirow{6}{*}{n/a} & \multirow{6}{*}{59.6/42.6} & \multirow{6}{*}{n/a} & 57.3/25/52.3 & n/a & 60.2/26/34.3 & n/a \\
 &  & 500 &  &  &  &  & 58.1/34/47.2 & n/a & 60.9/35/31.3 & n/a \\
 &  & 1{,}000 &  &  &  &  & 58.5/44/41.7 & n/a & 60.7/43/28.5 & n/a \\
 &  & 1{,}500 &  &  &  &  & 59.5/53/36.6 & n/a & 61.6/47/27.1 & n/a \\
 &  & 2{,}000 &  &  &  &  & 59.5/54/35.7 & n/a & 62.0/49/26.3 & n/a \\
 &  & 2{,}500 &  &  &  &  & 59.7/59/32.9 & n/a & 61.8/55/24.4 & n/a \\
\cmidrule(l){2-11}
 & \multirow{6}{*}{GR00T N1.5} & 250 & \multirow{6}{*}{75.6/16.4} & \multirow{6}{*}{71.7/8.3} & \multirow{6}{*}{75.4/12.3} & \multirow{6}{*}{69.6/6.3} & 75.9/23/13.4 & 71.2/11/7.6 & 75.5/23/10.2 & 69.4/13/5.6 \\
 &  & 500 &  &  &  &  & 75.7/30/12.4 & 71.0/14/7.4 & 75.4/29/9.7 & 69.5/14/5.6 \\
 &  & 1{,}000 &  &  &  &  & 76.0/40/11.1 & 70.9/18/7.1 & 75.3/39/8.8 & 69.5/18/5.4 \\
 &  & 1{,}500 &  &  &  &  & 76.5/43/10.6 & 70.9/20/7.0 & 75.8/41/8.6 & 69.7/20/5.3 \\
 &  & 2{,}000 &  &  &  &  & 77.0/44/10.5 & 71.1/20/7.0 & 75.9/42/8.5 & 69.7/21/5.3 \\
 &  & 2{,}500 &  &  &  &  & 76.9/46/10.3 & 70.9/21/6.9 & 76.0/45/8.2 & 69.8/22/5.2 \\
\cmidrule(l){2-11}
 & \multirow{6}{*}{Diffusion Policy} & 250 & \multirow{6}{*}{61.1/105.7} & \multirow{6}{*}{59.8/53.6} & \multirow{6}{*}{59.5/12.3} & \multirow{6}{*}{58.3/6.2} & 61.5/26/78.6 & 59.4/15/45.7 & 61.2/30/8.9 & 58.8/27/4.7 \\
 &  & 500 &  &  &  &  & 61.2/34/70.0 & 59.1/17/44.7 & 61.3/41/7.6 & 60.1/30/4.5 \\
 &  & 1{,}000 &  &  &  &  & 62.1/44/59.4 & 58.7/21/42.4 & 62.5/51/6.6 & 60.1/34/4.3 \\
 &  & 1{,}500 &  &  &  &  & 62.7/52/51.4 & 58.4/29/38.1 & 63.5/55/6.1 & 60.1/41/3.9 \\
 &  & 2{,}000 &  &  &  &  & 62.9/59/43.8 & 58.5/33/35.8 & 63.7/60/5.5 & 59.9/42/3.8 \\
 &  & 2{,}500 &  &  &  &  & 62.9/65/37.3 & 58.1/33/36.2 & 64.0/62/5.2 & 59.9/43/3.8 \\
\cmidrule(l){2-11}
 & \multirow{6}{*}{Fast-WAM} & 250 & \multirow{6}{*}{56.5/99.2} & \multirow{6}{*}{54.6/50.2} & \multirow{6}{*}{58.6/53.0} & \multirow{6}{*}{56.7/26.8} & 56.8/27/70.6 & 54.9/19/39.7 & 59.1/25/37.1 & 56.8/15/21.8 \\
 &  & 500 &  &  &  &  & 57.0/33/64.4 & 55.3/20/39.4 & 59.5/32/34.7 & 56.9/18/21.1 \\
 &  & 1{,}000 &  &  &  &  & 57.4/40/57.8 & 55.4/25/37.1 & 60.0/37/31.9 & 56.8/24/19.8 \\
 &  & 1{,}500 &  &  &  &  & 57.8/46/52.1 & 55.3/30/34.6 & 60.3/42/29.6 & 57.2/27/18.9 \\
 &  & 2{,}000 &  &  &  &  & 57.4/51/47.9 & 55.5/31/34.3 & 60.3/45/28.3 & 57.0/30/18.2 \\
 &  & 2{,}500 &  &  &  &  & 57.7/52/46.8 & 55.3/33/33.1 & 60.6/44/29.3 & 57.5/30/18.4 \\
\bottomrule
\end{tabular*}
\vspace{-8.0pt}
\caption{End-to-end results for every library size. Library gives the episodes collected for each LIBERO suite and for all 50 MT50 tasks. Pure inference does not use the library. Pure inference cells give success rate (\%) / latency (ms), and \name cells give success rate (\%) / hit rate (\%) / latency (ms). n/a marks a policy that already executes its whole chunk. \name cells average three libraries collected with different seeds.}
\label{tab:main-all}
\end{table*}

\begin{figure}[!htb]
\centering
\includegraphics[width=3.25in]{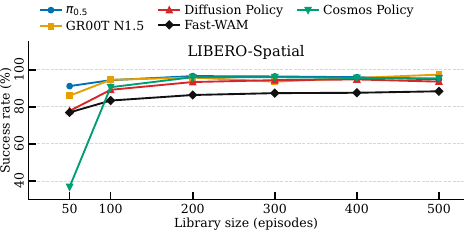}
\vspace{-10.3pt}
\caption{Cache alone of Figure~\ref{fig:libsize} on LIBERO-Spatial. The libraries keep every collected episode, including failed ones.}
\label{fig:libsize-spatial}
\end{figure}

\paragraph{Cache without refresh.}
In Tables~\ref{tab:abl-libero} and~\ref{tab:abl-mt50}, the Cache and +Corr. columns never call the policy.
With the corrector, the mean success rate over policies reaches that of pure inference on LIBERO-Spatial at 200 episodes (93.6\% against 93.5\%) and stays flat beyond (Figure~\ref{fig:libsize-spatial}).
On LIBERO-Long it rises from 44.6\% to 75.3\% between 50 and 500 episodes, still below 87.8\% for pure inference, and on MT50 it rises from 40.9\% to 55.9\% but gains only 1.3 points beyond 1{,}500 episodes.

\paragraph{Components.}
Averaged over policies and both LIBERO suites, the corrector adds 7.4 points to the cache at 50 episodes and 1.9 points at 500, and on MT50 3.4 points at 250 episodes and 1.9 points at 2{,}500.
The stagnation check adds 15.8 points on average and up to 35.0 points on LIBERO-Long while calling the policy at 8\% to 25\% of the decisions, but only 2.1 and 1.2 points on LIBERO-Spatial and MT50.
On LIBERO-Long, the two together beat the better of the two alone in 50 of 54 settings, and with the default horizon the corrector adds 5.0 points alone, the stagnation check 15.8 points, and both 19.4 points.

\begin{table*}[p]
\centering
\scriptsize
\renewcommand{\arraystretch}{0.78}
\setlength{\tabcolsep}{3pt}
\setlength{\aboverulesep}{0.5pt}\setlength{\belowrulesep}{0.8pt}
\begin{tabular*}{\textwidth}{@{\extracolsep{\fill}}llllcccccccc@{}}
\toprule
& & & & \multicolumn{4}{c}{Default execution horizon} & \multicolumn{4}{c}{Expanded execution horizon} \\
\cmidrule(lr){5-8}\cmidrule(l){9-12}
& Policy & Library & Pure & Cache & +Corr. & +Stag. & +Both & Cache & +Corr. & +Stag. & +Both \\
\midrule
\multirow{30}{*}{\rotatebox[origin=c]{90}{LIBERO-Spatial}} & \multirow{6}{*}{$\pi_{0.5}$} & 50 & \multirow{6}{*}{96.6} & 89.8/100/10.3 & 91.2/100/10.3 & 91.0/90/13.2 & 93.2/89/13.5 & 80.2/100/5.2 & 86.4/100/5.2 & 84.4/94/6.2 & 89.4/94/6.2 \\
 &  & 100 &  & 93.4/100/10.3 & 94.4/100/10.3 & 96.6/94/12.2 & 97.0/94/12.0 & 92.8/100/5.2 & 95.0/100/5.3 & 95.0/96/5.8 & 96.4/97/5.7 \\
 &  & 200 &  & 92.8/100/10.3 & 96.6/100/10.3 & 94.8/94/12.1 & 96.8/95/11.9 & 93.4/100/5.2 & 96.6/100/5.3 & 94.2/97/5.7 & 96.6/98/5.6 \\
 &  & 300 &  & 94.2/100/10.3 & 96.4/100/10.3 & 95.8/94/12.0 & 96.8/95/11.8 & 95.8/100/5.3 & 98.0/100/5.3 & 96.0/98/5.6 & 97.2/98/5.6 \\
 &  & 400 &  & 93.2/100/10.3 & 96.0/100/10.3 & 95.8/94/12.0 & 96.4/95/11.8 & 96.4/100/5.3 & 98.2/100/5.3 & 97.0/98/5.7 & 97.2/98/5.6 \\
 &  & 500 &  & 95.0/100/10.3 & 94.8/100/10.3 & 95.0/94/12.0 & 97.6/95/11.7 & 96.6/100/5.3 & 97.0/100/5.3 & 96.2/98/5.6 & 96.6/98/5.6 \\
\cmidrule(l){2-12}
 & \multirow{6}{*}{GR00T N1.5} & 50 & \multirow{6}{*}{93.4} & 85.6/100/6.1 & 86.0/100/6.1 & 87.0/91/7.1 & 87.4/91/7.1 & 83.2/100/3.1 & 89.6/100/3.1 & 84.8/95/3.4 & 88.4/95/3.4 \\
 &  & 100 &  & 91.8/100/6.1 & 94.6/100/6.1 & 91.8/93/6.9 & 95.0/94/6.8 & 93.2/100/3.1 & 94.4/100/3.1 & 94.2/96/3.3 & 95.0/97/3.3 \\
 &  & 200 &  & 93.8/100/6.1 & 95.8/100/6.1 & 95.4/94/6.8 & 95.4/94/6.7 & 95.2/100/3.1 & 95.6/100/3.1 & 95.0/96/3.3 & 95.6/97/3.3 \\
 &  & 300 &  & 94.4/100/6.1 & 93.6/100/6.1 & 93.6/94/6.8 & 96.0/95/6.6 & 95.6/100/3.1 & 95.2/100/3.1 & 94.6/97/3.3 & 94.8/97/3.3 \\
 &  & 400 &  & 96.6/100/6.1 & 95.6/100/6.1 & 97.0/95/6.6 & 92.8/94/6.8 & 96.6/100/3.1 & 96.2/100/3.1 & 94.8/97/3.3 & 94.8/97/3.3 \\
 &  & 500 &  & 95.8/100/6.1 & 97.4/100/6.1 & 93.4/94/6.8 & 95.2/94/6.7 & 97.0/100/3.1 & 95.6/100/3.1 & 95.8/97/3.3 & 95.4/97/3.3 \\
\cmidrule(l){2-12}
 & \multirow{6}{*}{Diffusion Policy} & 50 & \multirow{6}{*}{90.0} & 72.2/100/1.0 & 77.8/100/1.0 & 83.8/86/2.6 & 85.4/86/2.6 & 69.6/100/0.5 & 77.8/100/0.5 & 76.2/88/1.3 & 84.0/91/1.1 \\
 &  & 100 &  & 86.6/100/1.0 & 89.2/100/1.0 & 91.4/89/2.3 & 90.2/88/2.4 & 83.4/100/0.5 & 92.6/100/0.5 & 86.6/91/1.1 & 92.0/93/0.9 \\
 &  & 200 &  & 88.4/100/1.0 & 93.4/100/1.0 & 92.2/90/2.1 & 93.4/91/2.1 & 87.8/100/0.5 & 95.2/100/0.5 & 91.8/93/0.9 & 94.0/94/0.9 \\
 &  & 300 &  & 91.8/100/1.0 & 94.4/100/1.0 & 92.8/91/2.0 & 93.0/92/2.0 & 90.6/100/0.5 & 94.2/100/0.5 & 91.8/94/0.9 & 94.2/95/0.8 \\
 &  & 400 &  & 93.0/100/1.0 & 94.8/100/1.0 & 92.0/92/2.0 & 93.8/92/1.9 & 91.6/100/0.5 & 93.0/100/0.5 & 91.0/94/0.9 & 93.4/95/0.8 \\
 &  & 500 &  & 94.6/100/1.0 & 93.6/100/1.0 & 93.0/92/1.9 & 94.0/92/1.9 & 91.4/100/0.5 & 93.8/100/0.5 & 92.0/94/0.9 & 94.0/95/0.8 \\
\cmidrule(l){2-12}
 & \multirow{6}{*}{Fast-WAM} & 50 & \multirow{6}{*}{91.0} & 78.6/100/6.7 & 77.0/100/6.7 & 84.2/92/8.7 & 85.6/91/9.0 & 85.0/100/3.4 & 84.6/100/3.4 & 85.6/94/4.4 & 87.0/94/4.4 \\
 &  & 100 &  & 83.2/100/6.7 & 83.4/100/6.7 & 87.6/93/8.5 & 87.2/93/8.5 & 85.2/100/3.4 & 88.0/100/3.4 & 85.2/96/4.0 & 87.4/96/4.0 \\
 &  & 200 &  & 84.4/100/6.7 & 86.4/100/6.7 & 86.2/91/9.1 & 87.8/91/9.0 & 88.8/100/3.4 & 89.2/100/3.4 & 88.6/94/4.3 & 88.6/94/4.3 \\
 &  & 300 &  & 84.8/100/6.7 & 87.4/100/6.7 & 87.0/91/9.1 & 89.2/92/8.8 & 88.4/100/3.4 & 89.0/100/3.4 & 88.8/94/4.2 & 88.4/94/4.2 \\
 &  & 400 &  & 85.4/100/6.7 & 87.6/100/6.7 & 88.2/91/9.1 & 88.8/91/8.9 & 88.6/100/3.4 & 89.2/100/3.4 & 89.8/93/4.3 & 89.2/94/4.3 \\
 &  & 500 &  & 87.8/100/6.7 & 88.4/100/6.7 & 89.2/92/8.9 & 90.2/91/8.9 & 89.2/100/3.4 & 89.4/100/3.4 & 89.2/94/4.3 & 88.2/94/4.3 \\
\cmidrule(l){2-12}
 & \multirow{6}{*}{Cosmos Policy} & 50 & \multirow{6}{*}{96.4} & 11.2/100/90.0 & 36.8/100/90.5 & 25.6/76/103.4 & 45.2/81/101.6 & n/a & n/a & n/a & n/a \\
 &  & 100 &  & 85.6/100/92.2 & 90.6/100/93.1 & 86.6/93/96.2 & 91.8/95/96.0 & n/a & n/a & n/a & n/a \\
 &  & 200 &  & 93.6/100/92.9 & 96.0/100/93.9 & 92.8/96/95.1 & 94.8/97/95.7 & n/a & n/a & n/a & n/a \\
 &  & 300 &  & 95.2/100/93.6 & 96.0/100/93.6 & 94.6/96/95.8 & 96.2/97/95.1 & n/a & n/a & n/a & n/a \\
 &  & 400 &  & 94.0/100/93.7 & 95.6/100/93.6 & 94.6/96/96.0 & 95.2/97/95.6 & n/a & n/a & n/a & n/a \\
 &  & 500 &  & 95.0/100/93.4 & 95.4/100/93.3 & 95.8/97/95.4 & 95.8/97/95.0 & n/a & n/a & n/a & n/a \\
\midrule
\multirow{30}{*}{\rotatebox[origin=c]{90}{LIBERO-Long}} & \multirow{6}{*}{$\pi_{0.5}$} & 50 & \multirow{6}{*}{85.0} & 55.0/100/10.3 & 63.4/100/10.3 & 70.6/85/14.9 & 74.4/86/14.6 & 63.0/100/5.2 & 70.8/100/5.2 & 72.0/90/6.7 & 76.6/91/6.5 \\
 &  & 100 &  & 67.0/100/10.3 & 69.8/100/10.3 & 78.8/88/14.0 & 81.4/89/13.7 & 70.8/100/5.2 & 77.8/100/5.2 & 75.4/90/6.6 & 80.0/92/6.5 \\
 &  & 200 &  & 73.8/100/10.3 & 76.4/100/10.3 & 83.0/89/13.6 & 85.6/91/13.2 & 77.0/100/5.2 & 84.6/100/5.2 & 79.8/92/6.4 & 86.2/94/6.1 \\
 &  & 300 &  & 74.0/100/10.3 & 80.0/100/10.3 & 84.0/89/13.6 & 88.0/91/13.1 & 81.0/100/5.2 & 86.2/100/5.2 & 85.2/93/6.3 & 88.4/94/6.1 \\
 &  & 400 &  & 78.0/100/10.3 & 80.2/100/10.3 & 85.6/90/13.5 & 87.0/90/13.3 & 81.8/100/5.2 & 86.4/100/5.2 & 85.8/93/6.2 & 92.0/96/5.8 \\
 &  & 500 &  & 76.8/100/10.3 & 83.2/100/10.3 & 84.8/90/13.2 & 89.6/92/12.9 & 84.6/100/5.2 & 89.0/100/5.2 & 87.4/94/6.1 & 89.6/94/6.0 \\
\cmidrule(l){2-12}
 & \multirow{6}{*}{GR00T N1.5} & 50 & \multirow{6}{*}{86.2} & 50.2/100/6.1 & 52.2/100/6.1 & 62.0/86/7.7 & 64.4/87/7.5 & 58.4/100/3.1 & 65.2/100/3.1 & 62.0/88/3.7 & 67.0/89/3.7 \\
 &  & 100 &  & 53.2/100/6.1 & 50.6/100/6.1 & 71.2/86/7.7 & 75.2/86/7.6 & 66.0/100/3.1 & 65.8/100/3.1 & 72.0/89/3.7 & 73.8/89/3.6 \\
 &  & 200 &  & 62.4/100/6.1 & 66.0/100/6.1 & 80.0/84/7.8 & 81.6/85/7.7 & 75.8/100/3.1 & 75.2/100/3.1 & 77.6/89/3.7 & 79.2/90/3.6 \\
 &  & 300 &  & 63.8/100/6.1 & 69.0/100/6.1 & 80.0/86/7.6 & 84.0/87/7.5 & 75.2/100/3.1 & 80.6/100/3.1 & 80.0/92/3.5 & 83.0/93/3.5 \\
 &  & 400 &  & 65.0/100/6.1 & 72.4/100/6.1 & 85.6/87/7.6 & 87.4/88/7.5 & 78.8/100/3.1 & 79.6/100/3.1 & 82.0/90/3.6 & 83.4/91/3.6 \\
 &  & 500 &  & 72.6/100/6.1 & 74.8/100/6.1 & 86.4/88/7.4 & 84.2/88/7.4 & 82.8/100/3.1 & 86.6/100/3.1 & 85.6/92/3.5 & 88.2/93/3.5 \\
\cmidrule(l){2-12}
 & \multirow{6}{*}{Diffusion Policy} & 50 & \multirow{6}{*}{79.6} & 24.2/100/1.0 & 32.4/100/1.0 & 50.6/84/2.8 & 59.4/86/2.5 & 47.0/100/0.5 & 54.6/100/0.5 & 50.8/87/1.3 & 60.6/89/1.2 \\
 &  & 100 &  & 46.6/100/1.0 & 50.4/100/1.0 & 73.6/87/2.5 & 76.4/88/2.4 & 64.4/100/0.5 & 71.2/100/0.5 & 71.0/89/1.2 & 80.6/92/1.0 \\
 &  & 200 &  & 48.8/100/1.0 & 51.0/100/1.0 & 83.8/87/2.5 & 80.4/87/2.5 & 76.8/100/0.5 & 78.6/100/0.5 & 82.8/89/1.2 & 85.8/91/1.0 \\
 &  & 300 &  & 57.8/100/1.0 & 60.6/100/1.0 & 82.2/88/2.4 & 84.8/88/2.3 & 82.8/100/0.5 & 81.4/100/0.5 & 87.0/93/0.9 & 86.2/93/1.0 \\
 &  & 400 &  & 61.2/100/1.0 & 63.6/100/1.0 & 85.8/89/2.3 & 87.6/89/2.2 & 81.8/100/0.5 & 84.4/100/0.5 & 85.6/92/1.0 & 87.6/92/1.0 \\
 &  & 500 &  & 64.8/100/1.0 & 68.2/100/1.0 & 87.4/90/2.2 & 88.8/90/2.1 & 84.8/100/0.5 & 87.2/100/0.5 & 87.4/93/0.9 & 88.6/94/0.9 \\
\cmidrule(l){2-12}
 & \multirow{6}{*}{Fast-WAM} & 50 & \multirow{6}{*}{89.8} & 39.0/100/6.7 & 45.0/100/6.7 & 66.2/81/11.4 & 71.8/82/11.2 & 52.8/100/3.3 & 60.6/100/3.3 & 63.2/83/5.8 & 64.6/84/5.7 \\
 &  & 100 &  & 51.0/100/6.7 & 53.8/100/6.7 & 71.8/85/10.3 & 72.2/86/10.1 & 64.4/100/3.4 & 66.8/100/3.4 & 66.4/86/5.2 & 69.4/87/5.1 \\
 &  & 200 &  & 56.6/100/6.7 & 63.0/100/6.7 & 74.8/84/10.8 & 79.8/85/10.5 & 70.2/100/3.4 & 72.0/100/3.4 & 70.4/87/5.1 & 75.0/88/5.0 \\
 &  & 300 &  & 62.6/100/6.7 & 67.2/100/6.7 & 77.2/84/10.7 & 81.2/86/10.3 & 72.0/100/3.4 & 76.0/100/3.4 & 75.6/88/5.0 & 79.2/89/4.8 \\
 &  & 400 &  & 64.8/100/6.7 & 71.2/100/6.7 & 79.6/85/10.5 & 84.8/87/10.0 & 73.8/100/3.4 & 77.0/100/3.4 & 79.2/90/4.8 & 81.8/90/4.7 \\
 &  & 500 &  & 69.0/100/6.7 & 70.4/100/6.7 & 81.4/85/10.4 & 83.4/86/10.2 & 75.6/100/3.4 & 76.8/100/3.4 & 78.2/89/4.9 & 80.0/90/4.7 \\
\cmidrule(l){2-12}
 & \multirow{6}{*}{Cosmos Policy} & 50 & \multirow{6}{*}{98.6} & 12.0/100/89.4 & 29.8/100/89.6 & 22.4/75/103.4 & 38.4/78/101.8 & n/a & n/a & n/a & n/a \\
 &  & 100 &  & 45.0/100/89.7 & 60.2/100/89.8 & 54.0/85/98.0 & 65.2/88/96.7 & n/a & n/a & n/a & n/a \\
 &  & 200 &  & 66.0/100/90.0 & 71.4/100/90.0 & 74.4/89/96.1 & 76.4/91/95.2 & n/a & n/a & n/a & n/a \\
 &  & 300 &  & 70.6/100/90.1 & 75.6/100/90.1 & 77.0/90/95.8 & 83.6/92/94.6 & n/a & n/a & n/a & n/a \\
 &  & 400 &  & 77.2/100/90.3 & 81.4/100/90.2 & 84.4/92/94.7 & 84.0/92/94.5 & n/a & n/a & n/a & n/a \\
 &  & 500 &  & 75.2/100/90.2 & 79.8/100/90.2 & 80.0/91/95.4 & 85.6/93/94.3 & n/a & n/a & n/a & n/a \\
\bottomrule
\end{tabular*}
\vspace{-7.7pt}
\caption{Cache component ablation on LIBERO for every library size, without periodic refresh. Cache retrieves and fuses library chunks, +Corr. adds the corrector, +Stag. adds the stagnation check, and +Both adds both. Cells give success rate (\%) / hit rate (\%) / latency (ms), and Pure gives the success rate of pure inference with the default settings. Policy calls use reduced steps. n/a marks a policy without an expanded horizon. The libraries keep every collected episode, including failed ones.}
\label{tab:abl-libero}
\end{table*}

\paragraph{Expanded horizon.}
On LIBERO-Long, the expanded horizon raises the success rate of the cache in all 24 settings, by 12.6 points on average and by up to 28.0 points, since the robot retrieves half as often.
Once the stagnation check is on, the two horizons end within one point of each other on average.

\paragraph{Executing past the end of a chunk.}
For $\pi_{0.5}$ on LIBERO-Long with libraries of 200 episodes and no policy calls, executing 10 actions per retrieval, the length of a chunk, succeeds in 84.9\%, against 78.1\% with 5 actions, while following the retrieved episode to 15, 20, 30, and 40 actions lowers it to 77.1\%, 74.3\%, 66.5\%, and 59.8\%, even with the corrector refit for each length.
\name therefore never executes past the end of a chunk.

\begin{table*}[!htb]
\centering
\scriptsize
\renewcommand{\arraystretch}{0.78}
\setlength{\tabcolsep}{3pt}
\setlength{\aboverulesep}{0.5pt}\setlength{\belowrulesep}{0.8pt}
\begin{tabular*}{\textwidth}{@{\extracolsep{\fill}}llllcccccccc@{}}
\toprule
& & & & \multicolumn{4}{c}{Default execution horizon} & \multicolumn{4}{c}{Expanded execution horizon} \\
\cmidrule(lr){5-8}\cmidrule(l){9-12}
& Policy & Library & Pure & Cache & +Corr. & +Stag. & +Both & Cache & +Corr. & +Stag. & +Both \\
\midrule
\multirow{24}{*}{\rotatebox[origin=c]{90}{MT50}} & \multirow{6}{*}{$\pi_{0.5}$} & 250 & \multirow{6}{*}{56.3} & 34.8 & 39.1 & 35.9 & 40.0 & n/a & n/a & n/a & n/a \\
 &  & 500 &  & 41.6 & 46.3 & 42.2 & 46.5 & n/a & n/a & n/a & n/a \\
 &  & 1{,}000 &  & 47.4 & 50.5 & 47.5 & 51.1 & n/a & n/a & n/a & n/a \\
 &  & 1{,}500 &  & 50.1 & 52.4 & 50.6 & 53.0 & n/a & n/a & n/a & n/a \\
 &  & 2{,}000 &  & 50.6 & 52.2 & 50.8 & 53.0 & n/a & n/a & n/a & n/a \\
 &  & 2{,}500 &  & 50.4 & 52.8 & 50.8 & 53.1 & n/a & n/a & n/a & n/a \\
\cmidrule(l){2-12}
 & \multirow{6}{*}{GR00T N1.5} & 250 & \multirow{6}{*}{75.6} & 36.1 & 40.7 & 39.2 & 42.5 & 36.8 & 38.6 & 37.7 & 39.8 \\
 &  & 500 &  & 46.1 & 49.7 & 48.4 & 51.4 & 41.3 & 44.6 & 42.4 & 45.6 \\
 &  & 1{,}000 &  & 52.4 & 55.7 & 54.1 & 56.7 & 50.2 & 52.3 & 51.1 & 53.2 \\
 &  & 1{,}500 &  & 56.2 & 59.1 & 57.6 & 60.3 & 55.6 & 56.2 & 55.8 & 56.5 \\
 &  & 2{,}000 &  & 57.4 & 60.2 & 59.2 & 61.5 & 57.7 & 59.7 & 58.3 & 59.9 \\
 &  & 2{,}500 &  & 59.3 & 62.1 & 61.0 & 63.0 & 58.2 & 59.9 & 58.6 & 59.8 \\
\cmidrule(l){2-12}
 & \multirow{6}{*}{Diffusion Policy} & 250 & \multirow{6}{*}{61.1} & 43.4 & 46.0 & 44.9 & 46.6 & 39.0 & 40.6 & 39.1 & 41.1 \\
 &  & 500 &  & 49.4 & 51.2 & 50.9 & 52.5 & 46.2 & 46.7 & 47.0 & 48.0 \\
 &  & 1{,}000 &  & 54.9 & 56.8 & 55.7 & 57.3 & 51.2 & 52.2 & 51.6 & 52.0 \\
 &  & 1{,}500 &  & 57.2 & 58.2 & 57.4 & 58.3 & 54.8 & 55.6 & 55.0 & 55.4 \\
 &  & 2{,}000 &  & 57.4 & 58.1 & 57.8 & 58.7 & 56.5 & 56.4 & 56.5 & 56.2 \\
 &  & 2{,}500 &  & 57.8 & 58.5 & 58.8 & 59.4 & 56.6 & 57.3 & 56.2 & 56.7 \\
\cmidrule(l){2-12}
 & \multirow{6}{*}{Fast-WAM} & 250 & \multirow{6}{*}{56.5} & 35.8 & 37.9 & 38.2 & 39.4 & 32.2 & 36.8 & 33.9 & 38.1 \\
 &  & 500 &  & 39.8 & 42.2 & 41.2 & 43.7 & 40.1 & 42.2 & 40.2 & 42.1 \\
 &  & 1{,}000 &  & 44.7 & 46.9 & 46.4 & 47.8 & 43.5 & 45.2 & 43.8 & 44.7 \\
 &  & 1{,}500 &  & 46.2 & 48.7 & 48.2 & 49.8 & 46.2 & 47.2 & 46.2 & 47.6 \\
 &  & 2{,}000 &  & 47.2 & 49.2 & 48.2 & 50.3 & 47.6 & 48.8 & 47.4 & 48.7 \\
 &  & 2{,}500 &  & 48.8 & 50.3 & 49.6 & 51.7 & 47.8 & 49.0 & 48.0 & 49.5 \\
\bottomrule
\end{tabular*}
\vspace{-7.7pt}
\caption{Cache component ablation on MT50 for every library size, as in Table~\ref{tab:abl-libero}. Cells give the mean success rate over tasks (\%), and Pure gives that of pure inference with the default settings. n/a marks a policy without an expanded horizon. The libraries keep every collected episode, including failed ones.}
\label{tab:abl-mt50}
\end{table*}

\paragraph{One interval for all tasks.}
Per-task intervals lower the latency in every row of Table~\ref{tab:abl-budget} and never fall more than 1.4 points below pure inference, whereas the single interval with the highest success rate falls 5.9 to 19.2 points below pure inference on MT50 and for Fast-WAM on LIBERO-Long.

\begin{table}[!htb]
\centering
\scriptsize
\setlength{\tabcolsep}{2.5pt}
\begin{tabular}{@{}lcccc@{}}
\toprule
& & & \multicolumn{2}{c}{One interval} \\
\cmidrule(l){4-5}
Policy & Pure & Per task & Closest & Best \\
\midrule
\multicolumn{5}{@{}l}{\emph{LIBERO-Spatial}} \\
$\pi_{0.5}$ & 98.4/20.9 & 97.4/7.5 & 95.4/5.7 & 98.4/14.0 \\
Cosmos Policy & 96.6/151.3 & 97.3/104.0 & 95.6/106.1 & 97.2/111.3 \\
\midrule
\multicolumn{5}{@{}l}{\emph{LIBERO-Long}} \\
$\pi_{0.5}$ & 91.8/20.6 & 93.7/10.1 & 93.0/11.4 & 93.0/11.4 \\
GR00T N1.5 & 88.2/8.5 & 89.3/5.6 & 85.4/6.2 & 91.0/7.8 \\
Fast-WAM & 94.8/22.5 & 93.9/20.2 & 75.6/10.0 & 75.6/10.0 \\
Cosmos Policy & 96.6/146.2 & 95.2/117.7 & 92.2/113.6 & 95.6/118.7 \\
\midrule
\multicolumn{5}{@{}l}{\emph{MT50}} \\
GR00T N1.5 & 69.6/6.3 & 69.5/5.4 & 57.4/4.0 & 57.4/4.0 \\
Diffusion Policy & 58.3/6.2 & 60.1/4.3 & 51.3/3.5 & 52.4/2.0 \\
Fast-WAM & 56.7/26.8 & 56.8/19.8 & 49.4/12.3 & 49.4/12.3 \\
\bottomrule
\end{tabular}
\vspace{-3.0pt}
\caption{Refresh intervals chosen per task against one interval for all tasks, with libraries of 20 episodes per task, reduced steps, and the expanded horizon (reduced steps only for Cosmos Policy). Closest is the single interval whose latency is closest to that of per-task intervals, and Best is the single interval with the highest success rate. Cells give success rate (\%) / latency (ms).}
\label{tab:abl-budget}
\end{table}

\FloatBarrier

\section{Choosing the Refresh Level of Each Task}\label{app:knob}
Each task chooses its refresh level online, on the episodes it serves, with the ladder of Section~\ref{sec:correct}: it starts at pure inference, runs 50 episodes at a level, steps down while the task succeeds in at most 2 fewer of them than at pure inference, and keeps the last level that passed.
This appendix reports what choosing online costs and whether a chosen level holds on initial states it was not chosen on.
Every other quantity that depends on the data is fitted on the library alone (Table~\ref{tab:fit}), and no library initial state appears in testing (Appendix~\ref{app:library}).

\begin{figure}[!htb]
\centering
\includegraphics[width=3.25in]{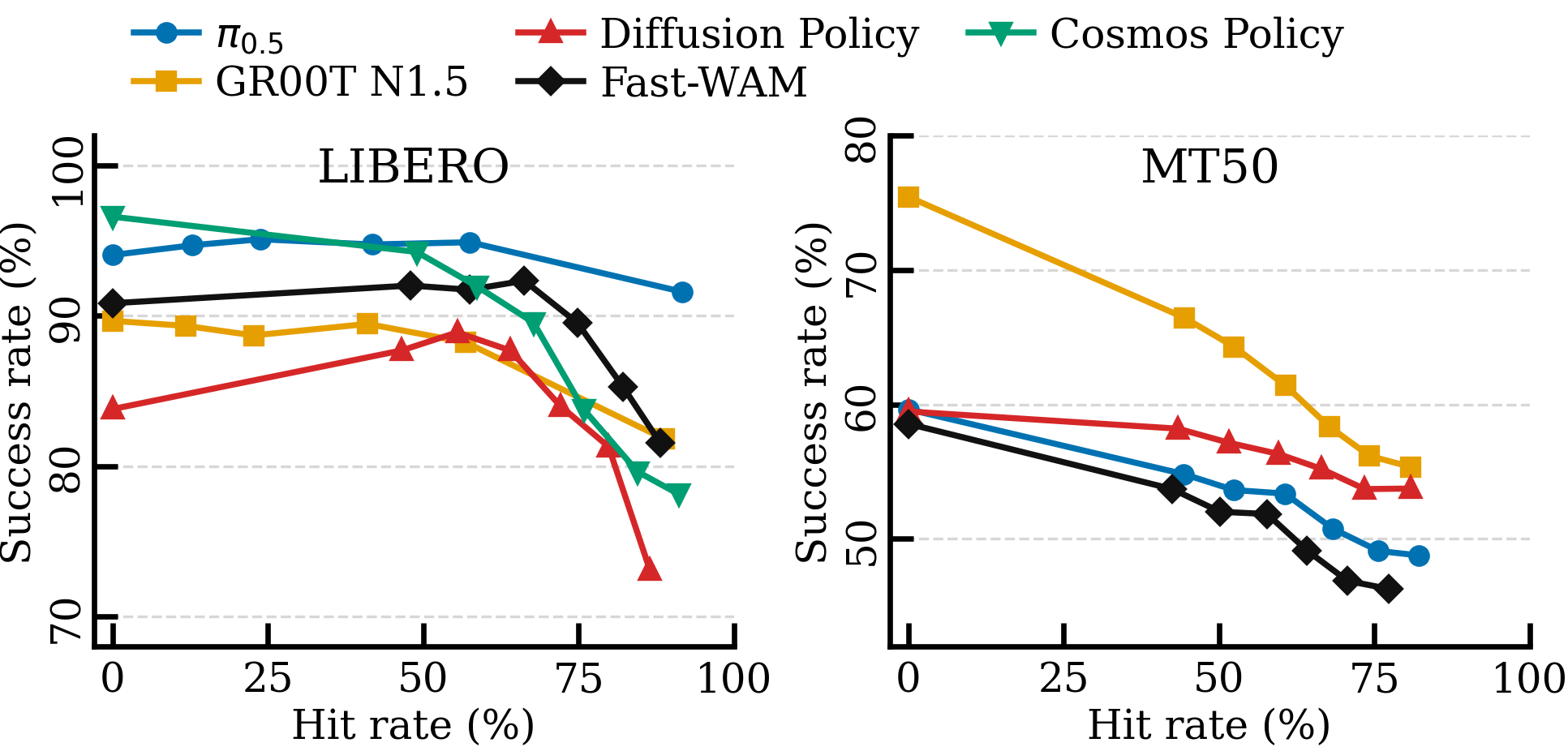}
\vspace{-10.0pt}
\caption{Success rate falls as the ladder lets the cache serve more decisions. Mean success rate against mean hit rate at each level, from pure inference (hit rate 0) to the level where only the stagnation check calls the policy, averaged over the tasks and library sizes of one library seed, with reduced steps and the default horizon. LIBERO averages both suites.}
\label{fig:knob}
\end{figure}

\paragraph{Success rate falls along the ladder.}
Figure~\ref{fig:knob} follows the ladder for every policy.
On LIBERO, the success rate stays near or above that of pure inference while the hit rate is low and drops once the hit rate passes a point that differs across policies, and on MT50 it falls from the first level on.
Across the 32{,}000 steps of the 5{,}410 task ladders in which every level was run, a more aggressive level succeeds in more than 2 more episodes than the level before it in only 10.5\% of the steps, against 16.4\% from binomial noise alone if no step changed the success rate.

\paragraph{Choosing online costs about five failures per task, once.}
In the main results of Section~\ref{sec:e2e}, 525 of the 900 tasks reach a level that fails, and choosing the levels costs about five failures per task beyond those of the level each task keeps, 3.0\% of the episodes run while choosing.
This is an upper bound, since we run every failing level for all 50 episodes, while a deployment can stop a level as soon as it can no longer pass.
Binary search over the ladder would try fewer levels per task (2.65 instead of 3.33), but it tests aggressive levels first and fails twice as many extra episodes (10.8 against 5.4) for nearly the same success rate at the chosen level (70.9\% against 70.8\%), so \name steps down.

\paragraph{The chosen levels hold on unseen initial states.}
Each task of the main results also chooses its level on half of its initial states and is evaluated on the other half, and the halves then swap.
With the same tolerance of 4 points, now 1 of 25 episodes, the success rate on the held-out halves is 1.4 points lower on average than in the main results, below the tolerance, at nearly the same latency.

\FloatBarrier

\section{Inside a Cached Episode}\label{app:inside}
To see where the cache errs, we serve LIBERO-Long with libraries of 200 episodes, with the settings of the stagnation-check row of Table~\ref{tab:ladder}, and at every decision also run the policy in the background without executing its chunk.
The action error of a chunk is the root mean square of its difference from the policy's chunk over the executed control steps, with each motion channel divided by its standard deviation in the library.

\begin{figure}[!htb]
\begin{minipage}[b]{0.49\columnwidth}
\centering
\footnotesize
\setlength{\tabcolsep}{2pt}
\begin{tabular*}{\linewidth}{@{\extracolsep{\fill}}lcccccc@{}}
\toprule
& \multicolumn{3}{c}{$\pi_{0.5}$} & \multicolumn{3}{c}{Diffusion Policy} \\
\cmidrule(lr){2-4}\cmidrule(l){5-7}
Chunk & All & Near & Far & All & Near & Far \\
\midrule
Nearest entry & 0.31 & 0.24 & 0.48 & 0.42 & 0.34 & 0.52 \\
Fused & 0.24 & 0.20 & 0.39 & 0.33 & 0.28 & 0.43 \\
Corrected & 0.23 & 0.19 & 0.39 & 0.31 & 0.26 & 0.41 \\
\bottomrule
\end{tabular*}
\vspace{-7.7pt}
\captionof{table}{Fusion and correction lower the action error at every distance from the library. Median error against the policy's own chunk over all decisions, and over the fifth of decisions nearest to (Near) and farthest from (Far) the library.}
\label{tab:inside}
\end{minipage}\hfill
\begin{minipage}[b]{0.49\columnwidth}
\centering
\includegraphics[width=\linewidth]{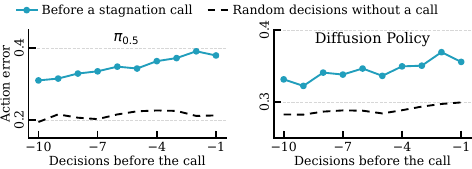}
\vspace{-23.0pt}
\captionof{figure}{The action error builds up before the stagnation check calls the policy. Median error of the served chunk over the ten decisions before each stagnation call and at random decisions without a call.}
\label{fig:inside}
\end{minipage}
\end{figure}

\paragraph{Fusion and correction cut the action mismatch at every distance.}
Replaying the nearest entry gives a median error of 0.31 for $\pi_{0.5}$ and 0.42 for Diffusion Policy (Table~\ref{tab:inside}).
Fusing the 16 nearest entries lowers it to 0.24 and 0.33, and the corrector to 0.23 and 0.31, about a quarter below the nearest entry.
The error grows with the distance of the query from the library, and fusion and correction lower it in every quintile of that distance.
Failed episodes sit farther from the library, with a median distance to the nearest entry of 1.19 for $\pi_{0.5}$ and 1.28 for Diffusion Policy, against 0.89 in successful episodes.

\paragraph{Errors build up before the stagnation check calls the policy.}
Over the ten decisions before a stagnation call, the median error of the served chunk rises from 0.31 to 0.38 for $\pi_{0.5}$ and from 0.33 to 0.36 for Diffusion Policy, while random decisions without a call stay near 0.21 and 0.29 (Figure~\ref{fig:inside}).
Within a run of consecutive cache hits, the error is higher at the sixteenth hit than at the second, 0.23 against 0.20 for $\pi_{0.5}$ and 0.35 against 0.29 for Diffusion Policy.

\begin{figure}[!htb]
\centering
\includegraphics[width=\columnwidth]{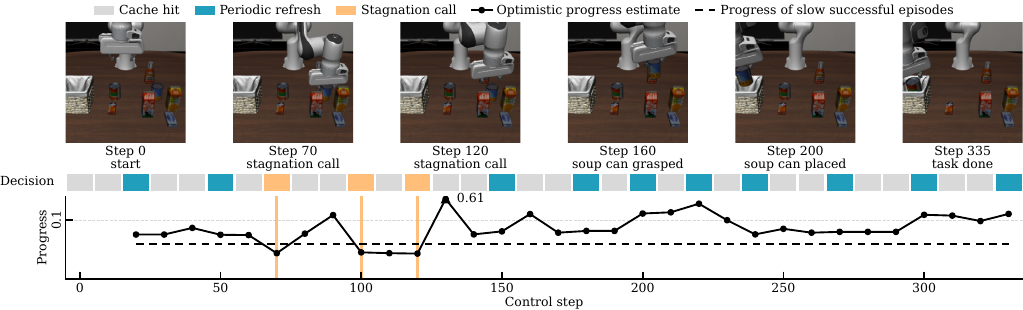}
\vspace{-22.0pt}
\caption{The stagnation check calls the policy when the robot stops advancing, and the robot then finishes the task. Third-person frames, the kind of each decision, and the progress estimate for one LIBERO-Long episode of $\pi_{0.5}$.}
\label{fig:timeline}
\end{figure}

\paragraph{One episode.}
Figure~\ref{fig:timeline} follows one episode of $\pi_{0.5}$ on LIBERO-Long from the main results.
Between control steps 60 and 120, the robot reaches for the soup can again and again without grasping it, and the optimistic progress estimate falls below the progress of slow successful episodes.
The stagnation check calls the policy three times, after which the robot grasps the can, places it in the basket, and finishes the task.
Periodic refresh calls the policy at other times, and the remaining decisions are cache hits.

\FloatBarrier

\section{Environment Perturbations and Out-of-Distribution Scenes}\label{app:ood}
The main experiments test new initial states of scenes that the library has seen.
Here we test scenes it has never seen, on LIBERO-plus \cite{fei2025liberoplusindepthrobustnessanalysis}, which perturbs LIBERO along seven axes at five levels each: lighting, background texture, camera viewpoint, paraphrased instructions, sensor noise on the third-person image, added objects, and the initial joint configuration of the robot.
We use its 2,402 LIBERO-Spatial and 2,519 LIBERO-Long variants with the official protocol of one episode per variant, and run $\pi_{0.5}$ with ten denoising steps and an execution horizon of five actions.
The libraries are those of the main experiments, built on the original scenes and used unchanged.
The official prompts of LIBERO-plus contain file-name tags, so the cache and $\pi_{0.5}$ both receive the clean task instruction; with the official prompts, $\pi_{0.5}$ succeeds in 80.3\% instead of 80.6\% of all variants.

\paragraph{Without adaptation, perturbations of the views and of the robot hurt the cache most.}
Table~\ref{tab:ood} compares $\pi_{0.5}$ with the cache and corrector alone, which never call the policy.
Over all variants they succeed in 64.3\% against 80.6\% for $\pi_{0.5}$.
The loss concentrates on camera viewpoint (29.6\% against 65.2\%), robot initial state (48.2\% against 75.9\%), and sensor noise (53.4\% against 79.0\%), which shift the third-person part of the key or change the kinematics, while lighting costs little.
On paraphrased instructions the cache is ahead (90.3\% against 84.9\%), because it uses the instruction only to pick the task, whereas $\pi_{0.5}$ conditions on the rewritten text.

\paragraph{Three policy episodes adapt the cache to a new scene.}
For a new scene, $\pi_{0.5}$ first runs three episodes from initial states other than the test one, and their decisions are encoded with the key of the cache.
Three steps then refit the cache on a CPU in seconds.
When the nearest-neighbor distances of the new decisions grow at least 1.5$\times$ more on the third-person features than on the wrist features, a ridge map, shrunk toward the identity, translates the new third-person features onto those of the paired library entries.
The metric of Equation~\ref{eq:metric} is then refit on the library together with the new decisions, which finally join the library with the chunks of $\pi_{0.5}$.
Serving again makes no policy calls.
We chose these steps on 40 variants from five categories (background, lighting, camera, sensor noise, and robot initial state; added objects and paraphrases do not fit this protocol) and apply them unchanged to all 3,451 variants of these categories.
Over all variants except added objects, adaptation alone raises the success rate from 62.6\% to 74.9\%, against 79.5\% for $\pi_{0.5}$, and on LIBERO-Spatial it matches $\pi_{0.5}$ (88.9\% against 88.6\%).

\paragraph{With the stagnation check, \name comes within 2.4 points of $\pi_{0.5}$ while calling it for less than a tenth of the decisions.}
Adding the stagnation check of Section~\ref{sec:correct}, whose policy calls run one denoising step, raises the success rate to 77.1\% with policy calls at 8.8\% of the decisions, so the gap to $\pi_{0.5}$ shrinks from 16.9 points without adaptation to 2.4 points.
On LIBERO-Spatial, \name reaches 89.5\% against 88.6\% for $\pi_{0.5}$ with calls at 6.6\% of the decisions.
The gains are largest where the unadapted cache lost most: camera viewpoint rises from 29.6\% to 63.5\% (65.2\% for $\pi_{0.5}$), sensor noise from 53.4\% to 84.2\%, above $\pi_{0.5}$ (79.0\%), and robot initial state from 48.2\% to 63.0\%.

\begin{table}[!htb]
\centering
\footnotesize
\begin{tabular}{@{}lrrrrr@{}}
\toprule
Perturbation & $\pi_{0.5}$ & No adaptation & Adaptation & + Stagnation check & Policy calls (\%) \\
\midrule
\multicolumn{6}{@{}l}{\emph{LIBERO-Spatial}} \\
All except added objects & 88.6 & 74.5 & 88.9 & 89.5 & 6.6 \\
Lighting & 95.9 & 99.3 & 99.0 & 99.3 & 0.4 \\
Background & 98.8 & 93.4 & 89.9 & 88.8 & 4.0 \\
Camera viewpoint & 80.9 & 35.9 & 78.2 & 80.1 & 9.6 \\
Sensor noise & 88.0 & 72.6 & 94.6 & 95.2 & 5.4 \\
Robot initial state & 83.1 & 54.6 & 73.1 & 74.9 & 9.9 \\
Paraphrased instruction & 89.2 & 100.0 & 100.0 & 100.0 & n/a \\
Added objects & 93.8 & 88.8 & n/a & n/a & n/a \\
\midrule
\multicolumn{6}{@{}l}{\emph{LIBERO-Long}} \\
All except added objects & 71.3 & 51.7 & 62.1 & 65.7 & 9.5 \\
Lighting & 75.2 & 67.9 & 59.5 & 65.3 & 8.0 \\
Background & 86.2 & 72.3 & 69.6 & 73.4 & 7.0 \\
Camera viewpoint & 51.1 & 23.9 & 48.2 & 48.7 & 12.1 \\
Sensor noise & 71.9 & 38.3 & 70.2 & 75.7 & 8.6 \\
Robot initial state & 69.5 & 42.5 & 46.1 & 52.4 & 10.1 \\
Paraphrased instruction & 80.4 & 80.4 & 80.4 & 80.4 & n/a \\
Added objects & 78.2 & 57.4 & n/a & n/a & n/a \\
\midrule
\multicolumn{6}{@{}l}{\emph{Both suites}} \\
All except added objects & 79.5 & 62.6 & 74.9 & 77.1 & 8.8 \\
Lighting & 85.9 & 84.1 & 79.9 & 82.9 & 6.2 \\
Background & 92.1 & 82.3 & 79.2 & 80.6 & 6.3 \\
Camera viewpoint & 65.2 & 29.6 & 62.4 & 63.5 & 11.5 \\
Sensor noise & 79.0 & 53.4 & 80.9 & 84.2 & 8.0 \\
Robot initial state & 75.9 & 48.2 & 58.8 & 63.0 & 10.0 \\
Paraphrased instruction & 84.9 & 90.3 & 90.3 & 90.3 & n/a \\
Added objects & 86.8 & 74.7 & n/a & n/a & n/a \\
\bottomrule
\end{tabular}
\vspace{-2.7pt}
\caption{Adaptation and the stagnation check recover most of the success rate of $\pi_{0.5}$ in unseen scenes. Success rate (\%) on LIBERO-plus with the official protocol of one test episode per variant. Without the stagnation check the cache never calls the policy, and Policy calls gives the share of decisions that call it with the check. Added objects cannot be adapted because each variant has a single initial state, and paraphrased instructions reuse the unadapted result because the key does not see the instruction.}
\label{tab:ood}
\end{table}

\paragraph{Adaptation needs the learned metric.}
We repeat the experiment with the Euclidean distance on standardized keys in place of the metric of Equation~\ref{eq:metric}.
Everything else stays the same, including the ridge map and the new decisions that join the library, and only the refit of the metric is dropped.
Over all variants except added objects, adaptation then reaches 66.8\% instead of 74.9\%, and with the stagnation check 70.2\% instead of 77.1\%, while the policy is called slightly more often (9.5\% against 8.8\% of the decisions).
Both gaps are significant on each suite (paired McNemar test, $p < 10^{-10}$).
Table~\ref{tab:ood-euclid} shows that the gap is largest on background and lighting.
These perturbations change many features that do not affect the action.
The refit metric gives such features little weight, whereas the Euclidean distance weighs them like any other difference.

\begin{table}[!htb]
\centering
\footnotesize
\begin{tabular}{@{}lrrrr@{}}
\toprule
 & \multicolumn{2}{c}{Adaptation} & \multicolumn{2}{c}{+ Stagnation check} \\
\cmidrule(lr){2-3}\cmidrule(l){4-5}
Perturbation & Learned & Euclidean & Learned & Euclidean \\
\midrule
LIBERO-Spatial, all except added objects & 88.9 & 82.8 & 89.5 & 84.2 \\
LIBERO-Long, all except added objects & 62.1 & 52.2 & 65.7 & 57.5 \\
\midrule
\multicolumn{5}{@{}l}{\emph{Both suites}} \\
All except added objects & 74.9 & 66.8 & 77.1 & 70.2 \\
Lighting & 79.9 & 69.6 & 82.9 & 70.1 \\
Background & 79.2 & 60.5 & 80.6 & 63.1 \\
Camera viewpoint & 62.4 & 56.6 & 63.5 & 59.6 \\
Sensor noise & 80.9 & 73.2 & 84.2 & 79.4 \\
Robot initial state & 58.8 & 47.1 & 63.0 & 54.5 \\
\bottomrule
\end{tabular}
\vspace{-2.7pt}
\caption{Adaptation gains much less with the Euclidean distance. Success rate (\%) on LIBERO-plus as in Table~\ref{tab:ood}, with the learned metric of Equation~\ref{eq:metric} or the Euclidean distance on standardized keys. Paraphrased instructions are left out because they reuse the unadapted result.}
\label{tab:ood-euclid}
\end{table}

\paragraph{Cost of adaptation.}
Adaptation costs three extra policy episodes per scene and pays off when a scene is served for many episodes.
Test episodes follow the official protocol, and the adaptation steps and the 1.5 threshold, chosen on the 40 selection variants, are held fixed for all variants.

\paragraph{A takeover at a distance jump recovers from objects moved during an episode.}
We also change the scene during an episode.
We move the first object of each LIBERO-Long task by 5~cm in a random horizontal direction once per episode, in most episodes just before the gripper closes on it, and run $\pi_{0.5}$ with a 200-episode library (Table~\ref{tab:moved}).
Without the move, \name succeeds in 90.4\% of these episodes, and with it even pure inference falls to 61.2\%.
\name falls to 48.8\%, since the stagnation check acts only after progress stops, and in the ten decisions after the move it calls the policy 1.8 times against 5 for pure inference.
A moved object shows up as a jump of the normalized distance $r$ of the corrector (Appendix~\ref{app:impl}).
We hand the next $K$ decisions to the policy whenever $r$ exceeds the 99th percentile of its values in episodes without the move, which happens at under 0.5\% of the decisions before the move.
The jump is caught in about 78\% of the moved episodes, usually within two decisions.
With $K=15$, \name succeeds in 63.2\%, as often as pure inference ($p=0.46$), while calling the policy 21.0 times per episode against 38.1.
Unlike a distance threshold, which decides every cache hit, this trigger sits far above the usual distances and fires only on a rare jump.

\begin{table}[!htb]
\centering
\footnotesize
\begin{tabular*}{\columnwidth}{@{\extracolsep{\fill}}lcc@{}}
\toprule
Setting & Success (\%) & Policy calls per episode \\
\midrule
Pure inference & 61.2 & 38.1 \\
\name & 48.8 & 15.9 \\
\quad without the stagnation check & 45.6 & 12.8 \\
\quad with takeover, $K=5$ & 57.8 & 20.5 \\
\quad with takeover, $K=15$ & 63.2 & 21.0 \\
\bottomrule
\end{tabular*}
\vspace{-8.0pt}
\caption{A takeover on a jump of the retrieval distance brings \name back to pure inference when an object is moved during the episode. $\pi_{0.5}$ on LIBERO-Long with a 200-episode library, 500 episodes per setting, with the first object of each task moved by 5~cm once per episode.}
\label{tab:moved}
\end{table}

\FloatBarrier

\end{document}